\documentclass{article}
\PassOptionsToPackage{numbers,sort&compress}{natbib}
\usepackage[preprint]{format}
\usepackage{verbatim}
\usepackage{comment}
\usepackage[utf8]{inputenc}
\usepackage[T1]{fontenc}
\usepackage[colorlinks=true,
    citecolor=blue,
    linkcolor=red,
    anchorcolor=green]{hyperref}
\usepackage{url}
\usepackage{booktabs}
\usepackage{amsfonts}
\usepackage{nicefrac}
\usepackage{microtype}
\usepackage{multirow}
\usepackage{adjustbox}
\usepackage[table]{xcolor}
\usepackage{colortbl}
\usepackage{xspace}
\usepackage{algorithm}
\usepackage{algorithmicx}
\usepackage{algpseudocode}
\newcommand{\mycomment}[1]{\hfill\textcolor{gray}{\footnotesize$\triangleright$ #1}}

\usepackage{balance}
\usepackage{subfigure}
\usepackage{graphicx}
\usepackage{mdframed}
\usepackage{mathtools}
\usepackage[noabbrev,capitalise]{cleveref}
\usepackage{caption}
\usepackage{float}
\usepackage{amsthm}
\usepackage{amsmath}
\usepackage{bm}
\usepackage{ragged2e}
\usepackage{colortbl}
\usepackage{tcolorbox}
\usepackage{pifont}
\usepackage{makecell}
\usepackage{enumitem}
\tcbuselibrary{skins, breakable}
\usepackage{setspace}
\usepackage{listings}
\usepackage{fancyhdr}
\usepackage{fontawesome5}
\usepackage{tabularx}
\usepackage[table]{xcolor}
\usepackage{array}
\usepackage[utf8]{inputenc}
\usepackage{CJKutf8}

\definecolor{hunyuanblue}{RGB}{232,242,255}
\newcommand{\hybcell}{\cellcolor{hunyuanblue}}
\newcommand{\hybr}{\rowcolor{hunyuanblue}}
\newcolumntype{H}{>{\columncolor{hunyuanblue}}c}
\newcolumntype{C}[1]{>{\centering\arraybackslash}m{#1}}
\newcolumntype{L}[1]{>{\raggedright\arraybackslash}m{#1}}

\newcommand{\cmark}{\ding{51}}

\newcommand{\mypara}[1]{\vspace{-0.2mm}\noindent\textbf{#1}\hspace{0.02cm}}

\newcommand{\logo}[1]{\raisebox{-0.2\height}{\includegraphics[height=1.2em]{figure/logo/org/#1}}\,}

\AtEndPreamble{
    \crefname{section}{Sec.}{Secs.}         \Crefname{section}{Sec.}{Secs.}
    \crefname{equation}{Eq.}{Eqs.}          \Crefname{equation}{Eq.}{Eqs.}
    \crefname{table}{Tab.}{Tabs.}           \Crefname{table}{Tab.}{Tabs.}
    \crefname{figure}{Fig.}{Figs.}          \Crefname{figure}{Fig.}{Figs.}
    \crefname{promptcount}{prompt}{prompts} \Crefname{promptcount}{Prompt}{Prompts}
    \crefname{algorithm}{Alg.}{Algs.}       \Crefname{algorithm}{Alg.}{Algs.}
    \creflabelformat{equation}{#2(#1)#3}
    \creflabelformat{table}{#2#1#3}
    \creflabelformat{figure}{#2#1#3}
    \crefformat{section}{\S#2#1#3}
    \Crefformat{section}{\S#2#1#3}
    \crefmultiformat{section}{\S#2#1#3}{, \S#2#1#3}{, \S#2#1#3}{, \S#2#1#3}
}

\newcommand{\thetitle}{}
\let\titleold\title
\renewcommand{\title}[1]{\titleold{#1}\renewcommand{\thetitle}{#1}}

\makeatletter
\newcommand{\safenolinenumbers}{\ifdefined\nolinenumbers\nolinenumbers\fi}
\newcommand{\safelinenumbers}{\ifdefined\linenumbers\linenumbers\fi}
\newcommand{\saferesetlinenumber}[1][]{\ifdefined\resetlinenumber\resetlinenumber[#1]\fi}
\makeatother

\def\maketitlesupplementary{
    {
            \newpage
            \par\safenolinenumbers
            \centering\Large
            \textbf{\thetitle}\\
            \vspace{0.3em}
            Supplementary Material \\
            \safelinenumbers
        }
}

\newcommand{\nextline}{\\} 
\newcommand{\ul}[1]{\underline{#1}}
\newcommand{\tbf}[1]{\textbf{#1}}

\fancypagestyle{firstpagestyle}{
    \fancyhf{}
    \fancyhead[L]{\includegraphics[height=0.66cm]{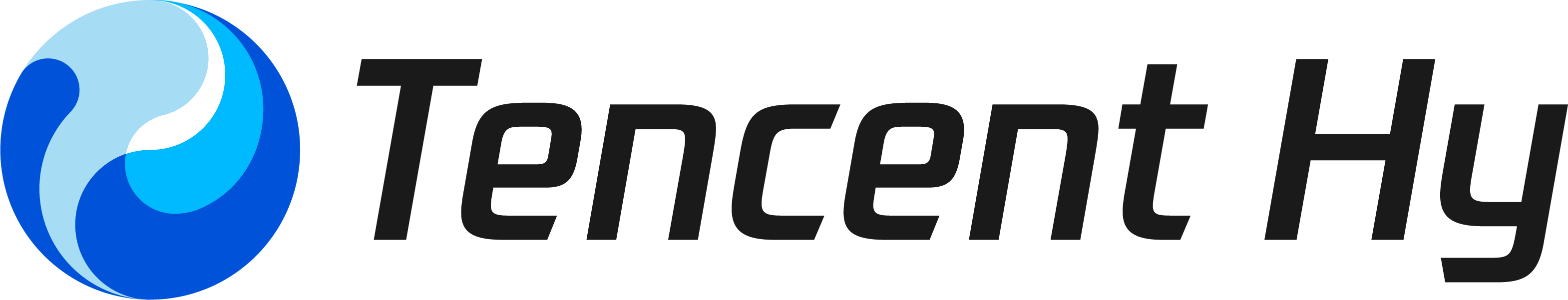}}
    \fancyhead[R]{\textcolor{gray}{\fontsize{11pt}{10pt}\selectfont July 7, 2026}}

}

\title{HunyuanOCR-1.5: Making Lightweight OCR VLMs Faster and Better}

\author{
    \begin{minipage}{\textwidth}
        \centering
        \textbf{Gengluo Li}$^{1,\,\scalebox{1}{$\ast$}}$\quad
        \textbf{Xingyu Wan}$^{2,\,\scalebox{1}{$\ast$}}$\quad
        \textbf{Shangpin Peng}$^{2,\,\scalebox{1}{$\ast$}}$\quad
        \textbf{Weinong Wang}$^{2,\,\scalebox{1}{$\ast$}}$\quad
        \textbf{Hao Feng}$^{2,\,\scalebox{1}{$\ast$}}$
        \\[0.3em]
        \textbf{Yongkun Du}$^{2,\,\scalebox{1}{$\ast$}}$\quad
        \textbf{Binghong Wu}$^{2}$\quad
        \textbf{Zheng Ruan}$^{2}$\quad
        \textbf{Zhiqiong Lu}$^{2}$\quad
        \textbf{Liang Wu}$^{2}$\quad
        \textbf{Pengyuan Lyu}$^{2}$
        \\[0.3em]
        \textbf{Huawen Shen}$^{2}$\quad
        \textbf{Zibin Lin}$^{2}$\quad
        \textbf{Shijing Hu}$^{2}$\quad
        \textbf{Jieneng Yang}$^{2}$\quad
        \textbf{Hongbing Wen}$^{2}$\quad
        \textbf{Guanghua Yu}$^{2}$\quad
        \\[0.3em]
        \textbf{Hong Liu}$^{2}$\quad
        \textbf{Bochao Wang}$^{2}$\quad
        \textbf{Can Ma}$^{1}$\quad
        \textbf{Han Hu}$^{2}$\quad
        \textbf{Chengquan Zhang}$^{2,\,\dagger,\,}$\textsuperscript{\scalebox{0.8}{\faEnvelope}}\quad
        \textbf{Yu Zhou}$^{3,\,}$\textsuperscript{\scalebox{0.8}{\faEnvelope}}
        \\[0.6em]
        {
        \normalfont
        $^{1}$Institute of Information Engineering, Chinese Academy of Sciences
        \\
        $^{2}$Large Language Model Department, Tencent\quad
        $^{3}$Nankai University
        }
        \\[0.6em]
        {
        \normalfont
        \small
        \texttt{
            ligengluo@iie.ac.cn\quad
            zchengquan@gmail.com\quad
            yzhou@nankai.edu.cn
        }
        }
    \end{minipage}
}

\begin{document}
\maketitle

\vspace{-2.5em}
\begin{center}
    \begin{tabular}{c l}
        \raisebox{-0.15\height}{\includegraphics[height=12pt]{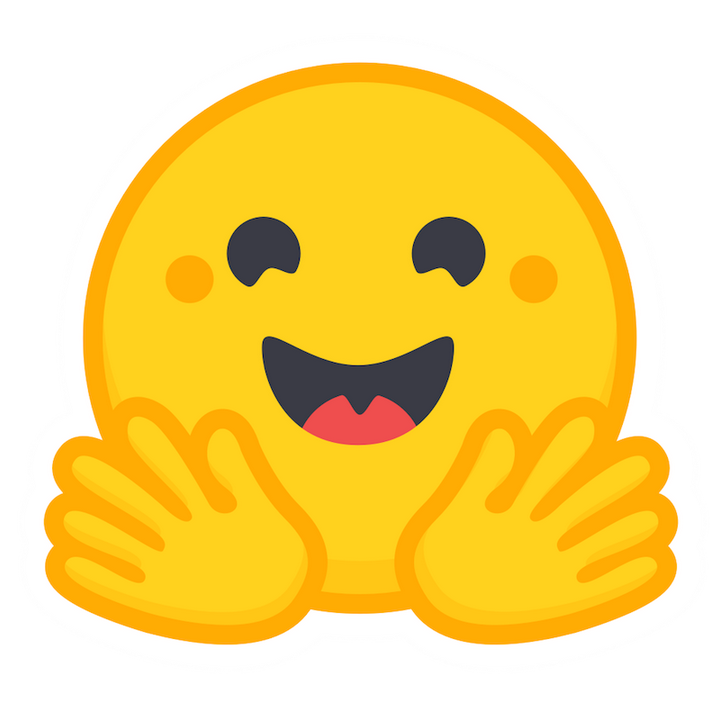}}
         &
        \url{https://huggingface.co/tencent/HunyuanOCR}
        \\[4pt]
        \raisebox{-0.15\height}{\includegraphics[height=12pt]{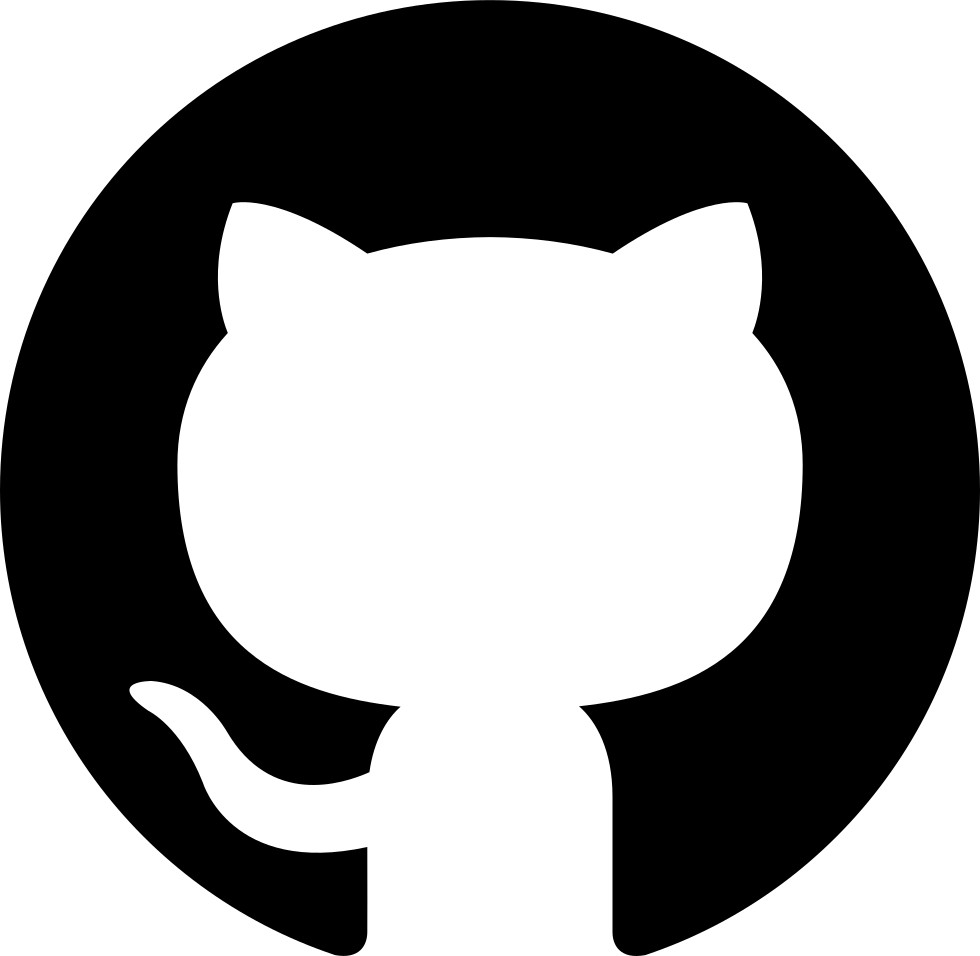}}
         &
        \url{https://github.com/Tencent-Hunyuan/HunyuanOCR}
    \end{tabular}
\end{center}
\vspace{1.5em}

\thispagestyle{firstpagestyle}

\let\oldthefootnote\thefootnote
\let\thefootnote\relax\footnotetext{
    $^{\scalebox{1.0}{\hspace{-0.7em} $\ast$}}$Equal contribution.
    \hspace{1em}
    $^{\dagger}$Project leader.
    \hspace{1em}
    \textsuperscript{\scalebox{0.8}{\faEnvelope}}Corresponding author.
}
\let\thefootnote\oldthefootnote

\begin{abstract}
    We present HunyuanOCR-1.5, a lightweight and end-to-end OCR-specialized vision-language model. HunyuanOCR targets a broad range of text-centric visual tasks, unifying document parsing, text spotting, information extraction, text-image translation, and multi-image document understanding within a single end-to-end VLM. Building upon the validated lightweight architecture of HunyuanOCR-1.0, HunyuanOCR-1.5 does not redesign the model backbone, but instead performs a systematic upgrade around two goals: making the model faster and better. For efficiency, we adapt \textit{DFlash} inference acceleration to OCR decoding, significantly reducing the decoding latency of long structured outputs such as dense documents, tables, and formulas while preserving the output distribution.
    Powered by \textit{DFlash}, HunyuanOCR-1.5 achieves a \textbf{6.37$\times$} speedup in Transformer inference and a \textbf{2.14$\times$} speedup under vLLM, delivering the fastest inference speed among all lightweight OCR VLMs.
    For capability, we propose \textit{Agentic Data Flow}, an agent-driven data construction system that transforms model weaknesses into executable data requirements and autonomously performs material search, quality verification, and data pipeline development. 
    Through this framework, we significantly enhance the model's long-tail capabilities across ancient-script OCR, fine-grained chart and table parsing, multi-image text-centric QA, low-resource multilingual parsing, and document hallucination evaluation. 
    Crucially, HunyuanOCR-1.5 stands as \textbf{the top-tier end-to-end OCR solution} on OmniDocBench v1.6, paired with unrivaled inference efficiency and new performance milestones across the aforementioned long-tail domains.
    Combined with an upgraded pretraining and post-training recipe, HunyuanOCR-1.5 further extends the capability boundary of the model in high-resolution, long-context, and multi-task scenarios. We characterize these upgrades through a capability-oriented evaluation, and experiments show that HunyuanOCR-1.5 achieves both faster inference and broader OCR capability coverage while retaining the deployment advantages of a lightweight end-to-end model. We will release the model weights and training code to the community to promote the research, reproduction, and real-world application of OCR-specialized vision-language models.
\end{abstract}

\section{Introduction}
\label{sec:introduction}

Visual text serves as the most ubiquitous and dense carrier of human knowledge. For decades, Optical Character Recognition (OCR)~\cite{long2021scene, zhang2024docparsing} has been the foundational technology for digitizing this information, traditionally functioning as a simple text transcription tool. However, as the demand for machine intelligence grows, this narrow definition is no longer sufficient. Modern applications require a comprehensive interface capable of supporting diverse text-centric visual tasks, ranging from document parsing~\cite{OmniDocBench_2025, Chronicles_OCR_2026, ChartArena_2026} and information extraction to visual question answering~\cite{OCRBench_2024}, text-image translation~\cite{DOTA_2024, MMTIT_Bench_2026}, and multi-image document understanding~\cite{DUDE_2023, MMLongBench_Doc_2024}.

To tackle these complex tasks, traditional cascaded pipelines relying on disjointed modules for detection, recognition, and downstream processing often struggle with error propagation and architectural redundancy. In contrast, the development of vision-language models (VLMs)~\cite{GPT_4o_2024, Gemini_2023, Gemini_1_5_2024, Gemini_2_5_2025, Gemini_3_2026, Gemini_3_1_2026, Qwen_VL_2023, Qwen2_VL_2024, Qwen2_5_VL_2025, Qwen3_VL_2025, InternVL_2024, InternVL1_5_2024, Mini_InternVL_2024, InternVL2_2024, InternVL2_5_2024, InternVL3_2025, InternVL3_5_2025} has paved the way for an elegant, end-to-end alternative. Driven by this trend, the community is developing OCR-specialized VLMs. These models are expected to tackle OCR as a unified visual-text understanding problem, where fine-grained perception, layout modeling, structured generation, and semantic reasoning are jointly performed within a single architecture.

However, most existing OCR-specialized VLMs are still primarily designed around document parsing, with the objective often restricted to converting a single page into structured outputs such as Markdown, HTML, or LaTeX. While document parsing is indeed one of the core OCR capabilities, the demands of real-world OCR go far beyond it, including text spotting in open scenarios, structured field extraction, question answering grounded in textual images, multilingual text-image translation, and reasoning across multiple pages or images. Fundamentally, a true OCR-specialized model should not be reduced to a mere document parser, but should be a unified end-to-end model covering diverse OCR tasks.

HunyuanOCR-1.0~\cite{HunyuanOCR_2025} has validated the feasibility of this philosophy: a lightweight, end-to-end OCR-specialized VLM that achieves leading performance across document parsing, text spotting, information extraction, visual question answering, and text-image translation. It demonstrated the effectiveness of unifying OCR capabilities within a compact architecture and highlighted the crucial role of high-quality OCR data and task-oriented training strategies in building practical OCR systems. Building on this foundation, HunyuanOCR-1.5 does not pursue a redesign of the model architecture, but instead addresses a more deployment-oriented question: \textit{on top of the validated HunyuanOCR framework, how can the model become faster and better}?

\mypara{Faster: DFlash-based inference acceleration.}
End-to-end OCR is often accompanied by long autoregressive decoding~\cite{Attention_2017}. In scenarios such as dense documents, tables, formulas, and long structured outputs, the decoding overhead becomes a major bottleneck in practical deployment~\cite{MinerU_Diffusion_2026}. To this end, HunyuanOCR-1.5 introduces a speculative decoding~\cite{leviathan2023fast, EAGLE_2024, EAGLE_2_2024, EAGLE_3_2026, cai2024medusa} framework based on DFlash~\cite{chen2026dflash} for inference acceleration: a lightweight block-diffusion~\cite{Block_Diffusion_2025, SDAR_2026, Fast_dLLM_v2_2025} draft model drafts multiple candidate tokens in parallel, which are then verified by the target model in a single pass. While preserving the output distribution of the target model, this significantly improves decoding efficiency for long outputs, achieving a $6.37\times$ speedup with Transformers and a $2.14\times$ speedup with vLLM in our evaluation, making the model more practical in real-world deployment environments that demand both accuracy and speed. 
Beyond server-grade deployment with vLLM~\cite{vLLM_2023}, HunyuanOCR-1.5 also supports PC-side inference through llama.cpp~\cite{llama_cpp_2023}, enabling deployment on CPUs, consumer GPUs, and laptops.

\mypara{Better: Agentic data flow and refined training recipes.}
Driven by comprehensive upgrades on both the data and training sides, HunyuanOCR-1.5 establishes itself as the \textbf{SOTA end-to-end OCR solution} on OmniDocBench v1.6. To achieve this capability boundary extension, we propose \textit{Agentic Data Flow} on the data side, an agent-driven data-construction system that translates model weaknesses into executable data requirements~\cite{AgentInstruct_2024, TaskCraft_2025, MetaSynth_2025}. Different from conventional pipelines that rely entirely on manually written scripts and manually collected materials, Agentic Data Flow allows agents to deeply participate in material search, tool-based verification, sample cleaning, and data pipeline development, and to iterate in a closed loop with algorithm engineers. In HunyuanOCR-1.5, this system is used for targeted data construction of long-tail capabilities such as low-resource OCR, ancient-script OCR~\cite{Chronicles_OCR_2026}, and multi-image QA~\cite{DUDE_2023, MMLongBench_Doc_2024}.

On the training side, we systematically upgrade the training recipe around capability boundary extension. In the pretraining stage, we revisit and re-plan Stage3 of HunyuanOCR-1.0, incorporating the new capability data produced by Agentic Data Flow, multi-image data, and historical OCR data, while increasing the maximum image resolution to 4K and extending the context window to 128K, so that the model can robustly adapt to high-resolution documents, long contexts, and multi-page or multi-image inputs. In the post-training stage, we refine the SFT data and introduce new high-quality training data and further explore RL across different OCR tasks to amplify the gains brought by reinforcement learning.

To systematically characterize the practical benefits of these upgrades, HunyuanOCR-1.5 is evaluated from a capability-oriented perspective rather than relying on a single benchmark. The evaluation covers both inherited and newly added capabilities, including end-to-end document parsing~\cite{OmniDocBench_2025}, text spotting, multilingual OCR, ancient-script recognition~\cite{Chronicles_OCR_2026}, text-image translation~\cite{DOTA_2024, MMTIT_Bench_2026}, multi-image QA~\cite{DUDE_2023, MMLongBench_Doc_2024}, information extraction, and hallucination-related reliability. This evaluation perspective is aligned with the design goal of HunyuanOCR-1.5: extending HunyuanOCR into a faster and more comprehensive unified end-to-end OCR-specialized VLM. In addition, we plan to release all the model weights and training code of HunyuanOCR-1.5 to the community, providing infrastructure for reproducing, fine-tuning, and extending OCR-specialized VLMs, and further promoting research and applications in OCR perception, document understanding, and multi-task modeling.

The main contributions of this report are summarized as follows:
\begin{itemize}[
    label=\raisebox{0.5ex}{\tiny$\bullet$},
    leftmargin=1em,
    itemsep=2pt,
    parsep=0pt,
    topsep=0pt,
    partopsep=0pt
]
    \item We present \textbf{HunyuanOCR-1.5}, an upgraded lightweight end-to-end OCR-specialized VLM that further extends diverse OCR task capabilities on top of HunyuanOCR-1.0, and we plan to release the model weights and training code to support community reproduction, fine-tuning, and capability extension.

    \item We adapt \textbf{DFlash} to HunyuanOCR inference and support PC-side deployment through llama.cpp, significantly improving the decoding efficiency of long structured OCR outputs while enabling both server-grade and local OCR deployment.

    \item We propose \textbf{Agentic Data Flow} and systematically upgrade the training recipe: an agent-driven data system produces long-tail capability data such as low-resource OCR, ancient-script OCR, and multi-image QA; in pretraining, we re-plan Stage3 and extend to 4K resolution and a context of 128K; and in post-training, we improve the capability ceiling through high-quality SFT data and task-specific RL exploration.
\end{itemize}
\section{Related Work}
\label{sec:related_work}

\mypara{General vision-language models.}
Recent general VLMs have demonstrated strong multimodal perception and reasoning abilities and have shown promising OCR-related capabilities in diverse visual scenarios. Representative models, such as GPT-4o~\cite{GPT_4o_2024}, Gemini~\cite{Gemini_2023, Gemini_1_5_2024, Gemini_2_5_2025, Gemini_3_2026, Gemini_3_1_2026}, Qwen-VL~\cite{Qwen_VL_2023, Qwen2_VL_2024, Qwen2_5_VL_2025, Qwen3_VL_2025}, and InternVL~\cite{InternVL_2024, InternVL1_5_2024, Mini_InternVL_2024, InternVL2_2024, InternVL2_5_2024, InternVL3_2025, InternVL3_5_2025}, can recognize text in natural images, documents, charts, and screenshots and further perform text-centric question answering or reasoning based on visual content. However, these models are primarily designed as general-purpose multimodal assistants rather than OCR-specialized systems. As a result, they often require large model sizes and high inference costs, and their performance may become unstable in OCR-intensive scenarios that require fine-grained text perception, dense document parsing, strict reading order preservation, or faithful structured output. In addition, general VLMs are not explicitly optimized for deployment-oriented OCR workloads, especially high-resolution long-document parsing and large-scale production serving.

\mypara{OCR-specific vision-language models.}
To address the limitations of general VLMs in OCR-centric scenarios, recent works have explored OCR-specific vision-language models~\cite{HunyuanOCR_2025, PaddleOCR_VL_2025, PaddleOCR_VL_1_5_2026}. Most existing OCR expert VLMs are primarily designed for document parsing~\cite{Infinity_parser_2025, HSD_2026, Logics_Parsing_2025, Logics_Parsing_Omni_2026}, aiming to convert page-level document images into structured outputs such as Markdown, HTML, or LaTeX. In addition to large OCR expert models, recent lightweight designs have also shown promising results. For example, UniRec-0.1B~\cite{du2025unirec} optimizes a compact 0.1B-parameter model for text blocks and formula blocks, demonstrating competitive OCR performance under a highly lightweight setting. According to their modeling paradigm, OCR-specific VLMs can be roughly divided into modular and end-to-end approaches. Modular methods usually cascade a layout analysis model before OCR recognition: a page-level document is first decomposed into block-level regions, and each block is then parsed by an OCR VLM. Such designs can reduce the difficulty of local parsing through region-level cropping, but the overall pipeline still depends on the preceding layout analysis results and may suffer from errors in region detection, reading-order recovery, and cross-block relation modeling.
In contrast, end-to-end OCR-specific models directly model page-level documents and parse the entire page image within a unified framework, with representative examples including dots.ocr~\cite{dots_ocr_2025} and DeepSeek-OCR~\cite{DeepSeek_OCR_2025}. This paradigm avoids explicit layout splitting and the associated error propagation, allowing the model to jointly capture text, tables, formulas, charts, and reading order in the full-page context. Therefore, end-to-end modeling is more beneficial for improving the native OCR capability of VLMs. We argue that an OCR-specialized VLM should not be defined only as a document parsing model but should support a broader range of OCR-related tasks, including text spotting, information extraction, document question answering, text-image translation, and multi-image understanding. HunyuanOCR~\cite{HunyuanOCR_2025} follows the lightweight end-to-end OCR-specific VLM paradigm, and HunyuanOCR-1.5 further extends this direction by keeping the validated architecture unchanged while improving capability boundaries through data construction, recipe upgrades, and inference acceleration.

\mypara{Multi-token prediction.}
Autoregressive decoding~\cite{Attention_2017} is a key latency bottleneck for long-output OCR scenarios such as document parsing, table reconstruction, and formula transcription. Speculative decoding accelerates generation through a draft-then-verify paradigm~\cite{Speculative_Decoding_survey_2024}, where candidate tokens proposed by a lightweight draft model are verified by the target model while preserving the original output distribution. However, many speculative methods still rely on autoregressive drafting, so the draft cost grows with the number of proposed tokens. Recent parallel drafting methods, such as multi-head prediction~\cite{cai2024medusa, gloeckle2024better} and diffusion-based generation, aim to predict multiple future tokens simultaneously. Among them, DFlash~\cite{chen2026dflash} trains a block-diffusion draft model conditioned on target-model hidden states, enabling an entire candidate block to be proposed in one parallel forward pass and then verified by the target model. This makes DFlash well-suited for HunyuanOCR-1.5, where OCR-centric generation often produces long and structured outputs.

\section{Model Design}
\label{sec:model_design}

\subsection{Model Architecture}
\label{subsec:model_architecture}

\begin{figure*}[t]
    \centering
    \includegraphics[width=\textwidth]{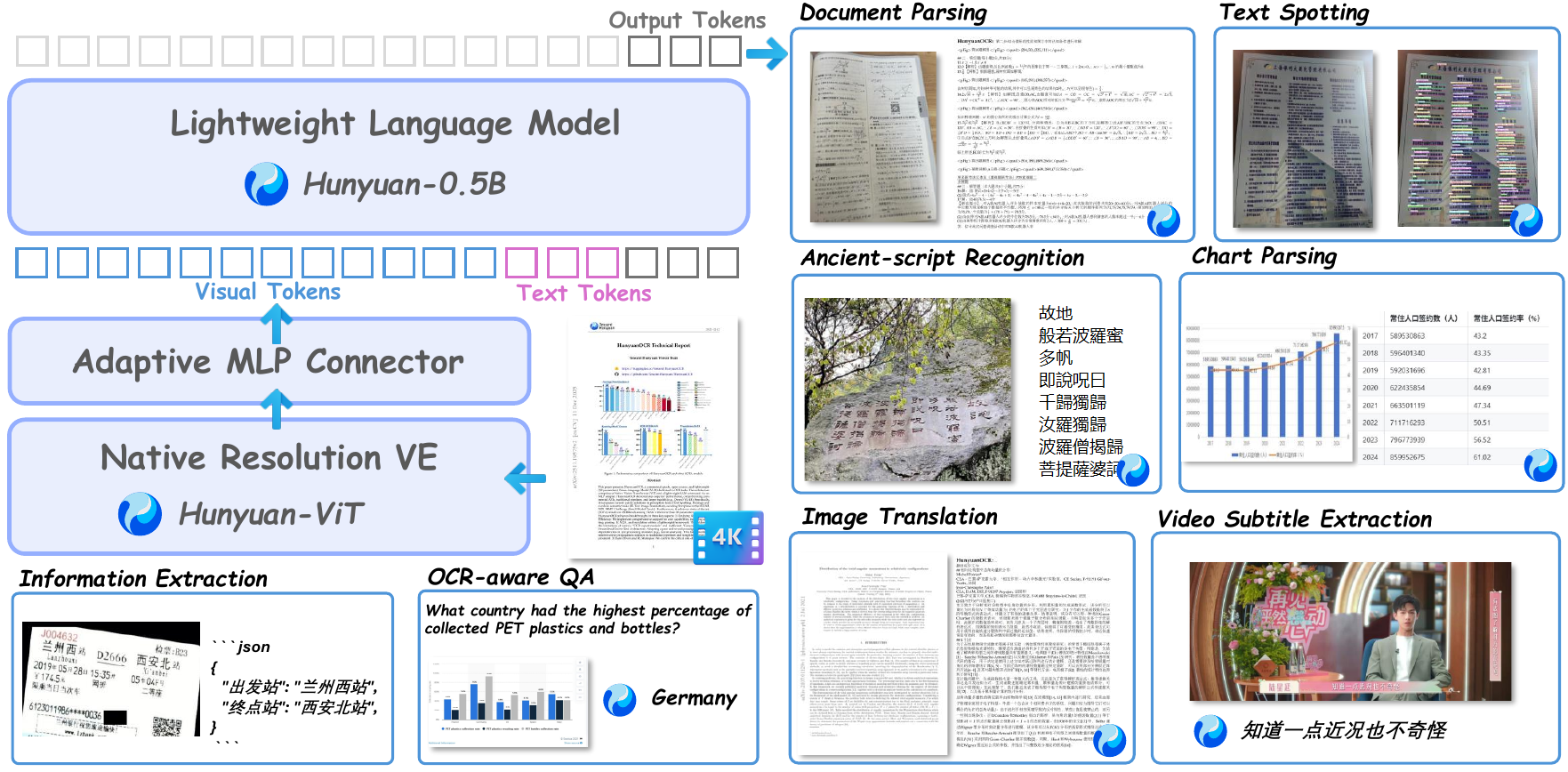}
    \vspace{-1.6em}
    \caption{
        \textbf{Overview of the HunyuanOCR-1.5 architecture.}
        A compact end-to-end model that unifies diverse OCR-centric capabilities, including document parsing, text spotting, information extraction, OCR-aware QA, ancient-script recognition, chart parsing, image translation, and video subtitle extraction.
    }
    \vspace{-0.4em}
    \label{fig:model_arch}
\end{figure*}

HunyuanOCR-1.5 follows the compact, fully end-to-end architecture of HunyuanOCR-1.0~\cite{HunyuanOCR_2025}, comprising a native-resolution visual encoder, an adaptive MLP connector, and a lightweight language model (\cref{fig:model_arch}). The pivotal upgrade in the model backbone lies in the visual encoder: built upon Hunyuan-ViT~\cite{ViT_2020, NaViT_2023}, we extend the maximum input image resolution from 2K to 4K. This crucial scaling allows the model to preserve native aspect ratios and spatial layouts while capturing finer structural details, which is instrumental for processing highly dense documents, oversized tables, and complex charts.

The remaining components maintain their validated lightweight configurations to ensure deployment efficiency. The adaptive MLP connector compresses high-resolution visual features into compact tokens while preserving layout sensitivity. Concurrently, the language component, a lightweight Hunyuan-0.5B model with XD-RoPE~\cite{hunyuan2025llm}, processes these tokens to autoregressively generate structured OCR outputs. Through this streamlined formulation, HunyuanOCR-1.5 directly maps multi-modal inputs into diverse OCR-centric outputs (e.g., Markdown documents, HTML tables, LaTeX formulas, and chart descriptions) without relying on any task-specific post-processing modules.

\subsection{Multi-token Prediction}
\label{subsec:multi_token_prediction}

Autoregressive decoding~\cite{Attention_2017} is a major efficiency bottleneck for end-to-end OCR-centric VLMs, especially in document parsing scenarios~\cite{Infinity_parser_2025, HSD_2026, Logics_Parsing_2025, Logics_Parsing_Omni_2026} that require long structured outputs, such as dense tables, multi-column documents, and long formulas. Although speculative decoding~\cite{leviathan2023fast, EAGLE_2024, EAGLE_2_2024, EAGLE_3_2026, cai2024medusa} reduces latency by drafting multiple candidate tokens and verifying them with the target model, many existing methods still generate draft tokens autoregressively, making the draft cost grow with the number of candidates. To address this limitation, HunyuanOCR-1.5 adopts DFlash~\cite{chen2026dflash}, which uses a lightweight block-diffusion draft model to predict a block of candidate tokens in one parallel forward pass. Given a block size $B$, the draft model proposes $\hat{\mathbf{y}}_{1:B}$ at once, and the target model verifies the block in parallel and accepts the longest valid prefix, preserving the output distribution of the target model.

\begin{figure*}[t]
    \centering
    \includegraphics[width=0.75\textwidth]{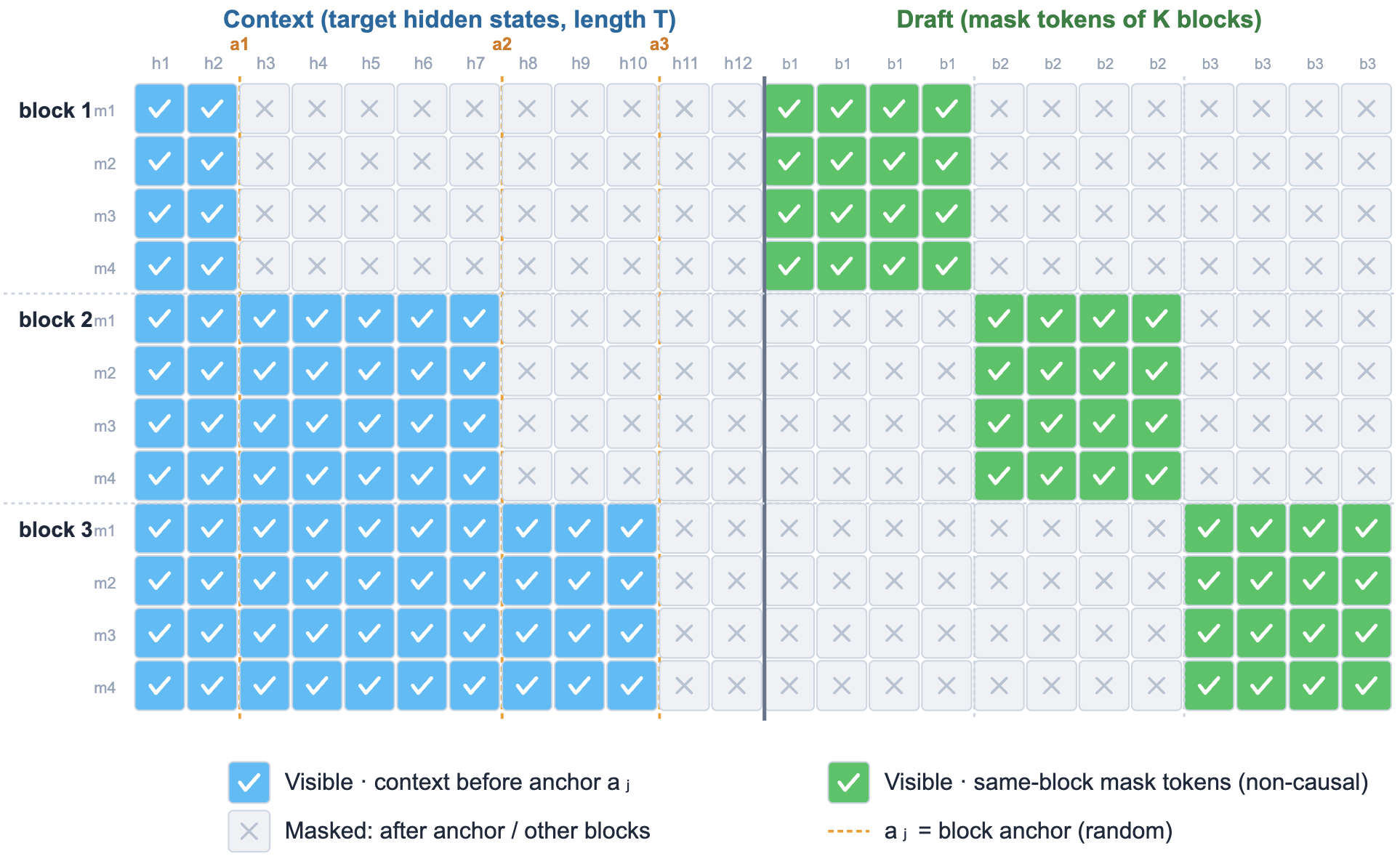}
    \vspace{-0.2em}
    \caption{
        \textbf{Overview of DFlash training with a joint FlexAttention mask.}
        One target forward is performed, $K$ anchors are sampled at random positions, and all $K$ blocks attend in a single pass. Rows denote Query tokens, and columns denote Key/Value tokens.
    }
    \vspace{-0.6em}
    \label{fig:dflash_mask}
\end{figure*}

During training, the HunyuanOCR-1.5 target model is frozen, and only the DFlash draft model is optimized. For each training sequence, we first run the target model once and cache its hidden states as conditional representations. We then randomly sample $K$ anchor positions, each corresponding to an independent block-drafting task. These $K$ blocks are concatenated and trained in a single forward pass with a FlexAttention~\cite{Flex_Attention_2024} block-diagonal mask, where each block can attend to the target hidden states before its anchor and to the mask tokens within the same block, while different blocks remain isolated, as shown in~\cref{fig:dflash_mask}. We use ground-truth continuation tokens as labels and optimize a position-weighted next-token cross-entropy loss:

\begin{equation}
    w^{(j)}_{k}
    =
    \mathbf{1}[k>0]\cdot\mathbf{1}[\mathrm{valid}]
    \cdot
    \exp\!\left(-\max(k-1,0)/\gamma\right),
\end{equation}
\begin{equation}
    \mathcal{L}_{\mathrm{DFlash}}
    =
    \frac{1}{Z}
    \sum_{j=1}^{K}
    \sum_{k=1}^{B-1}
    w^{(j)}_{k}
    \left[
        -\log p_{\theta}\!\left(
        y^{(j)}_{k} \mid \mathbf{h}_{<a_{j}},\, \mathbf{m}^{(j)}_{1:B}
        \right)
    \right],
    \quad
    Z = \sum_{j=1}^{K}\sum_{k=1}^{B-1} w^{(j)}_{k},
\end{equation}

where $a_j$ is the $j$-th anchor position, $\mathbf{h}_{<a_j}$ denotes the target hidden states before the anchor, $\mathbf{m}^{(j)}_{1:B}$ denotes the mask-token queries of the draft block, and $y^{(j)}_{k}$ is the ground-truth continuation token. The weight excludes the anchor token and invalid positions, while the exponential decay reduces the loss weight for farther positions that are harder to predict. In our implementation, the DFlash draft model has approximately 90.7M parameters, uses block size $B=16$, samples $K=16$ anchors per sequence, sets $\gamma=7.0$, and is implemented as a 5-layer Transformer initialized from the last 5 decoder layers of the target model.

\mypara{Discussion.}
DFlash~\cite{chen2026dflash} accelerates inference by trading otherwise underutilized computation for fewer autoregressive decoding steps. In single-request or low-concurrency scenarios, target-model AR decoding is often memory-bandwidth-bound, leaving a considerable amount of compute idle. DFlash leverages this idle compute to produce a block of draft tokens through a lightweight parallel forward pass, so that each target-model verification step can advance multiple tokens, thereby increasing the effective number of accepted tokens per target forward pass. This property is particularly beneficial for OCR outputs with strong local regularity, such as HTML tables, formulas, and structured Markdown parsing results, where future tokens are more predictable, and the accepted prefix length tends to be longer. As shown in~\cref{sec:exp_dflash_speed}, the acceleration ratio increases with output length and becomes significant for long, highly structured OCR generation tasks.

\section{Agentic Data Flow}
\label{sec:agentic_data_flow}

\begin{figure*}[t]
    \centering
    \includegraphics[width=\textwidth]{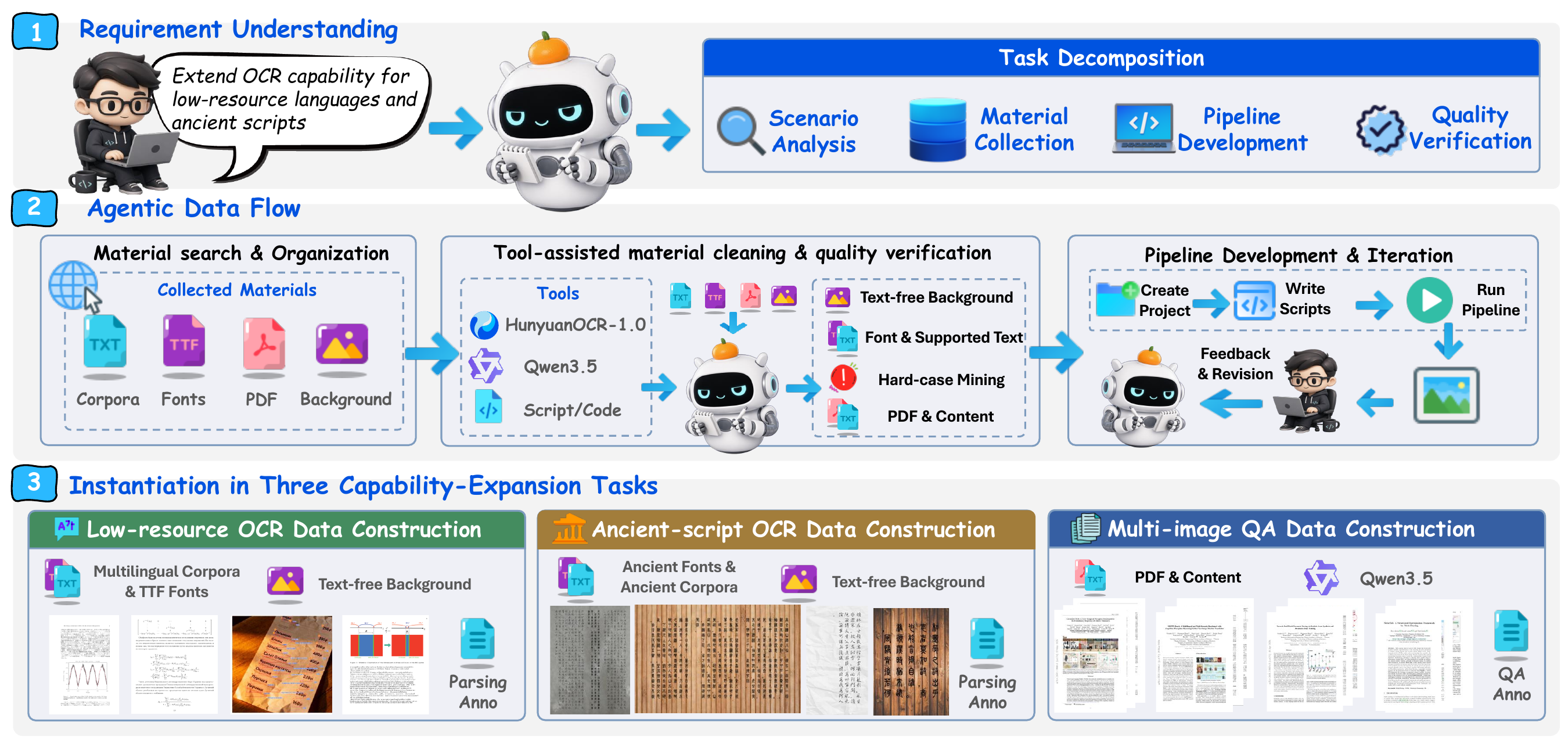}
    \vspace{-1.5em}
    \caption{
        \textbf{Overview of \textit{Agentic Data Flow}.}
        An agent-driven data construction system, instantiated in three capability-expansion tasks: low-resource OCR, ancient-script OCR, and multi-image QA.
    }
    \vspace{-2em}
    \label{fig:agentic_data_flow}
\end{figure*}

The capability boundary extension of HunyuanOCR-1.5 is not achieved by simply scaling up data volume, but is instead driven by a data construction system explicitly oriented toward addressing model weaknesses. We refer to this system as \textit{Agentic Data Flow}. It takes concrete capability gaps, such as insufficient coverage of low-resource languages, weak perception of ancient scripts, lack of multi-image document understanding, and insufficient hard cases in complex scenarios, and systematically converts them into executable data requirements that further drive the subsequent data production loop.

\subsection{Agentic Data Flow Pipeline}
\label{subsec:agentic_pipeline}

We equip the agent with tool-calling structures and usage instructions, enabling access to web search, OCR services, vision-language model services, file processing scripts, image cleaning tools, and data generation tools. Algorithm engineers provide target capability requirements in natural language, such as constructing synthetic data for low-resource OCR, generating ancient-script parsing samples, mining failure cases of HunyuanOCR-1.0, or constructing multi-image QA data. The agent autonomously decomposes the task, determines the required materials, tool calls, scripts, and quality criteria, and interacts with algorithm engineers during data production. By inspecting intermediate samples, identifying quality issues, and adding constraints, algorithm engineers iteratively refine the data pipeline, eventually forming a reusable workflow for the target weakness. Within this loop, as illustrated in~\cref{fig:agentic_data_flow}, the agent mainly undertakes three key operations.

\mypara{Material search and organization.}
The agent autonomously invokes web search and other tools to collect materials required for data construction. For low-resource OCR, it searches for multilingual text corpora, TTF font files, and rendering backgrounds. For ancient-script OCR, it searches for fonts related to the seven historical forms of Chinese characters, as well as backgrounds with ancient-book or historical-document styles. For multi-image QA, it mainly searches and organizes multi-page PDF documents and uses PDF tools to extract page-level text and structural information as the basic context for subsequent QA generation. Compared with manual collection, the agent can organize scattered resources into structured material directories and maintain mappings among corpora, fonts, backgrounds, PDF documents, and other resources.

\mypara{Tool-assisted material cleaning and quality verification.}
Beyond material collection, the agent further invokes tools to clean and refine the collected resources. For image backgrounds, it can call HunyuanOCR-1.0~\cite{HunyuanOCR_2025} and Qwen3.5~\cite{Qwen3_5_2026} services for automatic inspection, filtering out candidate images that contain interfering text, overly complex foreground objects, or unstable visual quality, thereby maintaining high-quality text-free backgrounds suitable for synthesis. For fonts, the agent tests the rendering compatibility of candidate TTF files for the corresponding languages or scripts and maintains the supported rendering vocabulary of each font. For hard-case mining, the agent can run HunyuanOCR-1.0 inference on candidate images in batch and automatically maintain hard-sample sets according to parsing results such as missed recognition, structural disorder, table parsing failure, and incorrect multi-column reading order.

\mypara{Weakness-oriented data pipeline development and iteration.}
After the materials are prepared, the agent autonomously develops data production pipelines for specific weakness topics. It creates data projects, writes rendering or QA generation scripts, organizes material paths, defines task formats, and progressively supports different layout renderings, background combinations, degradation augmentations, and output schemas. During development, the agent continuously interacts with algorithm engineers: it first generates initial demos and then performs multiple rounds of revision based on feedback regarding layout quality, visual realism, task difficulty, label format, and data diversity. As iteration proceeds, the pipeline gradually evolves from a single-template prototype into a data production system supporting multiple layouts, augmentations, and task formats.

\subsection{Instantiation on Capability-Expansion Tasks}
\label{subsec:agentic_instantiation}

Through the above mechanism, Agentic Data Flow connects model weakness identification, material construction, data cleaning, pipeline development, and training data injection into a closed loop. In HunyuanOCR-1.5, we instantiate this system in three representative capability-expansion tasks: low-resource OCR, ancient-script OCR, and multi-image QA.

\mypara{Low-resource OCR data construction.}
For low-resource OCR, the agent automatically collects multilingual text corpora and corresponding TTF font files from the web. Since different fonts vary significantly in character coverage, the agent tests the rendering compatibility of candidate fonts for each language and maintains a mapping among languages, fonts, and supported rendering vocabularies. Based on the design philosophy of SynthText~\cite{SynthText_2016} and SynthDoG~\cite{SynthDoG_2022}, the agent develops a multilingual synthetic data production pipeline that renders texts from different languages onto diverse backgrounds with controllable layouts and visual styles. Through this process, we maintain parsing data covering 331 languages, providing pretraining supervision to improve the multilingual perception capability of HunyuanOCR-1.5.

\mypara{Ancient-script OCR data construction.}
For ancient-script OCR, we focus on the seven historical forms of Chinese characters. For each historical script, the agent autonomously searches for multiple TTF font files with different rendering styles, and maintains diverse text-free background materials through both autonomous collection and tool-assisted verification. For example, when collecting background images, the agent can invoke HunyuanOCR-1.0 and Qwen3.5 services for multi-model validation, filtering out candidate images that contain interfering text or unstable visual quality. The agent then develops an ancient-script parsing data synthesis pipeline according to the writing directions, layout patterns, and visual styles of historical documents, supporting different backgrounds, fonts, layouts, and degradation augmentations. The generated data are mainly used in the pretraining stage to supplement rare historical character forms and improve the model's fundamental perception of ancient documents.

\mypara{Multi-image QA data construction.}
For multi-image document understanding, we extend Agentic Data Flow to a QA data production process based on multi-page PDFs. The agent first collects and organizes multi-page PDF documents, and invokes PDF tools to extract page-level text and basic structural information. The extracted text is then organized into cross-page contexts according to page order and provided to a strong text model to generate multi-image QA samples, including cross-page information retrieval, multi-page content comparison, evidence aggregation, and document-level reasoning questions. To ensure that the generated data truly require multi-page understanding, we further filter out questions that can be answered from a single page, samples whose answers are inconsistent with the extracted PDF context, and questions without explicit textual evidence. This pipeline extends the capability boundary of HunyuanOCR-1.5 from single-image OCR and single-page document parsing to multi-page and multi-image document understanding.

Overall, Agentic Data Flow serves as a capability-expansion data system for HunyuanOCR-1.5. It is not limited to a specific data type, but provides a reusable data construction paradigm: defining data requirements around model weaknesses, automatically completing material search, tool-based verification, sample cleaning, pipeline development, and human-agent iteration through agents, and injecting the resulting data into subsequent training stages. This system supports the improvement of HunyuanOCR-1.5 in long-tail directions such as low-resource languages, ancient scripts, multi-image understanding, and hard-case robustness.

\section{Training Recipe}
\label{sec:training_recipe}

The training recipe of HunyuanOCR-1.5 follows the staged training paradigm of HunyuanOCR~\cite{HunyuanOCR_2025}, while shifting the objective from building general OCR capabilities to extending capability boundaries and improving task ceilings. Overall, we structure the training pipeline of HunyuanOCR-1.5 into three main phases: \textit{pretraining} (\cref{subsec:pretraining}), \textit{supervised fine-tuning} (SFT, \cref{subsec:SFT}), and \textit{reinforcement learning} (RL, \cref{subsec:RL}). 

The pretraining stage mainly injects newly constructed capability-expansion data into the model and improves its adaptation to complex inputs through resolution and context-window extension. The subsequent SFT and RL stages collaboratively focus on \textbf{capability ceiling improvement}, pushing the upper bound of each OCR task while enhancing output stability and mitigating hallucinations in document scenarios. Specifically, SFT establishes a clean and highly structured foundation by refining the training data and unifying the prompt interface. Building upon this high-quality basis, RL further pushes the capability ceilings using verifiable rewards and judge-based supervision, forming a complementary optimization pipeline.

\subsection{Pretraining: Revisiting Stage3 for Capability Boundary Extension}
\label{subsec:pretraining}

In the pretraining stage, we do not redesign the full pretraining procedure of HunyuanOCR-1.0. Instead, we reuse its first two stages and only re-plan the third stage (Stage3). Two upgrades are applied to Stage3: (a) we inject new capability-expansion data to broaden what the model can recognize, and (b) we enlarge the input specification so that the model can handle high-resolution and long-context inputs.

\mypara{Data upgrade.} Stage3 now mixes three sources: the new capability data produced by \textit{Agentic Data Flow} (\cref{sec:agentic_data_flow}), multi-image understanding data, and historical OCR data from HunyuanOCR-1.0. The new data target the model's weak spots, covering low-resource OCR, ancient-script OCR, multi-image document understanding, hard cases, and long-tail layouts, and thus drive capability expansion. The historical OCR data are kept to preserve existing strengths in general OCR, document parsing, and structured output. Training on both jointly lets the re-planned Stage3 expand new capabilities without regressing on old ones.

\mypara{Input specification.} We also raise the maximum image resolution to 4K and extend the context window to 128K. This lets the model take in far more demanding inputs, such as dense documents, multi-page and multi-image contexts, and long structured outputs. As a result, HunyuanOCR-1.5 pushes its capability boundary toward high-resolution, long-context, and multi-image scenarios while keeping the architecture unchanged.
\subsection{SFT: Building a High-Quality Foundation for the RL Stage}
\label{subsec:SFT}

As the first step toward capability ceiling improvement, the SFT stage prepares a clean, well-organized, and interface-consistent training set for the subsequent RL stage. This preparation consists of three parts: refining the data quality, splitting the data between SFT and RL, and unifying the prompt design across tasks.

\mypara{Data refinement.}
We start from the post-training data of HunyuanOCR-1.0 and clean it thoroughly, removing annotation errors, format inconsistencies, image-text mismatches, ambiguous task objectives, and low-quality duplicated samples. We then enrich the data pool with the new capability data produced by Agentic Data Flow, user-provided hard cases, and high-quality data for the newly introduced capabilities, with careful manual annotation and verification for key samples. Through this process, the SFT data are upgraded from the general task coverage of HunyuanOCR-1.0 to a high-quality training set oriented toward capability ceiling improvement and robustness in complex scenarios.

\mypara{Data splitting for SFT and RL.}
We divide the curated data into two disjoint portions. One portion is used for supervised fine-tuning in the current stage, while the other, consisting mainly of high-difficulty samples, is reserved for the subsequent RL stage. This split lets SFT establish broad task competence while keeping the most challenging samples for RL to push the capability ceiling.

\mypara{Unified prompt design.}
Finally, we unify the prompt design across tasks, routing each task capability to its own specialized prompt. This reduces instruction ambiguity, gives each task a clear and consistent interface, and thereby provides well-defined task boundaries for the subsequent RL stage.
\subsection{Reinforcement Learning}
\label{subsec:RL}
\begin{figure}[t!]
    \centering
    \includegraphics[width=\linewidth]{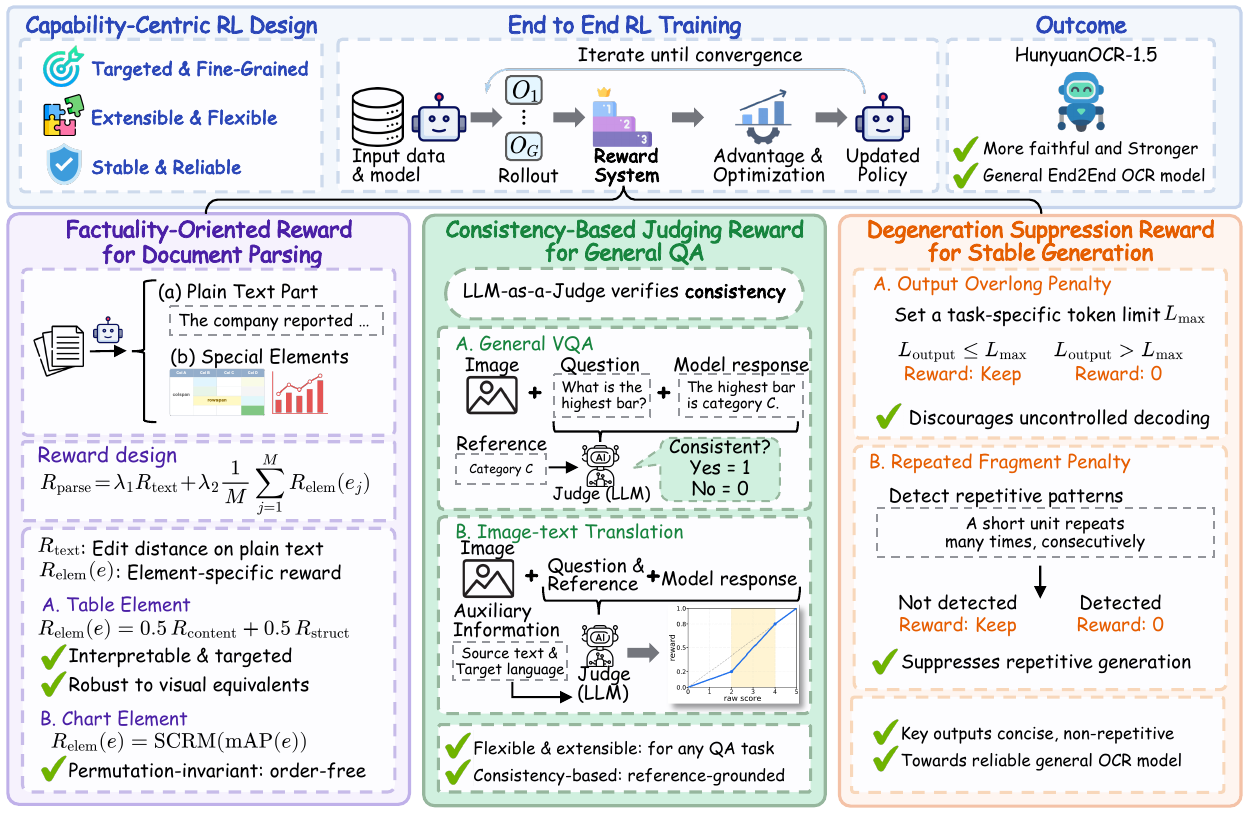}
    \vspace{-1.8em}
    \caption{
        \textbf{Overview of the RL framework.}
        The RL framework that optimizes the general OCR model toward more faithful, stronger, and more comprehensive behavior through three complementary reward components.
    }
    \label{fig:RL_framework}
    \vspace{-10pt}
\end{figure}

Reinforcement learning (RL) has emerged as a powerful paradigm for large language models (LLMs) and multimodal large language models (MLLMs), with success in mathematical reasoning~\cite{DeepSeekMath_2024} and image segmentation~\cite{Seg_Zero_2025}. This is largely attributed to RL's ability to align model outputs with verifiable metrics~\cite{wen2025reinforcement} or human preferences~\cite{Uni_DPO_2026, SENTINEL_2025}. HunyuanOCR~\cite{HunyuanOCR_2025} has already validated this potential in the OCR domain: through high-quality RL data and an ability-adaptive reward design, it achieves stable and effective training, showing that RL can substantially improve lightweight OCR models across diverse tasks.

HunyuanOCR-1.5 pushes this direction further. As shown in~\cref{fig:RL_framework}, we build a reward system tailored to the capabilities an OCR model must acquire, providing fine-grained, discriminative signals. Concretely, it consists of three complementary components: (i) a \textit{capability-routed, structure-aware rule reward} that targets the factual fidelity of document parsing; (ii) a \textit{consistency-based judging reward} for general question answering that flexibly scores arbitrary QA tasks and can be easily extended; and (iii) a \textit{degeneration-suppression reward} that detects overlong and repetitive outputs to keep generation stable. Together, these components drive HunyuanOCR-1.5 toward a more faithful, stronger, and more comprehensive general OCR model.

\subsubsection{Training Strategy}

\mypara{Data curation.}
Following HunyuanOCR~\cite{HunyuanOCR_2025}, we curate the RL data with an emphasis on quality, diversity, and difficulty balance. Starting from the SFT policy, we perform $n=16$ on-policy rollouts for each candidate query and estimate its difficulty based on the rollout outcomes. Queries that are already solved consistently by the policy are discarded, retaining only informative examples with non-trivial reward variance for RL training.

\mypara{Optimization.}
%
%
%
We adopt IcePop~\cite{IcePop_2025}, a GRPO-style policy optimization variant, as our main reinforcement learning framework to mitigate the training--inference mismatch. Let $\pi_{\mathrm{infer}}$ and $\pi_{\mathrm{train}}$ denote the same policy as executed by the inference and training engines. In each iteration, for a query $q$, IcePop samples a group of $G$ responses $\{o_1,o_2,\ldots,o_G\}$ from the old inference policy $\pi_{\mathrm{infer}}(\cdot\mid q;\theta_{\mathrm{old}})$, while the update is computed with $\pi_{\mathrm{train}}(\cdot;\theta)$. To suppress unstable updates from train--inference discrepancies, we compute a token-level calibration ratio between the two policies and only retain tokens whose ratio lies in a prescribed interval. Since our implementation adopts a token-mean loss~\cite{DAPO_2026}, we normalize over all valid tokens that pass the IcePop mask, instead of first averaging each response by its length $|o_{i}|$:

\vspace{-1.5em}
\begingroup
\allowdisplaybreaks
\begin{align}
    \mathcal{J}_{\mathrm{IcePop}}^{\mathrm{tok}}(\theta)
     & =
    \mathbb{E}_{q,\{o_i\}_{i=1}^{G}}
    \left[
        \frac{1}{Z}
        \sum_{i=1}^{G}\sum_{t=1}^{|o_i|}
        a_{i,t}s_{i,t}
        \left(
        \mathcal{L}^{\mathrm{PG}}_{i,t}(\theta)
        - \gamma\,\mathbb{D}_{\mathrm{KL},i,t}
        \right)
        \right],
    \label{eq:icepop-token-mean}
    \\[0.2em]
    \mathcal{L}^{\mathrm{PG}}_{i,t}(\theta)
     & =
    c_{i,t}
    \min\!\left(
    r_{i,t}(\theta) A_{i},
    \operatorname{clip}\!\left(r_{i,t}(\theta),1-\epsilon,1+\epsilon\right) A_i
    \right),
    \label{eq:icepop-pg-loss}
    \\[0.2em]
    Z
     & =
    \sum_{i=1}^{G}\sum_{t=1}^{|o_i|} a_{i,t}s_{i,t},
    \label{eq:icepop-token-mean-normalizer}
    \\[0.2em]
    r_{i,t}(\theta)
     & =
    \frac{
        \pi_{\mathrm{train}}(o_{i,t}\mid q,o_{i,<t};\theta)
    }{
        \pi_{\mathrm{train}}(o_{i,t}\mid q,o_{i,<t};\theta_{\mathrm{old}})
    },
    \label{eq:icepop-ratio-r}
    \\[0.2em]
    c_{i,t}
     & =
    \frac{
        \pi_{\mathrm{train}}(o_{i,t}\mid q,o_{i,<t};\theta_{\mathrm{old}})
    }{
        \pi_{\mathrm{infer}}(o_{i,t}\mid q,o_{i,<t};\theta_{\mathrm{old}})
    },
    \quad
    s_{i,t}=\mathbf{1}\!\left[\alpha_{\mathrm{m}}\le c_{i,t}\le\beta_{\mathrm{m}}\right].
    \label{eq:icepop-mask}
\end{align}
\endgroup
\vspace{-0.5em}

Here $a_{i,t}\in\{0,1\}$ is the valid-token mask, $A_{i}$ is the group-relative advantage of response $o_{i}$, and $\mathbb{D}_{\mathrm{KL},i,t}$ is the token-level KL term. The bounds $\alpha_{\mathrm{m}}$ and $\beta_{\mathrm{m}}$ control the acceptable train--inference ratio region; tokens outside it have $s_{i,t}=0$ and do not contribute to the update. The hyperparameters $\epsilon$ and $\gamma$ control PPO-style clipping and KL strength. If no token in a mini-batch satisfies the IcePop mask, the update is skipped.

\subsubsection{Factuality-Oriented Reward for Document Parsing}
\label{subsubsec:factual_reward}

The first and most fundamental component targets factual fidelity, since a reliable OCR model must faithfully transcribe what is visually present rather than hallucinate plausible content. For text spotting, we follow the rule-based reward of HunyuanOCR~\cite{HunyuanOCR_2025}. Document parsing, however, requires a dedicated design: it converts a document image into a structured representation that may contain plain text, tables, and charts. Tables and charts are structured elements whose correctness is not well captured by the edit-distance criterion used for plain text, so a single text-level metric yields coarse and sometimes misleading signals.

To provide a fine-grained, structure-aware reward, we parse the output and reference into a plain-text part and a set of special elements, each a table or a chart. The parsing reward is then computed as:

\vspace{-5mm}
\begin{equation}
    R_{\text{parse}}
    =
    \lambda_{1}\, R_{\text{text}}
    +
    \lambda_{2}\, \frac{1}{M}\sum_{j=1}^{M} R_{\text{elem}}(e_j)\,,
    \label{eq:parsing_reward}
\end{equation}
\vspace{-3mm}

\noindent
where $R_{\text{text}}$ is the text reward based on normalized edit distance, $\{e_{1}, \dots, e_{M}\}$ are the $M$ special elements parsed from the reference, $R_{\text{elem}}(e_j)$ is the element-specific score defined below, and $\lambda_1,\lambda_2$ balance the two terms. Averaging over the $M$ special elements keeps the reward well scaled.

\mypara{Table.}
Table content is usually expressed in HTML with structure-specific tags for rows, columns, and cell merging, so it requires a dedicated reward rather than plain text matching. TEDS and TEDS-S~\cite{data_pubtabnet} are the standard metrics for table parsing and are natural reward candidates, but both have limitations in an RL setting. As a structural reward, TEDS-S is computed in a rather black-box manner, making it hard for the model to identify which structural part of the output is wrong and to optimize accordingly. As a content reward, TEDS relies on character-level edit distance, which produces inaccurate scores in cases that are visually equivalent but expressed differently, such as mathematical formulas. We therefore improve both terms: we replace the TEDS-S score with a 1D-probe structural reward $R_{\text{struct}}$, and enhance the TEDS-based content reward with an anchor-guided destylization mechanism to obtain $R_{\text{content}}$~\cite{StrucTab_2026}. The score of a table element is

\vspace{-3mm}
\begin{equation}
    R_{\text{elem}}(e)
    =
    0.5\,R_{\text{content}} + 0.5\,R_{\text{struct}}\,.
\end{equation}
\vspace{-3mm}

\mypara{Chart.}
Chart content is typically represented in a Markdown-style table, where the row and column orderings usually do not affect correctness, and even transposing rows and columns expresses the same chart. A reward that is sensitive to such orderings would penalize correct predictions, so charts also require a dedicated design. We first convert both the prediction and the reference into a tabular CSV form and then apply SCRM (Structuring Chart-oriented Representation Metric)~\cite{StructChart_2026} to compute the mean Average Precision (AP) between them, which serves as the score $R_{\text{elem}}(e)$ of the chart element~\cite{ChartArena_2026}. This makes the reward invariant to order permutations while remaining sensitive to the underlying chart semantics.

\subsubsection{Consistency-Based Judging Reward for General QA}
\label{subsubsec:qa_reward}

Beyond factual parsing, we extend the model toward broader capabilities through a consistency-based judging reward for general question answering. Instead of designing a bespoke metric for every task, this component uses an LLM-as-a-judge to verify the consistency between the model response and a high-quality reference. Its key advantage is flexibility and extensibility: it can score arbitrary QA tasks, and can incorporate additional annotation fields to support more fine-grained evaluation of specific downstream tasks.

\mypara{Visual question answering.}
For general VQA, the reward is binary: the judge assigns $1$ if the model's answer is semantically consistent with the reference and $0$ otherwise, focusing on factual correctness while tolerating minor stylistic variations. This provides a clear and robust supervision signal for answer correctness.

\mypara{Translation.}
Translation is a representative case where additional annotation fields enable more precise judging. We provide the judge with auxiliary metadata such as the source-language text and the target-language label, and let it assign a soft score in the range $[0,5]$ based on consistency with the reference translation. The score is normalized to $[0,1]$ via a debiased mapping that expands the resolution of mid-range scores, making the reward more sensitive to subtle quality differences and better able to capture improvements.

\subsubsection{Degeneration Suppression Reward for Stable Generation}
\label{subsubsec:degeneration_reward}

The third component keeps generation stable by explicitly suppressing degenerate outputs. In OCR settings, degeneration typically appears as overlong outputs, repeated fragments, or cyclic generation patterns when the model encounters uncertain or out-of-distribution inputs, which severely undermines reliability in real-world deployment. To address this, we introduce two complementary penalties during training.

\mypara{Overlong output penalty.}
For each task, we set an appropriate upper bound on the output length according to its expected format and maximal valid response length. Since excessive length is a common indicator of repeated or drifting generation, we directly assign a reward of zero to rollouts that exceed the predefined token limit, discouraging uncontrolled decoding and encouraging concise, task-aligned outputs.

\mypara{Repeated fragment detection and penalty.}
For rollouts within the length limit, we further detect repetitive patterns, which are a common failure mode of long OCR outputs, where a short unit of at most \textit{max\_unit} tokens repeats at least \textit{min\_repeats} times consecutively at the end of the sequence. Such rollouts also receive a zero reward. This penalty suppresses repetitive generation and improves the stability of model outputs.

Overall, the three reward components form a coherent and layered design: the factuality-oriented reward secures faithful parsing of structured documents, the consistency-based judging reward extends the model toward general and open-ended capabilities, and the degeneration-suppression reward stabilizes generation throughout. Together, they provide fine-grained, discriminative, and extensible optimization signals that drive HunyuanOCR-1.5 toward a more faithful, stronger, and more comprehensive general OCR model.

\section{Evaluation Tree}
\label{sec:evaluation_tree}

Rather than relying on a single or isolated benchmark, HunyuanOCR-1.5 is evaluated through a capability-oriented \textbf{OCR evaluation tree}. This evaluation design mainly answers two questions: (a) whether the core OCR capabilities established in HunyuanOCR-1.0~\cite{HunyuanOCR_2025}, such as document parsing, general OCR-aware QA, text spotting, and information extraction, are further strengthened; and (b) whether newly introduced capabilities, including low-resource languages, ancient scripts, multi-image understanding, and faithful seen-text parsing, are effectively incorporated into the model boundary.

Along an orthogonal dimension, all evaluation sources are categorized by their origin into \textbf{open-source} and \textbf{in-house} benchmarks. Only Spotting, IE, and Video Subtitle Extraction rely on in-house and real-world production benchmarks, while all remaining dimensions are evaluated on open-source benchmarks.

\subsection{OCR Capability Evaluation Tree}

The evaluation tree is organized into three groups by evaluation purpose. The first group verifies that the fundamental OCR capabilities inherited from HunyuanOCR-1.0 are preserved and further strengthened. The second group examines whether the newly extended boundary capabilities, such as long-tail languages, ancient scripts, multi-image understanding, and structured element parsing, are effectively incorporated into the model. The third group targets output reliability, focusing on the model's seen-text preservation ability and its hallucination risk in long OCR sequences. The following subsection details the specific dimensions within each group and their corresponding benchmarks.

\subsection{Evaluation Dimensions and Benchmark Mapping}

We first organize the evaluation dimensions into several capability groups and then provide the benchmark mapping for each group. This organization highlights what types of OCR capabilities are evaluated before specifying how each capability is measured.

\begin{itemize}[
            label=\raisebox{0.5ex}{\tiny$\bullet$},
            leftmargin=1em,
            itemsep=2pt,
            parsep=0pt,
            topsep=0pt,
            partopsep=0pt
      ]
      \item \textbf{Basic OCR and document understanding.}
            This group evaluates the fundamental capabilities of OCR model.
            \textit{End-to-end document parsing} is evaluated by OmniDocBench~\cite{OmniDocBench_2025}, following the latest official evaluation protocol, which measures structured parsing of mainstream printed and scanned documents, including body text, tables, formulas, and reading order.
            \textit{General OCR capability} is evaluated by OCRBench~\cite{OCRBench_2024}, focusing on OCR-aware QA across scene text recognition, document question answering, information extraction, formula recognition, and chart understanding.
            \textit{Text spotting} is evaluated by an in-house Spotting Benchmark, which measures text localization and recognition across document images, scene text, artistic text, handwriting, advertisements, cards and receipts, screenshots, street views, and video frames.

      \item \textbf{Long-tail capability expansion.}
            This group evaluates whether HunyuanOCR-1.5 effectively extends its capability boundary to long-tail languages and scripts.
            \textit{Low-resource multilingual parsing} is evaluated by MORE~\cite{MORE_2026}, which covers parsing across 149 languages and focuses on low-resource languages and rare writing systems.
            \textit{Ancient script recognition} is evaluated by Chronicles-OCR~\cite{Chronicles_OCR_2026}, which measures recognition ability on the seven historical forms of Chinese characters, historical documents, and ancient-script images.

      \item \textbf{Structured visual element parsing.}
            This group evaluates structured visual element parsing beyond plain text recognition.
            \textit{Table parsing} is evaluated by TableVerse-5K~\cite{StrucTab_2026}, which measures table structure and content reconstruction.
            \textit{Chart parsing} is evaluated by ChartArena~\cite{ChartArena_2026}, which evaluates chart text, structure, and semantic parsing.

      \item \textbf{Cross-page and cross-lingual understanding.}
            This group evaluates capabilities that go beyond single-page OCR.
            \textit{Multi-image QA} is evaluated by DUDE~\cite{DUDE_2023}, which measures multi-page document understanding, cross-page information retrieval, multi-image content comparison, and evidence aggregation.
            \textit{Text image translation} is evaluated by DoTA~\cite{DOTA_2024} and MMTIT~\cite{MMTIT_Bench_2026}. DoTA focuses on English document image translation into Chinese, while MMTIT evaluates multilingual text image translation from 14 non-Chinese and non-English languages into Chinese or English across multiple scenarios.

      \item \textbf{Application-oriented and reliability evaluation.}
            This group evaluates practical OCR applications and output faithfulness.
            \textit{Information extraction} is evaluated by an in-house IE Benchmark~\cite{HunyuanOCR_2025}, which targets structured field extraction from cards, receipts, and forms, requiring the model to transcribe target fields and return them as structured key-value outputs rather than free-form text.
            \textit{Video subtitle extraction} is evaluated by an in-house Video Subtitle Extraction Benchmark~\cite{HunyuanOCR_2025}, which measures subtitle recognition from video frames together with the temporal consistency of recognized text and the robustness to dynamic backgrounds and compression noise.
            \textit{Document hallucination} is evaluated by \textbf{CHAOS-Bench}, short for \textbf{Comprehensive Hallucination Assessment for OCR Sequences}. CHAOS-Bench evaluates faithfulness with a controlled WYSIWYG protocol. For each document page, we modify one character in 2 to 3 selected words in the rendered image, turning them into meaningless perturbed words. Degenerate edits and modified strings that remain dictionary-valid words are removed. Given the set of perturbed words $\mathcal{P}_i$ on page $i$ and the model output $O_i$, a hit $\mathbf{1}_{\mathrm{hit}}(w,O_i)$ is counted when word $w \in \mathcal{P}_i$ appears in $O_i$ as a case-insensitive whole-word match. The page-level recall is computed as:
            \begin{equation}
                  R_{i} = \frac{1}{|\mathcal{P}_i|}\sum_{w\in\mathcal{P}_i}\mathbf{1}_{\mathrm{hit}}(w,O_i).
            \end{equation}
            The final score is the page-averaged recall over all $N$ pages:
            \begin{equation}
                  \mathrm{Recall}_{\mathrm{page}} = \frac{1}{N}\sum_{i=1}^{N}R_i.
            \end{equation}
            This metric directly measures whether the model preserves visually observed words when visual evidence conflicts with language priors.
\end{itemize}

Based on this evaluation tree, the experimental results in the next section are reported from two complementary perspectives. We first highlight the boundary capability evaluations that are newly introduced or strengthened in HunyuanOCR-1.5, including Chronicles-OCR, ChartArena, TableVerse-5K, DUDE, MORE, and CHAOS-Bench. We then analyze the changes to the existing evaluation dimensions inherited from HunyuanOCR-1.0, including OmniDocBench, Spotting, text-image translation, IE, Video Subtitle Extraction, and OCRBench. This organization keeps the capability taxonomy in the evaluation tree while focusing the results discussion on what is newly monitored and what is preserved or improved from the previous version.

\makeatletter
\newcommand{\ghopensource}[1]{
    \def\gh@arg{#1}
    \ifx\gh@arg\@empty
        Open-source
    \else
        \href{#1}{\raisebox{-0.15\height}{\includegraphics[height=8pt]{figure/logo/github.png}} Open-source}
    \fi
}
\makeatother

\begin{table*}[t]
    \centering
    \small
    \caption{
        Grouped benchmark mapping for the OCR capability evaluation tree.
    }
    \vspace{-0.5em}
    \label{tab:benchmark_mapping}
    \setlength{\tabcolsep}{4pt}
    \begin{tabularx}{\textwidth}{p{0.25\textwidth} p{0.3\textwidth} p{0.25\textwidth} p{0.2\textwidth}}
        \toprule
        \textbf{Capability Group}     &
        \textbf{Evaluation Dimension} &
        \textbf{Benchmark}            &
        \textbf{Source}
        \\
        \midrule
        \multirow{3}{=}{Basic OCR and \nextline doc understanding}
                                      & End-to-end document parsing & OmniDocBench~\cite{OmniDocBench_2025}     & \ghopensource{https://github.com/opendatalab/OmniDocBench}                                    \\
                                      & General OCR-aware QA        & OCRBench~\cite{OCRBench_2024}             & \ghopensource{https://github.com/Yuliang-Liu/MultimodalOCR}                                   \\
                                      & Text spotting               & Spotting Benchmark                        & In-house                                                                                      \\
        \midrule
        \multirow{2}{=}{Long-tail ability expansion}
                                      & Multilingual parsing        & MORE~\cite{MORE_2026}                     & \ghopensource{https://github.com/zimoqingfeng/MORE}                                           \\
                                      & Ancient-script recognition  & Chronicles-OCR~\cite{Chronicles_OCR_2026} & \ghopensource{https://github.com/VirtualLUOUCAS/Chronicles-OCR}                               \\
        \midrule
        \multirow{2}{=}{Structured visual \nextline element parsing}
                                      & Table parsing               & TableVerse-5K~\cite{StrucTab_2026}        & \ghopensource{https://github.com/VirtualLUOUCAS/StrucTab}                                     \\
                                      & Chart parsing               & ChartArena~\cite{ChartArena_2026}         & \ghopensource{https://github.com/pspdada/ChartArena}                                          \\
        \midrule
        \multirow{3}{=}{Cross-page and \nextline cross-lingual understanding}
                                      & Multi-image QA              & DUDE~\cite{DUDE_2023}                     & \ghopensource{https://github.com/duchallenge-team/dude}                                       \\
                                      & Text image translation      & DoTA~\cite{DOTA_2024}                     & \ghopensource{https://github.com/liangyupu/DIMTDA}                                            \\
                                      & Text image translation      & MMTIT~\cite{MMTIT_Bench_2026}             & \ghopensource{https://github.com/VirtualLUOUCAS/MMTIT_Bench}                                  \\
        \midrule
        \multirow{3}{=}{Practical applications \nextline and reliability}
                                      & Information extraction      & IE Benchmark                              & In-house                                                                                      \\
                                      & Video subtitle extraction   & VSE Benchmark                             & In-house                                                                                      \\
                                      & Document hallucination      & CHAOS-Bench                               & \ghopensource{https://github.com/Tencent-Hunyuan/HunyuanOCR/tree/main/benchmarks/CHAOS-Bench} \\
        \bottomrule
    \end{tabularx}
\end{table*}
\section{Experimental Results}
\label{sec:experimental_results}

\subsection{Inference Speed with DFlash}
\label{sec:exp_dflash_speed}

We evaluate the inference speed of HunyuanOCR-1.5 with standard autoregressive (AR) decoding and DFlash-accelerated decoding on OmniDocBench~\cite{OmniDocBench_2025}. Unless otherwise specified, we report per-sample metrics:
\begin{equation}
    \mathrm{Latency} = \frac{1}{N}\sum_{i=1}^{N} t_i\,,\quad
    \mathrm{Token/s} = \frac{\sum_{i=1}^{N} c_i}{\sum_{i=1}^{N} t_i}\,,\quad
    \mathrm{Page/s} = \frac{N}{\sum_{i=1}^{N} t_i}\,,
\end{equation}
where $t_{i}$ and $c_{i}$ denote the latency and generated tokens of the $i$-th sample. The speedup is computed by comparing DFlash with AR under the same metric.

\mypara{Overall speed.}
We first compare AR decoding and DFlash decoding under single-request inference. As shown in~\cref{tab:dflash_overall_speed}, DFlash significantly accelerates HunyuanOCR-1.5 under both Transformers~\cite{Transformers_2020} and vLLM~\cite{vLLM_2023}. In vLLM, DFlash reduces the average latency from 3.032s to 1.408s, improves throughput from 466.9 token/s to 1002.3 token/s, and achieves a $2.14\times$ speedup. The gain is even larger under Transformers, whose AR baseline is closer to naive token-by-token decoding and therefore benefits more from speculative decoding. These results show that DFlash effectively reduces the decoding latency of long OCR outputs while preserving the original end-to-end generation paradigm.

\begin{table*}[t]
    \centering
    \caption{
        \textbf{Overall inference speed comparison.}
        Comparison between AR and DFlash decoding on OmniDocBench under batch size 1. The vLLM results are aligned with the latest 930-sample SOTA comparison.
    }
    \label{tab:dflash_overall_speed}
    \vspace{-0.4em}
    \renewcommand{\arraystretch}{1.1}
    \setlength{\tabcolsep}{8pt}
    \resizebox{\textwidth}{!}{
        \begin{tabular}{lccc HHH cc}
            \toprule
            \multirow{2}{*}[-1.0mm]
            {\textbf{Framework}}       &
            \multicolumn{3}{c}
            {\textbf{AR Decoding}}     &
            \multicolumn{3}{H}
            {\textbf{DFlash Decoding}} &
            \multirow{2}{*}[-1.0mm]
            {\textbf{Speedup}}         &
            \multirow{2}{*}[-1.0mm]
            {\makecell{\textbf{Effective} \nextline \textbf{Acc. Length}}}
            \\
            \cmidrule(lr){2-4}
            \cmidrule(lr){5-7}
                                       & Latency (s) $\downarrow$ & TPS   & Page/s & Latency (s) $\downarrow$ & TPS    & Page/s &              &      \\
            \midrule
            Transformers               & 34.850                   & 40.9  & 0.029  & 5.474                    & 245.7  & 0.183  & 6.37$\times$ & 8.89 \\
            vLLM                       & 3.032                    & 466.9 & 0.330  & 1.408                    & 1002.3 & 0.706  & 2.14$\times$ & 8.36 \\
            \bottomrule
        \end{tabular}
    }
\end{table*}

\mypara{Comparison with SOTA OCR systems.}
We further compare HunyuanOCR-1.5 with DFlash against representative OCR systems, including two-stage pipeline methods and end-to-end OCR VLMs. The evaluation is conducted on the same OmniDocBench test set under single-request inference, where each system is assigned one accelerator instance with comparable compute capacity. For two-stage systems, GLM-OCR~\cite{GLM_OCR_2026} and PaddleOCR-VL-1.6~\cite{PaddleOCR_VL_1_6_2026}, we report the full page-level pipeline latency, including layout analysis, region-level OCR/VLM inference, and result merging. Since different systems use different tokenizers, prompts, and output formats, cross-model token/s is not directly comparable; we mainly compare average latency and page throughput.

As shown in~\cref{tab:sota_speed_comparison}, HunyuanOCR-1.5 with DFlash achieves the fastest end-to-end inference speed among all evaluated systems, reaching 1.408s per page and 0.706 page/s. It is about $1.17\times$ faster than GLM-OCR and $1.24\times$ faster than PaddleOCR-VL-1.6, while keeping a unified end-to-end OCR VLM formulation without explicit layout decomposition or region-wise cascaded inference. Compared with other OCR VLMs, including Unlimited-OCR~\cite{Unlimited_OCR_2026}, DeepSeek-OCR 2~\cite{DeepSeek_OCR_2_2026}, and dots.ocr~\cite{dots_ocr_2025}, HunyuanOCR-1.5 with DFlash reduces average latency by $2.60\times$, $3.88\times$, and $5.08\times$, respectively.

\begin{table*}[t]
    \centering
    \caption{
        \tbf{End-to-end speed comparison with representative OCR systems.}
        Evaluated on the OmniDocBench test set. GLM-OCR and PaddleOCR-VL-1.6 are two-stage multi-model pipeline methods, while others are single-model end-to-end VLMs. Speedup is measured against the HunyuanOCR-1.5 AR setting.
    }
    \vspace{-0.4em}
    \label{tab:sota_speed_comparison}
    \renewcommand{\arraystretch}{1.2}
    \resizebox{\textwidth}{!}{
        \begin{tabular}{lllccc}
            \toprule
            \tbf{Model}                                                & \tbf{Paradigm} & \tbf{Inference Method} & \tbf{Avg. Latency / Page (s)} $\downarrow$ & \tbf{Page/s} $\uparrow$ & \tbf{Speedup}      \\
            \midrule
            \logo{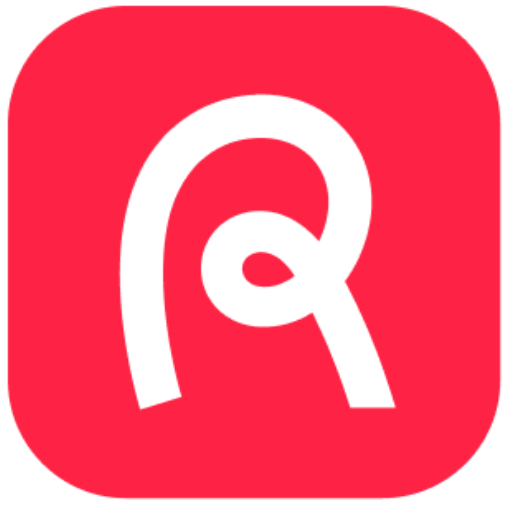}dots.ocr~\cite{dots_ocr_2025}            & End-to-end     & Auto-regressive        & 7.154                                      & 0.136                   & 0.41$\times$       \\
            \logo{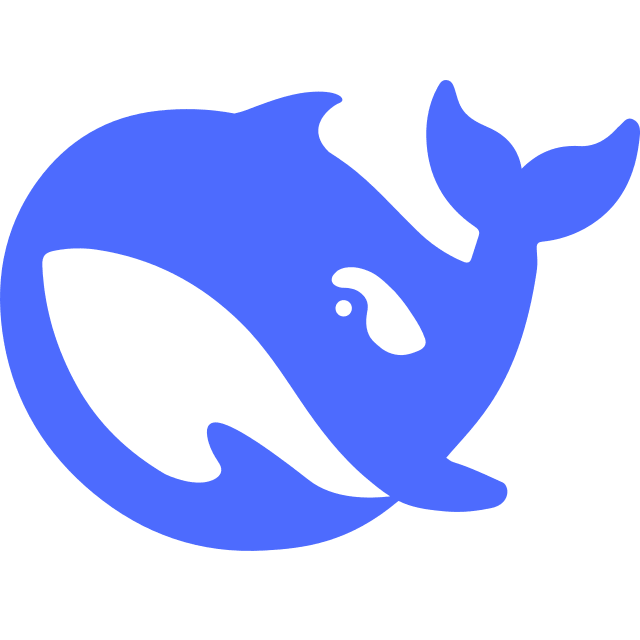}DeepSeek-OCR 2~\cite{DeepSeek_OCR_2_2026}   & End-to-end     & Auto-regressive        & 5.460                                      & 0.179                   & 0.54$\times$       \\
            \logo{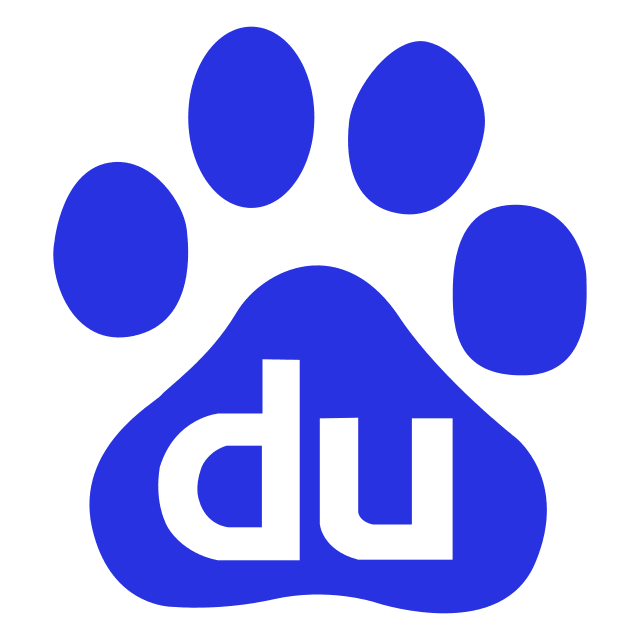}Unlimited-OCR~\cite{Unlimited_OCR_2026}        & End-to-end     & Auto-regressive        & 3.659                                      & 0.255                   & 0.77$\times$       \\
            \hybr
            \logo{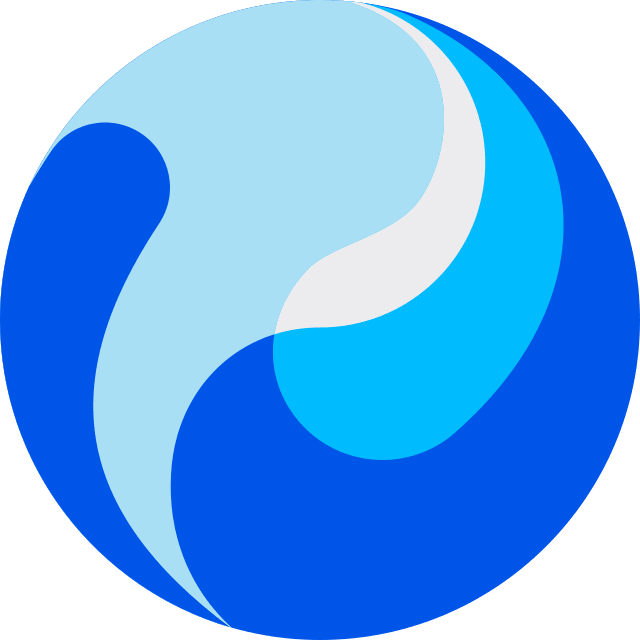}HunyuanOCR-1.5                               & End-to-end     & Auto-regressive        & 3.032                                      & 0.330                   & 1.00$\times$       \\
            \logo{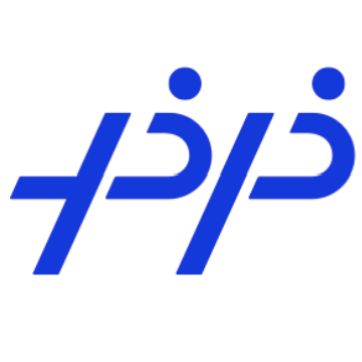}PaddleOCR-VL-1.6~\cite{PaddleOCR_VL_1_6_2026} & Two-stage      & Cascade                & 1.744                                      & 0.562                   & 1.71$\times$       \\
            \logo{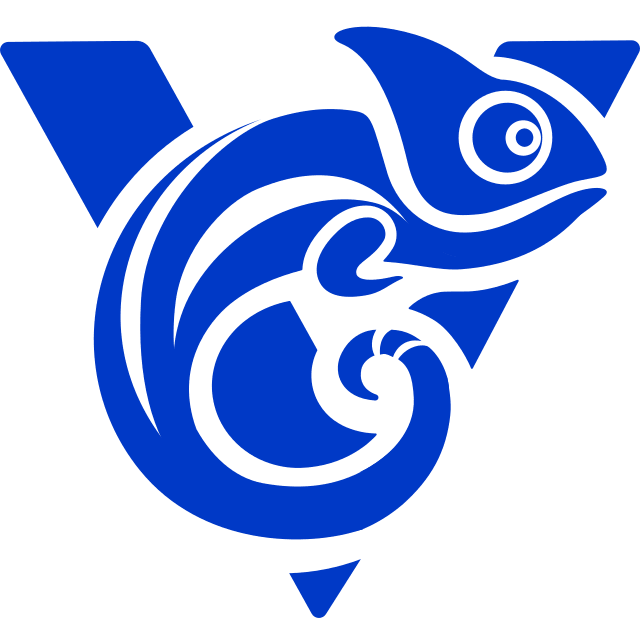}GLM-OCR~\cite{GLM_OCR_2026}                     & Two-stage      & Cascade                & 1.649                                      & 0.604                   & 1.83$\times$       \\
            \hybr
            \logo{hunyuan}HunyuanOCR-1.5                               & End-to-end     & DFlash                 & \tbf{1.408}                                & \tbf{0.706}             & \tbf{2.14$\times$} \\
            \bottomrule
        \end{tabular}
    }
\end{table*}

\mypara{Speedup versus output length.}
We analyze DFlash speedup across different output length ranges in~\cref{tab:dflash_length_speed}. The speedup consistently increases as the output sequence becomes longer. In vLLM, DFlash improves from $1.31\times$ on 0--256 token outputs to $2.30\times$ on 2048+ token outputs; in Transformers, the speedup increases from $4.56\times$ to $6.67\times$. This matches the nature of speculative decoding: longer outputs require more decoding steps, so each parallel draft-and-verify iteration amortizes more target-model forward passes. Short outputs are instead dominated by prefill and fixed overheads, limiting the attainable speedup.

\begin{table*}[t]
    \centering
    \caption{
        \textbf{Inference speed comparison across different output length ranges.}
        Output length is measured by AR completion tokens. Effective acceptance length denotes the average number of tokens advanced per speculative decoding step, including the bonus token.
    }
    \vspace{-0.4em}
    \label{tab:dflash_length_speed}
    \renewcommand{\arraystretch}{1.1}
    \resizebox{\textwidth}{!}{
        \begin{tabular}{ll ccc H H H cc}
            \toprule
            \multirow{2}{*}[-1.0mm]
            {\textbf{Framework}}                          &
            \multirow{2}{*}[-1.0mm]
            {\textbf{\makecell{Output \nextline Length}}} &
            \multicolumn{3}{c}
            {\textbf{AR Decoding}}                        &
            \multicolumn{3}{H}
            {\textbf{DFlash Decoding}}                    &
            \multirow{2}{*}[-1.0mm]
            {\textbf{Speedup}}                            &
            \multirow{2}{*}[-1.0mm]
            {\textbf{\makecell{Effective \nextline Acc. Length}}}
            \\
            \cmidrule(lr){3-5}
            \cmidrule(lr){6-8}
                                                          &                   & Latency (s) $\downarrow$ & TPS   & Page/s & Latency (s) $\downarrow$ & TPS    & Page/s &              &      \\
            \midrule
            \multirow{5}{*}{Transformers}
                                                          & [0, 256]          & 7.298                    & 28.1  & 0.137  & 1.602                    & 126.1  & 0.624  & 4.56$\times$ & 8.43 \\
                                                          & (256, 512]        & 11.014                   & 35.7  & 0.091  & 2.069                    & 191.6  & 0.483  & 5.32$\times$ & 8.29 \\
                                                          & (512, 1024]       & 19.932                   & 38.3  & 0.050  & 3.448                    & 231.3  & 0.290  & 5.78$\times$ & 9.15 \\
                                                          & (1024, 2048]      & 34.905                   & 40.8  & 0.029  & 5.294                    & 276.8  & 0.189  & 6.59$\times$ & 9.26 \\
                                                          & $(2048, +\infty)$ & 93.756                   & 42.9  & 0.011  & 14.054                   & 239.2  & 0.071  & 6.67$\times$ & 8.34 \\
            \midrule
            \multirow{5}{*}{vLLM}
                                                          & [0, 256]          & 0.950                    & 217.9 & 1.052  & 0.723                    & 286.4  & 1.383  & 1.31$\times$ & 9.23 \\
                                                          & (256, 512]        & 1.156                    & 341.8 & 0.865  & 0.746                    & 529.5  & 1.340  & 1.55$\times$ & 8.39 \\
                                                          & (512, 1024]       & 1.926                    & 395.9 & 0.519  & 1.086                    & 702.4  & 0.921  & 1.77$\times$ & 9.38 \\
                                                          & (1024, 2048]      & 3.071                    & 466.5 & 0.326  & 1.435                    & 998.6  & 0.697  & 2.14$\times$ & 9.50 \\
                                                          & $(2048, +\infty)$ & 6.660                    & 514.3 & 0.150  & 2.901                    & 1183.0 & 0.345  & 2.30$\times$ & 8.65 \\
            \bottomrule
        \end{tabular}
    }
\end{table*}

\mypara{Speedup by content type.}
We further categorize pages into text, formula, and table pages according to the official OmniDocBench layout annotations~\cite{OmniDocBench_2025}. As shown in~\cref{tab:dflash_content_speed}, both Transformers and vLLM show the same trend: table pages obtain the largest speedup, followed by formula pages and text pages. This is because table outputs usually contain highly regular HTML structures, making future tokens easier to predict and yielding longer effective accepted prefixes.

\begin{table*}[t]
    \centering
    \caption{
        \textbf{Inference speed comparison across different content types.}
        We categorize the OmniDocBench pages into text, formula, and table types, and report the speed on each. The effective acceptance length denotes the average number of tokens advanced per speculative decoding step, including the bonus token.
    }
    \vspace{-0.4em}
    \label{tab:dflash_content_speed}
    \renewcommand{\arraystretch}{1.2}
    \resizebox{\textwidth}{!}{
        \begin{tabular}{ll ccc HHH cc}
            \toprule
            \multirow{2}{*}[-1.0mm]
            {\textbf{Framework}}                         &
            \multirow{2}{*}[-1.0mm]
            {\textbf{\makecell{Content \nextline Type}}} &
            \multicolumn{3}{c}
            {\textbf{AR Decoding}}                       &
            \multicolumn{3}{H}
            {\textbf{DFlash Decoding}}                   &
            \multirow{2}{*}[-1.0mm]
            {\textbf{Speedup}}                           &
            \multirow{2}{*}[-1.0mm]
            {\textbf{\makecell{Effective \nextline Acc. Length}}}
            \\
            \cmidrule(lr){3-5}
            \cmidrule(lr){6-8}
                                                         &         & Latency (s) $\downarrow$ & TPS   & Page/s & Latency (s) $\downarrow$ & TPS    & Page/s &              &       \\
            \midrule
            \multirow{3}{*}{Transformers}
                                                         & Text    & 36.357                   & 40.6  & 0.028  & 6.455                    & 206.8  & 0.155  & 5.63$\times$ & 7.68  \\
                                                         & Formula & 36.753                   & 41.3  & 0.027  & 5.987                    & 237.8  & 0.167  & 6.14$\times$ & 8.59  \\
                                                         & Table   & 32.253                   & 41.0  & 0.031  & 4.129                    & 319.3  & 0.242  & 7.81$\times$ & 10.40 \\
            \midrule
            \multirow{3}{*}{vLLM}
                                                         & Text    & 3.034                    & 450.8 & 0.330  & 1.675                    & 818.4  & 0.597  & 1.81$\times$ & 8.02  \\
                                                         & Formula & 2.626                    & 464.7 & 0.381  & 1.277                    & 955.4  & 0.783  & 2.06$\times$ & 9.04  \\
                                                         & Table   & 2.881                    & 465.1 & 0.347  & 1.207                    & 1110.3 & 0.829  & 2.39$\times$ & 10.45 \\
            \bottomrule
        \end{tabular}
    }
    \vspace{-0.8em}
\end{table*}

\mypara{Throughput under concurrency.}
Finally, we evaluate DFlash under different vLLM concurrency levels. As shown in~\cref{tab:dflash_concurrency_speed}, system throughput increases as concurrency grows, indicating improved GPU utilization from continuous batching. DFlash maintains more than $1.8\times$ speedup from concurrency 1 to 32, with the highest speedup of $2.26\times$ at concurrency 4. The speedup gradually decreases at higher concurrency because the GPU becomes increasingly saturated, leaving less idle compute for speculative decoding.

\begin{table*}[t]
    \centering
    \caption{
        \textbf{vLLM throughput comparison under different concurrency levels.}
        We report the throughput of AR and DFlash decoding as the concurrency level $c$ increases. The $c=1$ row is aligned with the latest 930-sample SOTA speed comparison, while higher-concurrency rows follow the concurrency sweep results.
    }
    \vspace{-0.4em}
    \label{tab:dflash_concurrency_speed}
    \renewcommand{\arraystretch}{1.05}
    \setlength{\tabcolsep}{14pt}
    \resizebox{\textwidth}{!}{
        \begin{tabular}{l ccc HHH c}
            \toprule
            \multirow{2}{*}[-1.0mm]
            {\textbf{Concurrency}}     &
            \multicolumn{3}{c}
            {\textbf{AR Decoding}}     &
            \multicolumn{3}{H}
            {\textbf{DFlash Decoding}} &
            \multirow{2}{*}[-1.0mm]
            {\textbf{Speedup}}
            \\
            \cmidrule(lr){2-4}
            \cmidrule(lr){5-7}
                                       & Latency (s) $\downarrow$ & TPS    & Page/s & Latency (s) $\downarrow$ & TPS    & Page/s &              \\
            \midrule
            $c=1$                      & 3.032                    & 466.9  & 0.330  & 1.408                    & 1002.3 & 0.706  & 2.14$\times$ \\
            $c=2$                      & 3.761 / 2                & 707.4  & 0.532  & 1.785 / 2                & 1493.5 & 1.121  & 2.11$\times$ \\
            $c=4$                      & 5.915 / 4                & 900.0  & 0.676  & 2.615 / 4                & 2039.7 & 1.529  & 2.26$\times$ \\
            $c=6$                      & 7.625 / 6                & 1047.5 & 0.787  & 3.526 / 6                & 2262.8 & 1.702  & 2.16$\times$ \\
            $c=8$                      & 9.433 / 8                & 1127.6 & 0.848  & 4.452 / 8                & 2390.7 & 1.797  & 2.12$\times$ \\
            $c=16$                     & 15.657 / 16              & 1360.9 & 1.022  & 8.395 / 16               & 2539.8 & 1.906  & 1.87$\times$ \\
            $c=32$                     & 29.138 / 32              & 1462.6 & 1.098  & 16.162 / 32              & 2633.9 & 1.980  & 1.80$\times$ \\
            \bottomrule
        \end{tabular}
    }
    \vspace{-0.8em}
\end{table*}

\subsection{Boundary Capability Evaluation}
\label{subsec:boundary_capability}

We first focus on the boundary capabilities of HunyuanOCR-1.5. Here, boundary capabilities include both newly introduced task abilities, such as ancient-script parsing, document-level multi-image QA, and output faithfulness evaluation, and previously supported abilities that are further monitored and strengthened with more fine-grained benchmarks, such as complex table parsing and structured chart parsing. Through Chronicles-OCR, ChartArena, TableVerse-5K, DUDE, MORE, and CHAOS-Bench, this section characterizes the capability boundary of HunyuanOCR-1.5 from long-tail scripts, complex structures, multi-image understanding, and reliability perspectives.

\begin{table}[t]
    \centering
    \caption{
        \textbf{Comparison of ancient-script OCR results on Chronicles-OCR.}
        We report the average Parsing scores on archaic scripts (Oracle Bone, Bronze, Seal) and mature scripts (Clerical, Regular, Running, Cursive).
    }
    \setlength{\tabcolsep}{12pt}
    \resizebox{\columnwidth}{!}{
        \begin{tabular}{cll ccc}
            \toprule
            \textbf{\makecell{Model Type}}          &
            \textbf{Model}                          &
            \textbf{Size}                           &
            \textbf{Think-mode}                     &
            \textbf{Archaic Average}                &
            \textbf{Mature Average}
            \\
            \midrule
            \multirow{11}{*}[-0.8em]
            {\makecell{Open-source \nextline General \nextline VLMs}}
                                                    & \logo{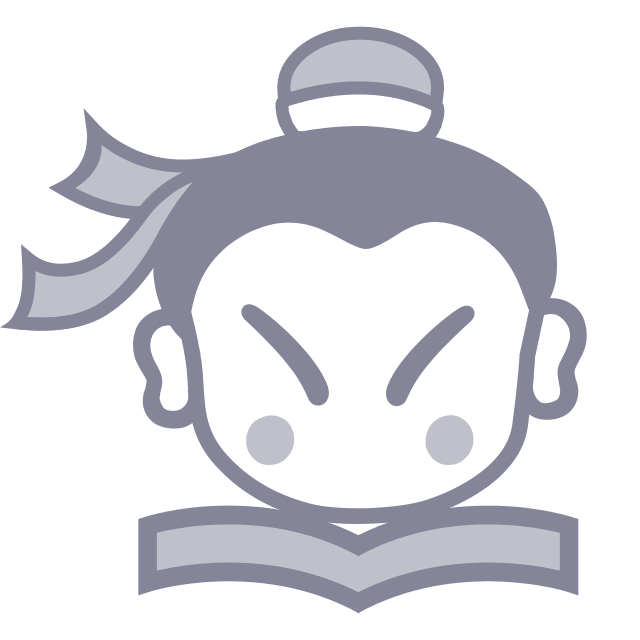}InternVL3.5-8B~\cite{InternVL3_5_2025}        & 8B        &        & 0.07          & 0.39          \\
                                                    & \logo{intern}InternVL3.5-A28B~\cite{InternVL3_5_2025}      & 241B-A28B &        & 0.13          & 0.56          \\
                                                    & \logo{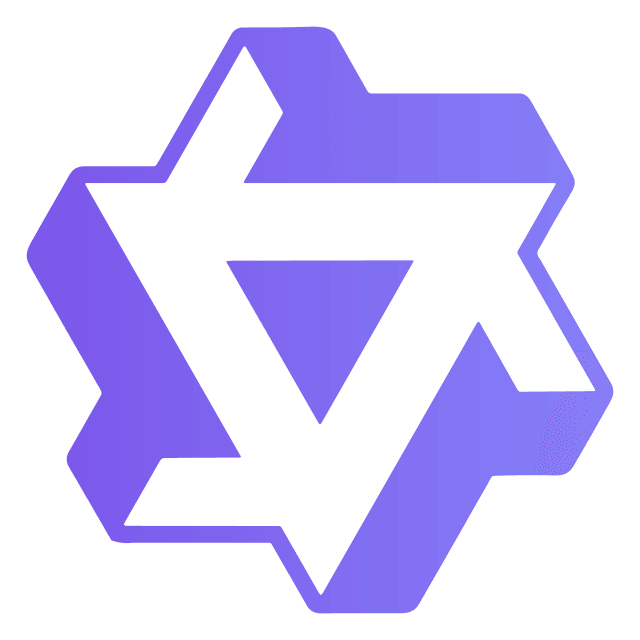}Qwen3-VL-8B~\cite{Qwen3_VL_2025}                & 8B        &        & 0.18          & 0.65          \\
                                                    & \logo{qwen}Qwen3-VL-A22B~\cite{Qwen3_VL_2025}              & 235B-A22B &        & 0.19          & 0.66          \\
                                                    & \logo{qwen}Qwen3.5-9B~\cite{Qwen3_5_2026}                  & 9B        &        & 0.09          & 0.60          \\
                                                    & \logo{qwen}Qwen3.5-A17B~\cite{Qwen3_5_2026}                & 397B-A17B &        & 0.22          & \ul{0.73}     \\
                                                    & \logo{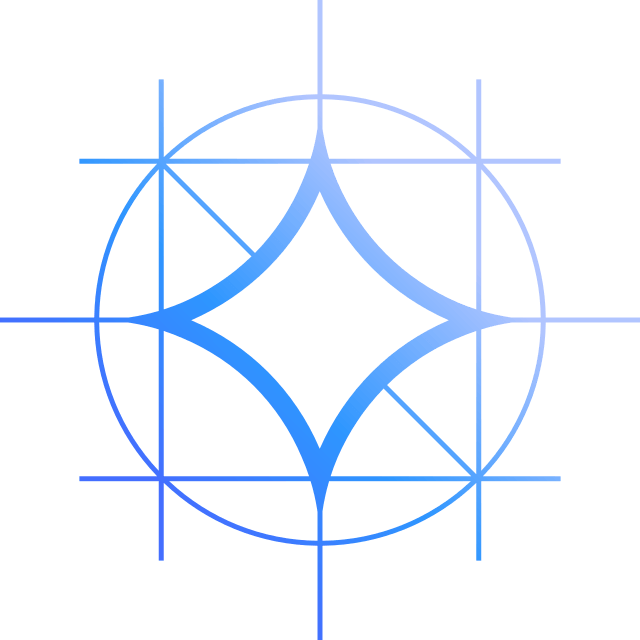}Gemma 4 31B it~\cite{Gemma_4_2026}               & 31B       &        & 0.04          & 0.35          \\
                                                    & \logo{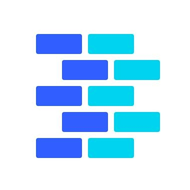}MiniCPM-V 4.5~\cite{MiniCPM_V_4_5_2025}      & 8B        & \cmark & 0.03          & 0.40          \\
                                                    & \logo{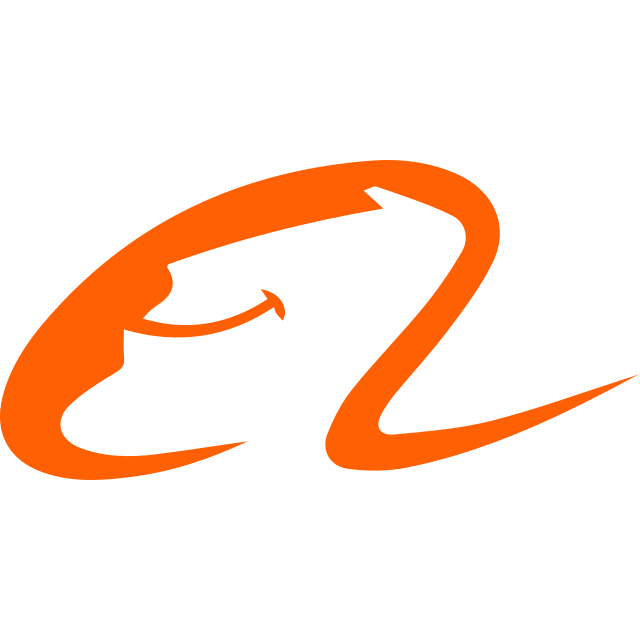}Ovis2.6-30B-A3B~\cite{Ovis2_5_2025}          & 30B-A3B   & \cmark & 0.11          & 0.51          \\
                                                    & \logo{glmv}GLM-4.5V~\cite{GLM_4_5V_2025}                   & 108B-A12B & \cmark & 0.06          & 0.43          \\
                                                    & \logo{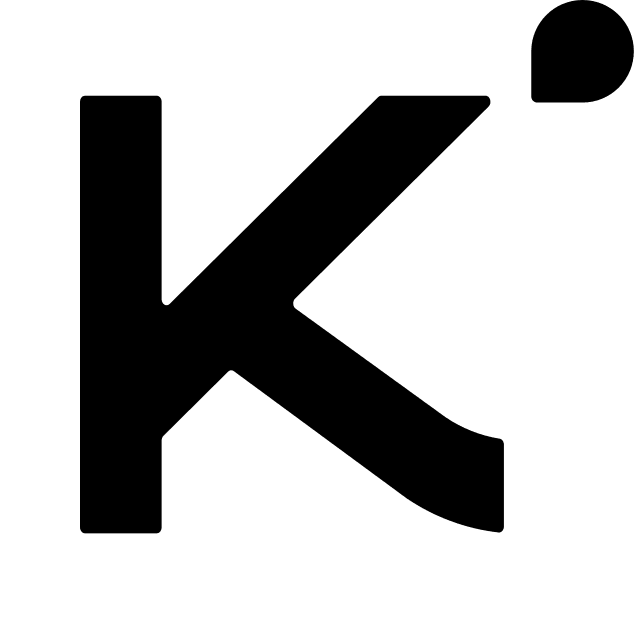}Kimi K2.5~\cite{Kimi_K2_5_2026}                 & 1T        &        & \ul{0.28}     & 0.71          \\
            \cmidrule(lr){1-6}
            \multirow{8}{*}[-0.6em]
            {\makecell{Proprietary \nextline General \nextline VLMs}}
                                                    & \logo{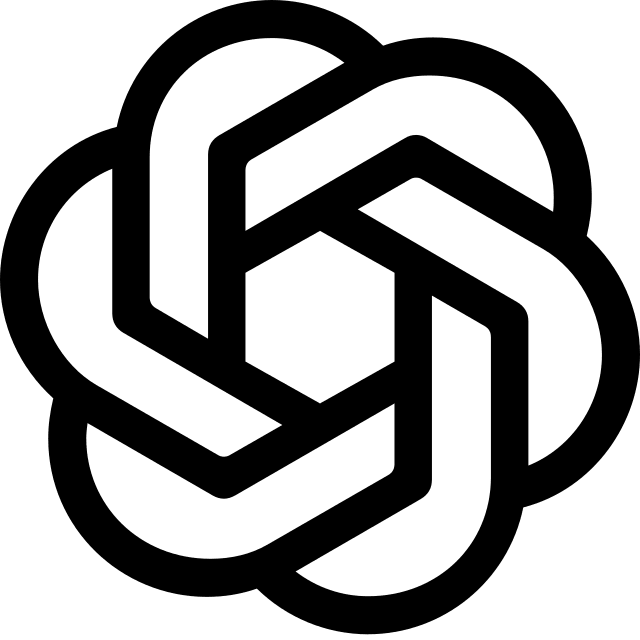}GPT-5~\cite{GPT_5_2025}                       & -         &        & 0.06          & 0.41          \\
                                                    & \logo{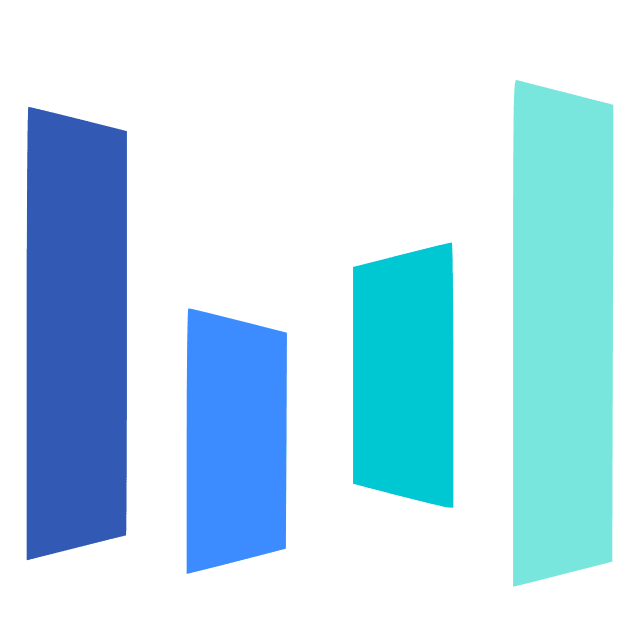}Seed1.8~\cite{Seed_1_8_2026}               & -         &        & 0.21          & 0.67          \\
                                                    & \logo{bytedance}Seed2.0 Pro~\cite{Seed_2_0_2026}           & -         &        & 0.18          & 0.71          \\
                                                    & \logo{bytedance}Seed2.0 Pro~\cite{Seed_2_0_2026}           & -         & \cmark & 0.26          & 0.72          \\
                                                    & \logo{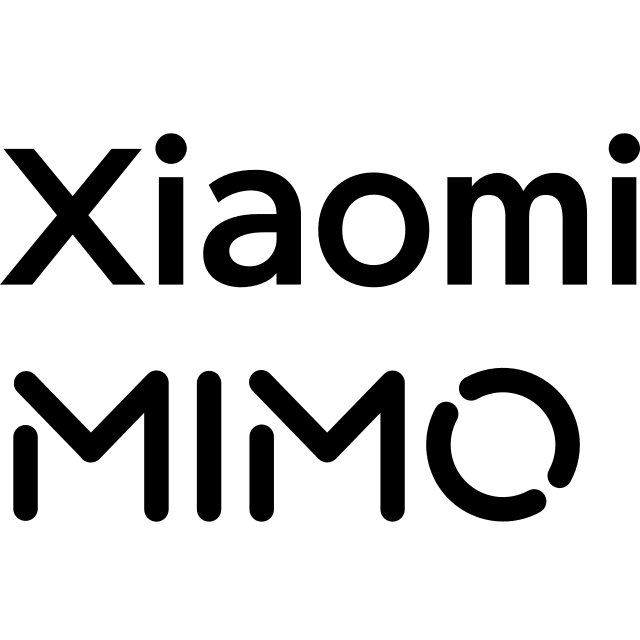}MiMo-V2-Omni~\cite{Mimo_V2_Omni_2026}     & -         & \cmark & 0.09          & 0.55          \\
                                                    & \logo{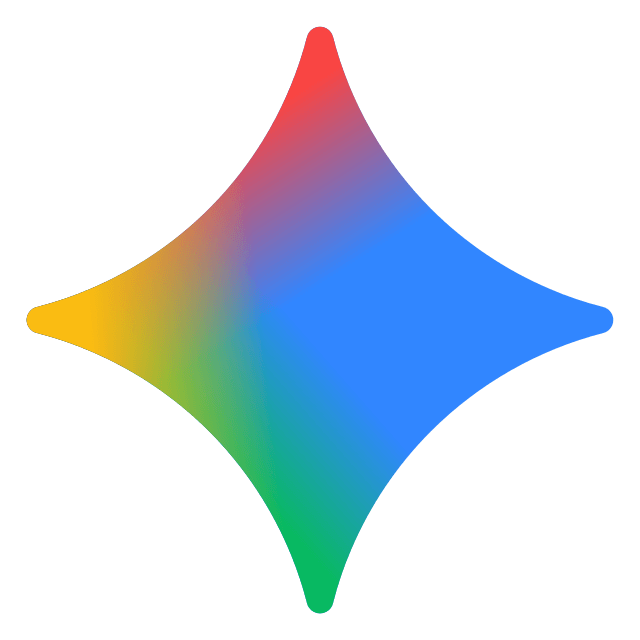}Gemini 2.5 Pro~\cite{Gemini_2_5_2025}         & -         & \cmark & 0.08          & 0.52          \\
                                                    & \logo{gemini}Gemini 3.1 Pro~\cite{Gemini_3_1_2026}         & -         & \cmark & 0.18          & 0.68          \\
                                                    & \logo{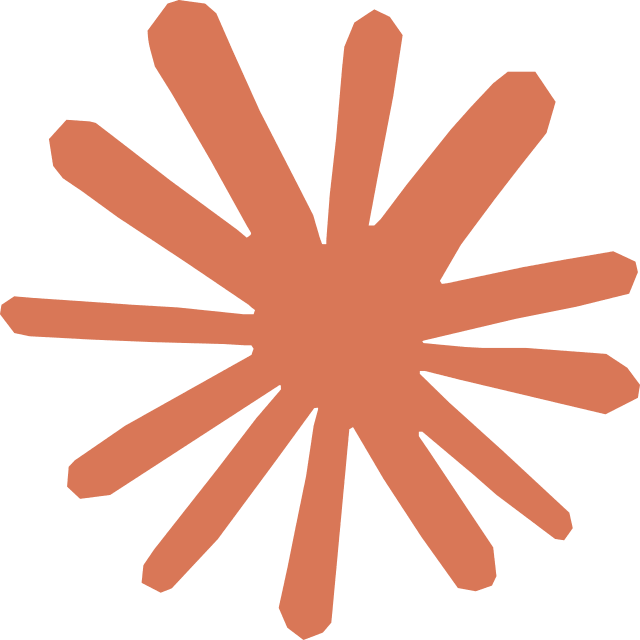}Claude Opus 4.7~\cite{Claude_Opus_4_7_2026}   & -         & \cmark & 0.10          & 0.50          \\
            \cmidrule(lr){1-6}
            \multirow{6}{*}[-0.6em]
            {\makecell{Expert \nextline OCR \nextline Models}}
                                                    & \logo{deepseek}DeepSeek-OCR~\cite{DeepSeek_OCR_2025}       & 3B-A0.5B  &        & 0.01          & 0.24          \\
                                                    & \logo{xiaohongshu}dots.ocr~\cite{dots_ocr_2025}            & 3B        &        & 0.05          & 0.47          \\
                                                    & \logo{glmv}GLM-OCR~\cite{GLM_OCR_2026}                     & 0.9B      &        & 0.06          & 0.38          \\
                                                    & \logo{paddle}PaddleOCR-VL-1.6~\cite{PaddleOCR_VL_1_6_2026} & 0.9B      &        & 0.05          & 0.41          \\
                                                    & \logo{baidu}Unlimited-OCR~\cite{Unlimited_OCR_2026}        & 3B-A0.5B  &        & 0.01          & 0.21          \\
            \rowcolor{hunyuanblue}\cellcolor{white} & \logo{hunyuan}\textbf{HunyuanOCR-1.5}                      & 1B        &        & \textbf{0.54} & \textbf{0.79} \\

            \bottomrule
        \end{tabular}
    }
    \label{tab:Chronicles_OCR_results}
\end{table}

\mypara{Chronicles-OCR.}
Chronicles-OCR~\cite{Chronicles_OCR_2026} evaluates ancient-script parsing, which is one of the key directions strengthened in HunyuanOCR-1.5. As shown in~\cref{tab:Chronicles_OCR_results}, HunyuanOCR-1.5 achieves SOTA performance within a 1B model, demonstrating substantially improved recognition ability on the seven historical forms of Chinese characters, historical documents, and ancient-script images. This result verifies that the data construction and training strategy for ancient scripts effectively improve the model's perception of historical glyphs.

\begin{table*}[t!]
    \small
    \centering
    \caption{
        \tbf{Comparison of chart deplotting results on ChartArena.}
        We report mAP$_{\text{high}}$ per chart type and the overall average, with separate \textit{EN} (English) and \textit{ZH} (Chinese) scores, each averaged over three visual styles.
    }
    \vspace{-2mm}
    {
        \renewcommand{\arraystretch}{1.1}
        \setlength{\tabcolsep}{3pt}
        \resizebox{\textwidth}{!}{%
            \begin{tabular}{cl cc cc cc cc cc cc cc cc cc}
                \toprule
                \multirow{2}{*}[-0.8mm]
                {\tbf{\makecell{Model \nextline Type}}} &
                \multirow{2}{*}[-0.8mm]
                {\tbf{Model}}                           &
                \multicolumn{2}{c}{\tbf{bar}}           &
                \multicolumn{2}{c}{\tbf{line}}          &
                \multicolumn{2}{c}{\tbf{pie}}           &
                \multicolumn{2}{c}{\tbf{radar}}         &
                \multicolumn{2}{c}{\tbf{box plot}}      &
                \multicolumn{2}{c}{\tbf{comb.}}         &
                \multicolumn{2}{c}{\tbf{flowchart}}     &
                \multicolumn{2}{c}{\tbf{mind map}}      &
                \multicolumn{2}{c}{\tbf{Average}}
                \\
                \cmidrule(lr){3-4}
                \cmidrule(lr){5-6}
                \cmidrule(lr){7-8}
                \cmidrule(lr){9-10}
                \cmidrule(lr){11-12}
                \cmidrule(lr){13-14}
                \cmidrule(lr){15-16}
                \cmidrule(lr){17-18}
                \cmidrule(lr){19-20}
                                                        &
                                                        &
                EN                                      & ZH                                                                &
                EN                                      & ZH                                                                &
                EN                                      & ZH                                                                &
                EN                                      & ZH                                                                &
                EN                                      & ZH                                                                &
                EN                                      & ZH                                                                &
                EN                                      & ZH                                                                &
                EN                                      & ZH                                                                &
                EN                                      & ZH
                \\
                \midrule
                \multirow{4}{*}[-0.3mm]
                {\makecell{General \nextline Purpose \nextline VLMs}}
                                                        & \logo{qwen}Qwen2.5-VL-7B-Ins.~\cite{Qwen2_5_VL_2025}              & 15.2               & 36.9               & 17.9               & 39.9               & 63.4              & 73.1               & 8.3               & 19.1               & 0.9                & 2.8                & 6.0            & 40.6              & 29.7               & 23.2               & 45.4           & 29.9           & 23.3               & 33.2               \\
                                                        & \logo{intern}InternVL3.5-8B~\cite{InternVL3_5_2025}               & 22.7               & 52.6               & 34.4               & 53.7               & 65.8              & 73.8               & 14.0              & 34.7               & 5.6                & 9.5                & 11.3           & 42.1              & 32.6               & 23.8               & 48.3           & 31.8           & 29.3               & 40.2               \\
                                                        & \logo{qwen}Qwen3-VL-8B-Ins.~\cite{Qwen3_VL_2025}                  & 27.5               & 58.6               & 35.5               & \ul{61.1}          & 77.3              & \ul{84.7}          & 16.8              & 42.6               & 11.6               & 12.1               & 13.2           & 47.9              & \ul{50.0}          & \ul{41.5}          & \ul{66.4}      & \ul{54.6}      & 37.3               & 50.4               \\
                                                        & \logo{qwen}Qwen3.5-9B~\cite{Qwen3_5_2026}                         & 32.5               & 45.1               & 45.5               & 54.1               & \tbf{82.6}        & 76.9               & 22.0              & 44.8               & 15.3               & 18.1               & 16.8           & 49.5              & 45.5               & 38.1               & 64.2           & 54.5           & \ul{40.6}          & 47.7               \\
                \midrule
                \multirow{7}{*}[-2mm]
                {\makecell{Expert\nextline Chart \nextline Deplotting \nextline Models}}
                                                        & \logo{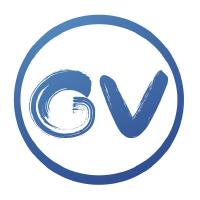}ChartAst (13B)~\cite{ChartAssisstant_2024}        & 5.2                & --                 & 4.2                & --                 & 0.3               & --                 & 1.5               & --                 & 0.3                & --                 & 0.0            & --                & --                 & --                 & --             & --             & 1.4                & --                 \\
                                                        & \logo{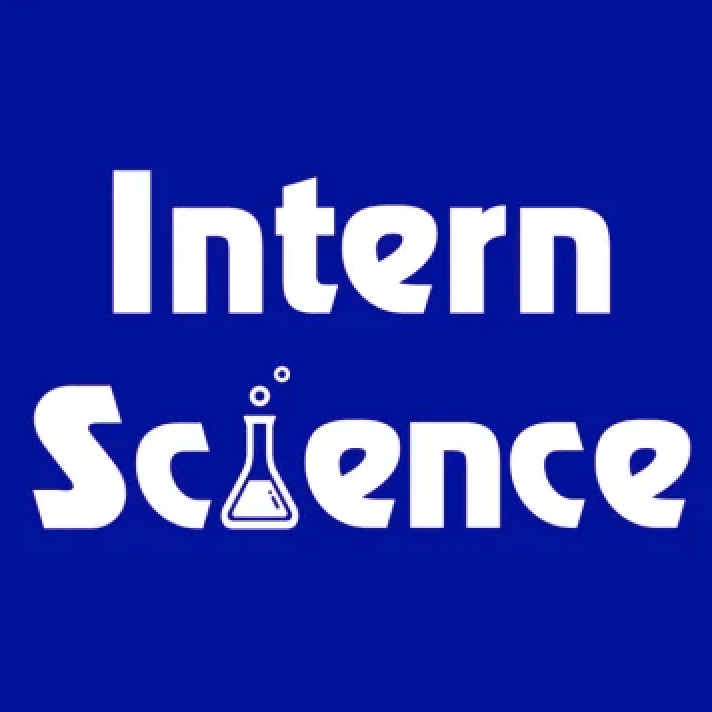}ChartVLM (8.3B)~\cite{ChartX_ChartVLM_2025}   & 11.2               & 5.3                & 11.5               & 4.3                & 12.9              & 8.2                & 2.1               & 5.0                & 0.7                & 0.4                & 4.1            & 4.4               & --                 & --                 & --             & --             & 5.3                & 3.5                \\
                                                        & \logo{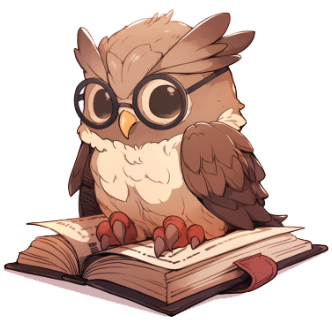}TinyChart (3B)~\cite{TinyChart_2024}                  & 6.1                & 6.3                & 9.7                & 3.2                & 5.7               & 5.4                & 0.5               & 3.4                & 0.2                & 1.3                & 0.7            & 4.2               & --                 & --                 & --             & --             & 2.9                & 3.0                \\
                                                        & \logo{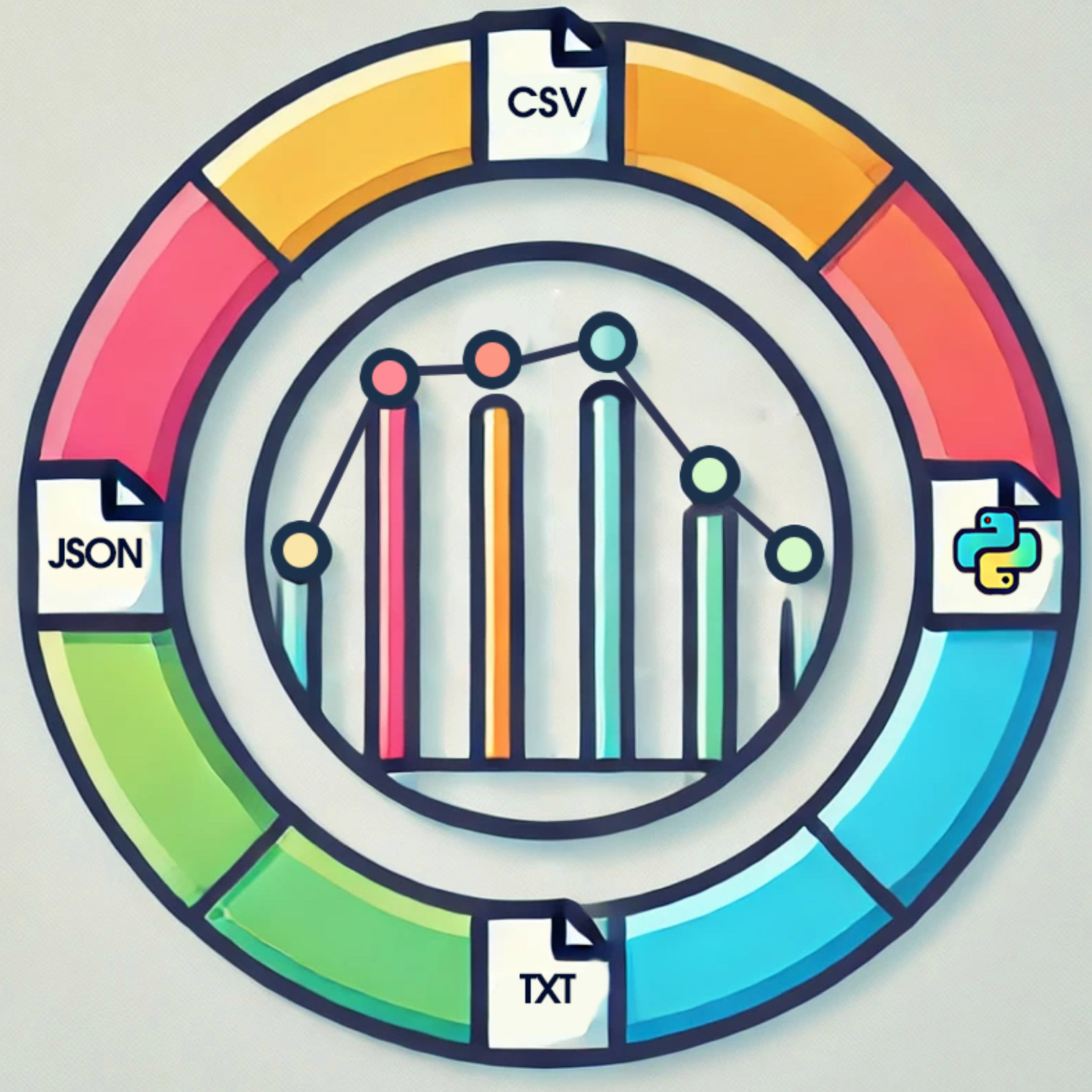}ChartMoE (8B)~\cite{ChartMoE_2024}                 & 18.7               & 24.4               & 14.7               & 22.3               & 15.0              & 48.5               & 3.7               & 16.1               & 2.7                & 1.6                & 5.1            & 19.5              & 4.0                & --                 & 4.1            & --             & 8.5                & 16.7               \\
                                                        & \logo{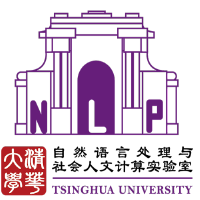}ChartCoder (7B)~\cite{ChartCoder_2025}               & 23.2               & 12.6               & 22.0               & 19.6               & 34.3              & 16.7               & 5.5               & 13.9               & 5.4                & 11.4               & 3.7            & 5.1               & 5.6                & --                 & 1.0            & --             & 12.6               & 9.9                \\
                                                        & \logo{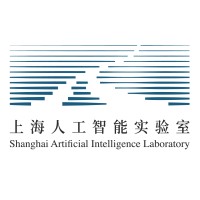}RRVF (7B)~\cite{RRVF_2025}                    & 35.8               & \ul{66.5}          & 41.5               & 54.3               & 51.6              & 75.3               & 16.6              & 40.3               & 14.7               & 14.1               & 23.5           & 61.2              & 36.4               & 32.4               & \tbf{68.4}     & \tbf{63.8}     & 36.0               & \ul{51.0}          \\
                                                        & \logo{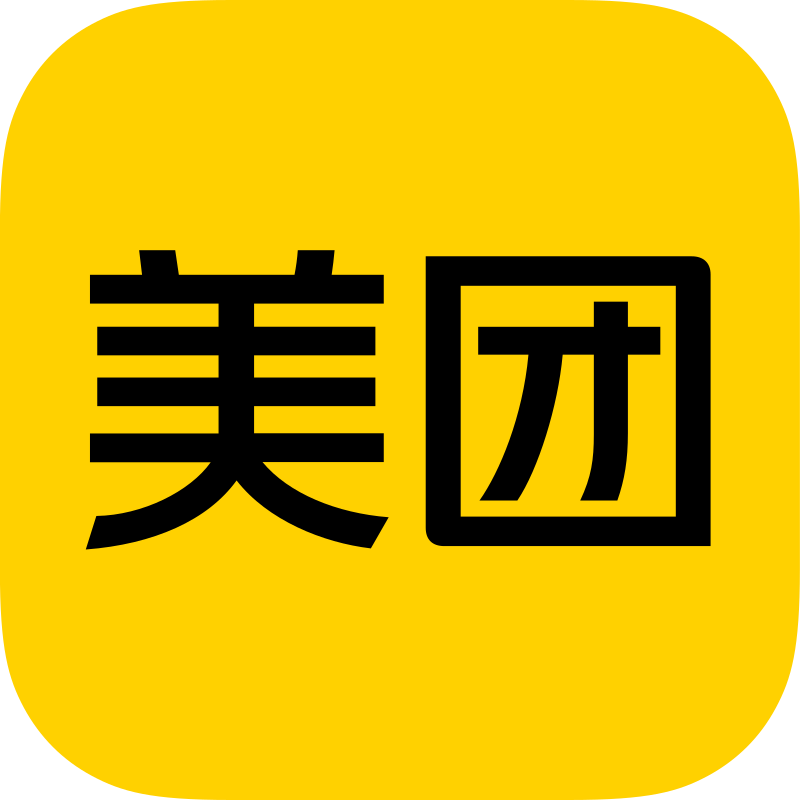}MSRL (7B)~\cite{MSRL_2025}                          & 32.7               & 45.2               & 35.2               & 34.3               & 41.2              & 67.9               & \tbf{25.9}        & \ul{48.0}          & 11.2               & 13.0               & 16.7           & 35.2              & 23.2               & 12.4               & 31.0           & 18.8           & 27.1               & 34.3               \\
                \midrule
                \multirow{4}{*}[-0.3mm]
                {\makecell{Expert \nextline OCR \nextline Models}}
                                                        & \logo{xiaohongshu}dots.mocr (3B)~\cite{Dots_mOCR_2026}            & 28.3               & 40.9               & 41.8               & 60.1               & 68.8              & 78.3               & 20.3              & 43.1               & \ul{24.1}          & 16.0               & \ul{26.9}      & 47.1              & 26.2               & 20.6               & 28.7           & 19.6           & 33.1               & 40.7               \\
                                                        & \logo{paddle}PaddleOCR-VL-1.5 (0.9B)~\cite{PaddleOCR_VL_1_5_2026} & 31.8               & 49.3               & 43.0               & 51.6               & 57.5              & 75.2               & 14.4              & 29.0               & 11.7               & 20.7               & 21.3           & 54.0              & --                 & --                 & --             & --             & 23.9               & 35.8               \\
                                                        & \logo{paddle}PaddleOCR-VL-1.6 (0.9B)~\cite{PaddleOCR_VL_1_6_2026} & \ul{39.9}          & 56.8               & \ul{53.5}          & 57.6               & 59.7              & 80.9               & 19.1              & 35.3               & 10.9               & \ul{31.4}          & \tbf{28.6}     & \tbf{65.5}        & --                 & --                 & --             & --             & 27.5               & 41.7               \\
                                                        & \hybcell\logo{hunyuan}\tbf{HunyuanOCR-1.5}                        & \hybcell\tbf{47.4} & \hybcell\tbf{73.9} & \hybcell\tbf{59.6} & \hybcell\tbf{73.4} & \hybcell\ul{79.7} & \hybcell\tbf{91.5} & \hybcell\ul{23.0} & \hybcell\tbf{50.5} & \hybcell\tbf{52.1} & \hybcell\tbf{64.3} & \hybcell{24.8} & \hybcell\ul{61.7} & \hybcell\tbf{67.8} & \hybcell\tbf{64.2} & \hybcell{36.5} & \hybcell{33.3} & \hybcell\tbf{48.9} & \hybcell\tbf{64.1} \\
                \bottomrule
            \end{tabular}
        }
    }
    \vspace{-0.4em}
    \label{tab:ChartArena_results}
\end{table*}

\mypara{ChartArena.}
ChartArena~\cite{ChartArena_2026} provides a fine-grained evaluation of structured chart parsing. While the HunyuanOCR series already had basic chart parsing ability, HunyuanOCR-1.5 further improves its parsing of chart text, legends, axes, visual element relations, and chart semantics. As shown in~\cref{tab:ChartArena_results}, HunyuanOCR-1.5 reaches a performance level comparable to 8B-scale models with only a 1B model, indicating strong capability in structured chart understanding and semantic recovery.

\begin{table}[t!]
  \small
  \centering
  \caption{
    \textbf{Comparison of table parsing results on TableVerse-5K.}
    We report TEDS and TEDS-S scores for table structure and content reconstruction.
  }
  {
    \renewcommand{\arraystretch}{0.95}
    \setlength{\tabcolsep}{14pt}
    \resizebox{\columnwidth}{!} {%
      \begin{tabular}{cll ccc}
        \toprule
        \multirow{2}{*}[-0.4mm]
        {\tbf{\makecell{Model Type}}}   &
        \multirow{2}{*}[-0.4mm]
        {\tbf{Model}}                   &
        \multirow{2}{*}[-0.4mm]
        {\tbf{Size}}                    &
        \multirow{2}{*}[-0.4mm]
        {\tbf{\makecell{Release Date}}} &
        \multicolumn{2}{c}{\tbf{TableVerse-5K}}
        \\
        \cmidrule(lr){5-6}
                                        &                                                            &             &                  & TEDS                & TEDS-S              \\
        \midrule
        \multirow{2}{*}
        {\makecell{Specialized Table \nextline Parsing Models}}
                                        & UniTable~\cite{UniTable_2024}                              & 125M        & 2024.03          & 48.55               & 78.65               \\
                                        & TRivia-3B~\cite{TRivia_2026}                               & 3B          & 2025.12          & \tbf{78.15}         & \tbf{85.41}         \\
        \midrule
        \multirow{8}{*}[-0.8em]
        {\makecell{General \nextline VLMs}}
                                        & \logo{openai}GPT-4o~\cite{GPT_4o_2024}                     & -           & 2024.05          & 63.62               & 76.41               \\
                                        & \logo{openai}GPT-5~\cite{GPT_5_2025}                       & -           & 2025.08          & 67.04               & 78.96               \\
                                        & \logo{qwen}Qwen2.5-VL-72B-Ins.~\cite{Qwen2_5_VL_2025}      & 72B         & 2025.02          & 75.23               & 82.65               \\
                                        & \logo{intern}InternVL3.5-A28B~\cite{InternVL3_5_2025}      & 241B-A28B   & 2025.08          & 76.08               & 84.96               \\
                                        & \logo{qwen}Qwen3-VL-A22B-Ins.~\cite{Qwen3_VL_2025}         & 235B-A22B   & 2025.10          & 78.26               & 84.23               \\
                                        & \logo{bytedance}Seed1.8 (no-think)~\cite{Seed_1_8_2026}    & -           & 2025.12          & \tbf{79.91}         & 86.03               \\
                                        & \logo{kimi}Kimi K2.5 (no-think)~\cite{Kimi_K2_5_2026}      & 1T          & 2026.02          & 78.75               & \ul{86.95}          \\
                                        & \logo{gemini}Gemini 2.5 Pro~\cite{Gemini_2_5_2025}         & -           & 2025.03          & \ul{79.46}          & \tbf{87.13}         \\
        \midrule
        \multirow{10}{*}[-1.0em]
        {\makecell{Expert \nextline OCR \nextline Models}}
                                        & \logo{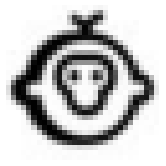}MonkeyOCR-pro-1.2B~\cite{MonkeyOCR_2025}   & 1.2B        & 2025.07          & 67.98               & 72.91               \\
                                        & \logo{monkeyocr}MonkeyOCR-pro-3B~\cite{MonkeyOCR_2025}     & 3B          & 2025.07          & 72.26               & 77.04               \\
                                        & \logo{deepseek}DeepSeek-OCR~\cite{DeepSeek_OCR_2025}       & 3B-A0.5B    & 2025.10          & 68.70               & 76.84               \\
                                        & \logo{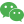}POINTS-Reader~\cite{POINTS_reader_2025}       & 3B          & 2025.08          & 72.03               & 81.13               \\
                                        & \logo{meituan}FD-RL~\cite{FD_RL_2026}                      & 4B          & 2025.11          & 74.31               & 80.51               \\
                                        & \logo{xiaohongshu}dots.ocr~\cite{dots_ocr_2025}            & 3B          & 2025.07          & 73.39               & 81.84               \\
                                        & \logo{paddle}PaddleOCR-VL~\cite{PaddleOCR_VL_2025}         & 0.9B        & 2025.10          & 77.55               & 84.08               \\
                                        & \logo{shanghaiailab}MinerU2.5~\cite{MinerU_2_5_2026}      & 1.2B        & 2025.09          & 77.41               & 84.31               \\
                                        & \logo{paddle}PaddleOCR-VL-1.6~\cite{PaddleOCR_VL_1_6_2026} & 0.9B        & 2026.06          & \ul{78.85}          & \ul{85.41}          \\
                                        & \hybcell\logo{hunyuan}\tbf{HunyuanOCR-1.5}                 & \hybcell 1B & \hybcell 2026.07 & \hybcell\tbf{79.37} & \hybcell\tbf{86.05} \\
        \bottomrule
      \end{tabular}
    }
  }
  \vspace{-1mm}
  \label{tab:TableVerse_5K_results}
\end{table}

\mypara{TableVerse-5K.}
TableVerse-5K~\cite{StrucTab_2026} evaluates table element parsing in complex table scenarios. Compared with table subsets in general document parsing benchmarks, this benchmark focuses more on table structure, cell content, row-column relations, and table reconstruction under complex layouts. As shown in~\cref{tab:TableVerse_5K_results}, HunyuanOCR-1.5 achieves the best performance among expert OCR models on TableVerse-5K, showing that the model further enhances its capability in table structure parsing and complex table reconstruction.

\mypara{DUDE.}
DUDE~\cite{DUDE_2023} evaluates document-level multi-image QA, where the model needs to retrieve information, aggregate evidence, and answer questions across multiple pages or images. This task goes beyond conventional single-image OCR and single-page document parsing, and is used here to examine whether an OCR-specialized VLM can extend toward document-level multi-image understanding. HunyuanOCR-1.5 achieves 54.64 on the DUDE validation set, which is close to the 56.41 result of the general multimodal model Qwen3.5-0.8B. This result indicates that, after multi-image data construction and training adaptation, HunyuanOCR-1.5 has acquired a certain degree of document-level multi-image QA capability and reaches a comparable level to a general-purpose VLM in this setting.

\begin{table}[t]
    \small
    \centering
    \caption{
        \textbf{Results on the MORE benchmark.}
        We evaluate low-resource multilingual parsing across text, formula, table, code, catalog, and reading-order dimensions.
    }
    \label{tab:MORE_results}
    \setlength{\tabcolsep}{6pt}
    \resizebox{\columnwidth}{!}{%
        \begin{tabular}{cll ccccccc}
            \toprule
            \tbf{\makecell{Model \nextline Type}} &
            \tbf{Model}                           &
            \tbf{Size}                            &
            \tbf{Overall$\uparrow$}               &
            Text$\uparrow$                        &
            Formula$\uparrow$                     &
            Table$\uparrow$                       &
            Code$\uparrow$                        &
            Catalog$\uparrow$                     &
            \makecell{Reading \nextline Order$\uparrow$}
            \\
            \midrule
            \multirow{3}{*}[-0.6mm]
            {\makecell{General \nextline Purpose \nextline VLMs}}
                                                  & \logo{qwen}Qwen3-VL~\cite{Qwen3_VL_2025}                   & 2B          & 83.56               & 92.02          & 65.45          & 65.21          & 92.38               & 93.76          & 92.53          \\
                                                  & \logo{qwen}Qwen2.5-VL~\cite{Qwen2_5_VL_2025}               & 3B          & 83.93               & 89.36          & 84.48          & 68.27          & 86.69               & 92.54          & 82.23          \\
                                                  & \logo{gemini}Gemini 3~\cite{Gemini_3_2026}                 & -           & 91.61               & \tbf{95.39}    & 90.27          & 81.02          & 93.05               & \tbf{94.31}    & 95.63          \\
            \cmidrule(lr){1-10}
            \multirow{8}{*}[-2.0mm]
            {\makecell{Expert \nextline OCR \nextline Models}}
                                                  & \logo{shanghaiailab}MinerU2.5~\cite{MinerU_2_5_2026}      & 1.2B        & 48.85               & 27.12          & 73.29          & 33.83          & 72.41               & 21.61          & 64.81          \\
                                                  & \logo{deepseek}DeepSeek-OCR 2~\cite{DeepSeek_OCR_2_2026}   & 3B-A0.5B    & 82.91               & 85.27          & 75.67          & 61.63          & 92.26               & 88.26          & 94.36          \\
                                                  & \logo{xiaohongshu}dots.ocr~\cite{dots_ocr_2025}            & 3B          & 84.31               & 94.45          & 90.77          & 39.81          & 95.38               & 88.26          & \tbf{97.18}    \\
                                                  & \logo{baidu}Unlimited-OCR~\cite{Unlimited_OCR_2026}        & 3B-A0.5B    & 84.90               & 86.75          & \tbf{92.22}    & 50.89          & 97.45               & 85.95          & 96.17          \\
                                                  & \logo{glmv}GLM-OCR~\cite{GLM_OCR_2026}                     & 0.9B        & 85.75               & 87.31          & 89.29          & \tbf{82.48}    & 95.83               & 67.12          & 92.48          \\
                                                  & \logo{paddle}PaddleOCR-VL~\cite{PaddleOCR_VL_2025}         & 0.9B        & 87.96               & 90.99          & 91.11          & 61.11          & 96.29               & 93.04          & 95.19          \\
                                                  & \logo{paddle}PaddleOCR-VL-1.6~\cite{PaddleOCR_VL_1_6_2026} & 0.9B        & 89.88               & 90.28          & 89.16          & 76.46          & 97.47               & 92.86          & 93.05          \\
                                                  & \hybcell\logo{hunyuan}\tbf{HunyuanOCR-1.5}                 & \hybcell 1B & \hybcell\tbf{91.90} & \hybcell 91.31 & \hybcell 91.10 & \hybcell 80.77 & \hybcell\tbf{99.10} & \hybcell 92.66 & \hybcell 96.48 \\
            \bottomrule
        \end{tabular}
    }
\end{table}

\mypara{MORE.}
MORE~\cite{MORE_2026} evaluates low-resource multilingual parsing across 149 languages, focusing on low-resource languages and long-tail writing systems. As shown in~\cref{tab:MORE_results}, HunyuanOCR-1.5 achieves SOTA performance among OCR expert models on MORE. This result demonstrates that the low-resource language data produced by Agentic Data Flow effectively improves multilingual perception, and further validates the value of systematic data construction for low-resource OCR capability expansion.

\begin{table}[t]
    \centering
    \small
    \caption{
        \textbf{Results on CHAOS-Bench.}
        We report the page-average recall of perturbed seen-text words, measuring output faithfulness under conflicts between visual evidence and language priors.
    }
    \label{tab:CHAOS_Bench_results}
    \setlength{\tabcolsep}{16pt}
    \resizebox{0.7\columnwidth}{!}{
        \begin{tabular}{llc}
            \toprule
            \textbf{Model}                                               &
            \textbf{Size}                                                &
            \textbf{Page-avg Recall$\uparrow$}
            \\
            \midrule
            \logo{xiaohongshu}dots.ocr~\cite{dots_ocr_2025}              & 3B          & 3.02                \\
            \logo{glmv}GLM-OCR~\cite{GLM_OCR_2026}                       & 0.9B        & 5.75                \\
            \logo{paddle}PaddleOCR-VL-1.6~\cite{PaddleOCR_VL_1_6_2026}   & 0.9B        & 5.95                \\
            \logo{deepseek}DeepSeek-OCR 2~\cite{DeepSeek_OCR_2_2026}     & 3B-A0.5B    & \ul{6.33}           \\
            \logo{shanghaiailab}MinerU2.5-Pro~\cite{MinerU_2_5_Pro_2026} & 1.2B        & \ul{6.33}           \\
            \hybcell\logo{hunyuan}\tbf{HunyuanOCR-1.5}                   & \hybcell 1B & \hybcell\tbf{14.15} \\
            \bottomrule
        \end{tabular}
    }
\end{table}

\mypara{CHAOS-Bench.}
CHAOS-Bench is introduced in this work to evaluate output faithfulness and the model's adherence to the seen-text principle. It modifies characters in selected words from academic paper images to create meaningless words, and then checks whether the model preserves these visually observed meaningless words in its parsed output. As shown in~\cref{tab:CHAOS_Bench_results}, HunyuanOCR-1.5 achieves the best result among compared models, with a page-average recall of 14.15. However, the absolute recall remains low, indicating that faithfully preserving visually observed but semantically invalid text is still a challenging problem for current OCR-centric VLMs. This result suggests that HunyuanOCR-1.5 is less biased toward language priors than existing models, but also highlights the need for further research on hallucination suppression and seen-text faithful generation.

Overall, these boundary capability evaluations show that the improvements of HunyuanOCR-1.5 are reflected not only in conventional OCR metrics, but also in more fine-grained boundary scenarios and reliability-oriented evaluations. Results on ancient scripts, low-resource languages, complex tables, structured charts, multi-image QA, and output faithfulness collectively demonstrate that Agentic Data Flow, the upgraded training recipe, and post-training optimization effectively promote the systematic improvement of HunyuanOCR-1.5 in both strengthened existing abilities and newly expanded capability boundaries.

\subsection{Updates on Existing Benchmarks}

\begin{table}[h]
    \small
    \centering
    \caption{
        \textbf{Comparison of document parsing results on OmniDocBench v1.6.}
        We report the overall score together with per-dimension metrics on text, formula, table, and reading order.
    }
    \label{tab:OmniDocBench_results}
    \setlength{\tabcolsep}{6pt}
    \resizebox{\columnwidth}{!}{
        \begin{tabular}{cll cccccc}
            \toprule
            \textbf{\makecell{Model \nextline Type}}          &
            \textbf{Model}                                    &
            \textbf{Size}                                     &
            \textbf{Overall$\uparrow$}                        &
            \textbf{Text\textsuperscript{Edit}$\downarrow$}   &
            \textbf{Formula\textsuperscript{CDM}$\uparrow$}   &
            \textbf{Table\textsuperscript{TEDS}$\uparrow$}    &
            \textbf{Table\textsuperscript{TEDS\_S}$\uparrow$} &
            \textbf{Order\textsuperscript{Edit}$\downarrow$}
            \\

            \midrule
            \multirow{7}{*}[-1.8mm]
            {\makecell{General \nextline Purpose \nextline VLMs}}
                                                              & \logo{intern}InternVL3.5-241B~\cite{InternVL3_5_2025}      & 241B      & 83.76          & 0.130       & 89.95       & 74.35       & 79.78       & 0.215       \\
                                                              & \logo{kimi}Kimi K2.5~\cite{Kimi_K2_5_2026}                 & 1T        & 84.53          & 0.107       & 83.50       & 80.76       & 84.00       & 0.211       \\
                                                              & \logo{openai}GPT-5.2~\cite{GPT_5_2025}                     & -         & 86.59          & 0.114       & 88.21       & 82.95       & 87.93       & 0.193       \\
                                                              & \logo{qwen}Qwen3-VL-235B~\cite{Qwen3_VL_2025}              & 235B-A22B & 89.78          & 0.063       & 92.55       & 83.07       & 86.75       & 0.166       \\
                                                              & \logo{gemini}Gemini 3 Flash~\cite{Gemini_3_2026}           & -         & 92.62          & 0.066       & 95.16       & 89.29       & 93.51       & 0.172       \\
                                                              & \logo{gemini}Gemini 3 Pro~\cite{Gemini_3_2026}             & -         & 92.91          & 0.064       & \tbf{95.99} & 89.15       & 92.96       & 0.165       \\
                                                              & \logo{alibaba}Ovis2.6-30B-A3B~\cite{Ovis2_5_2025}          & 30B-A3B   & 93.70          & \tbf{0.035} & 95.17       & 89.44       & 92.40       & 0.135       \\
            \cmidrule(lr){1-9}
            \multirow{12}{*}[-2.8mm]
            {\makecell{End2End \nextline Expert \nextline OCR \nextline Models}}
                                                              & \logo{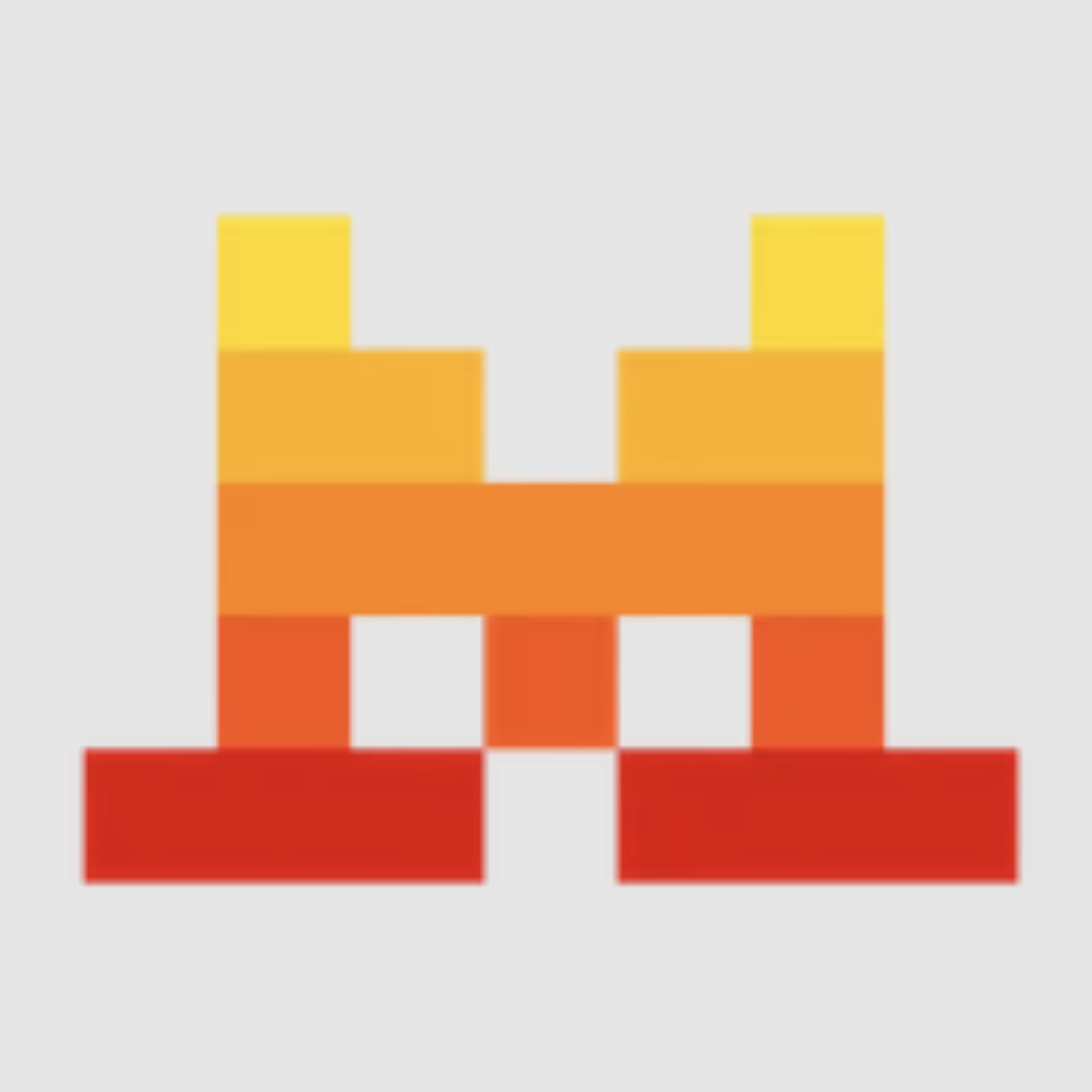}Mistral OCR~\cite{Mistral_OCR_2025}        & -         & 85.66          & 0.097       & 89.91       & 76.78       & 80.93       & 0.171       \\
                                                              & \logo{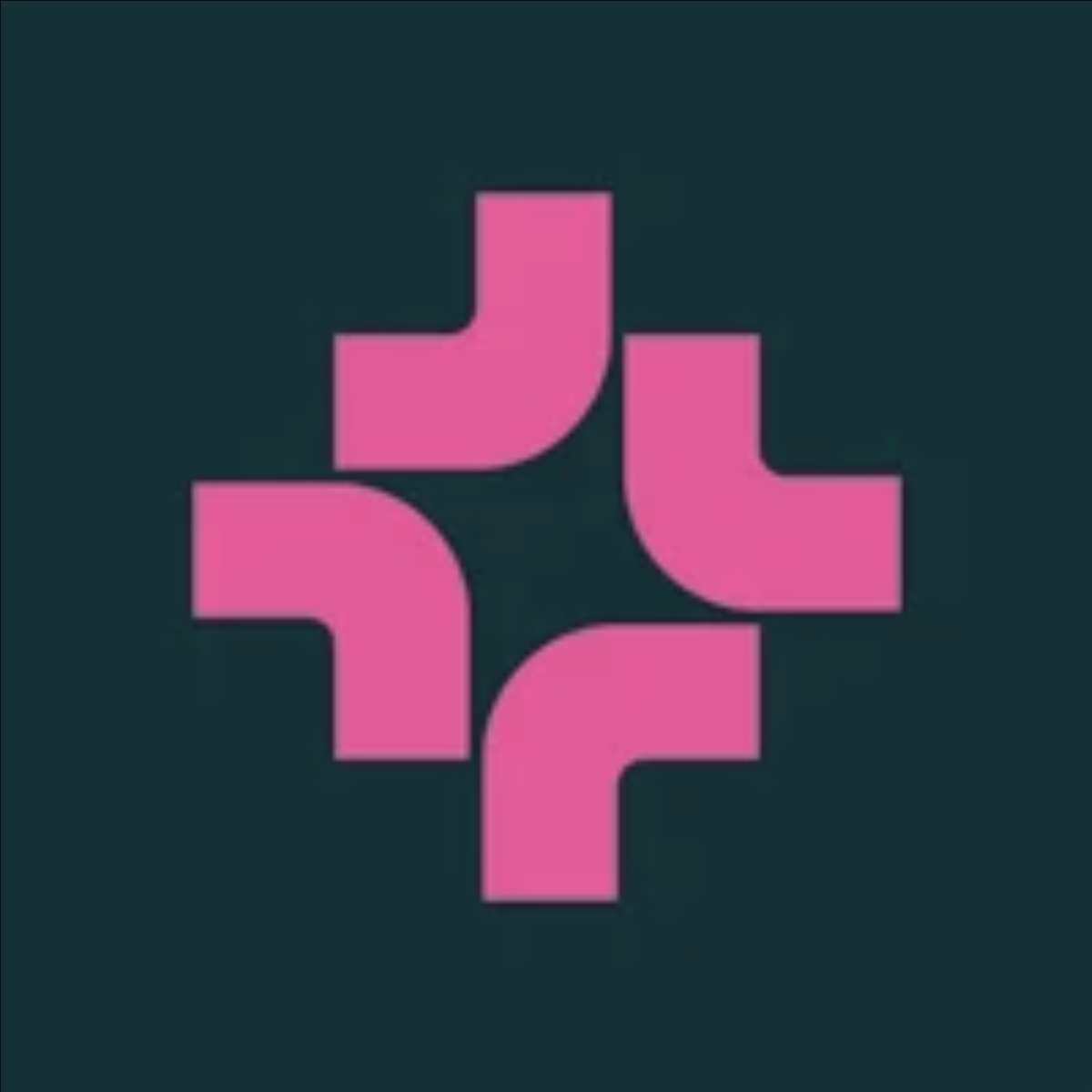}olmOCR~\cite{olmOCR_2025}                    & 7B        & 85.74          & 0.139       & 88.10       & 83.00       & 87.17       & 0.216       \\
                                                              & \logo{meituan}OCRVerse~\cite{OCRVerse_2026}                & 4B        & 88.60          & 0.063       & 89.61       & 82.44       & 86.27       & 0.163       \\
                                                              & \logo{deepseek}DeepSeek-OCR 2~\cite{DeepSeek_OCR_2_2026}   & 3B-A0.5B  & 90.25          & 0.050       & 91.84       & 83.89       & 87.75       & 0.144       \\
                                                              & \logo{xiaohongshu}dots.ocr~\cite{dots_ocr_2025}            & 3B        & 90.77          & 0.048       & 89.95       & 87.18       & 90.58       & 0.138       \\
                                                              & \logo{hunyuan}HunyuanOCR~\cite{HunyuanOCR_2025}            & 1B        & 92.03          & 0.048       & 88.60       & \ul{92.37}  & \ul{93.99}  & 0.138       \\
                                                              & \logo{xiaohongshu}FireRed-OCR~\cite{FireRed_OCR_2026}      & 2B        & 93.26          & \ul{0.037}  & 95.44       & 88.04       & 91.06       & 0.131       \\
                                                              & \logo{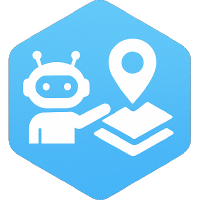}ABot-OCR~\cite{ABot_OCR_2026}                & 2B        & 93.30          & \ul{0.037}  & 94.86       & 88.69       & 91.87       & 0.137       \\
                                                              & \logo{alibaba}Logics-Parsing-v2~\cite{Logics_Parsing_2025} & 4B        & 93.33          & 0.041       & 95.65       & 88.42       & 91.98       & 0.137       \\
                                                              & \logo{baidu}Qianfan-OCR~\cite{Qianfan_OCR_2026}            & 4.7B      & 93.90          & 0.040       & 95.08       & 90.53       & 93.31       & 0.130       \\
                                                              & \logo{baidu}Unlimited-OCR~\cite{Unlimited_OCR_2026}        & 3B-A0.5B  & \ul{93.92}     & 0.042       & \ul{95.79}  & 90.16       & 93.32       & \tbf{0.129} \\
            \hybr\cellcolor{white}                            & \logo{hunyuan}\tbf{HunyuanOCR-1.5}                         & 1B        & \textbf{94.74} & 0.039       & 94.50       & \tbf{93.67} & \tbf{94.71} & \tbf{0.129} \\
            \bottomrule
        \end{tabular}
    }
\end{table}

After evaluating boundary capabilities, we further analyze the performance of HunyuanOCR-1.5 on existing evaluation dimensions already covered by HunyuanOCR-1.0. This part focuses on whether HunyuanOCR-1.5 can further improve or stably maintain its core OCR abilities, including end-to-end document parsing, text spotting, text image translation, information extraction, video subtitle extraction, and general OCR-aware QA.

\begin{table}[h!]
    \centering
    \footnotesize
    \setlength{\tabcolsep}{3pt}
    \caption{
        \textbf{Comprehensive evaluation of text spotting ability.}
        We report the overall score and per-scenario results across diverse image domains.
    }
    \label{tab:spotting_results}
    \resizebox{\linewidth}{!}{
        \begin{tabular}{llcccccccccc}
            \toprule
            \tbf{Model Type} &
            \tbf{Model}      &
            \tbf{Overall}    &
            Art              &
            Doc              &
            Game             &
            Hand             &
            Ads              &
            Receipt          &
            Screen           &
            Scene            &
            Video
            \\
            \midrule
            \multirow{2}{*}
            {\makecell{Traditional \nextline Methods}}
                             & \logo{paddle}PaddleOCR~\cite{PaddleOCR_3_0_2025}           & 53.38               & 32.83              & 70.23               & 51.59               & 56.39               & 57.38              & 50.59               & 63.38              & 44.68               & 53.35              \\
                             & \logo{baidu}BaiduOCR~\cite{BaiduOCRAPI_2025}               & 61.90               & 38.50              & \ul{78.95}          & 59.24               & 59.06               & 66.70              & \ul{63.66}          & 68.18              & 55.53               & 67.38              \\
            \midrule
            \multirow{6}{*}[-0.6mm]
            {\makecell{General \nextline Purpose \nextline VLMs}}
                             & \logo{gemini}Gemini 2.5 Pro~\cite{Gemini_2_5_2025}         & 23.44               & 21.79              & 35.16               & 10.02               & 38.49               & 29.89              & 20.80               & 17.59              & 18.33               & 18.90              \\
                             & \logo{qwen}Qwen3-VL-2B-Ins.~\cite{Qwen3_VL_2025}           & 29.68               & 29.43              & 19.37               & 20.85               & 50.57               & 35.14              & 24.42               & 12.13              & 34.90               & 40.10              \\
                             & \logo{qwen}Qwen3-VL-235B-A22B-Ins.~\cite{Qwen3_VL_2025}    & 53.62               & 46.15              & 43.78               & 48.00               & 68.90               & 64.01              & 47.53               & 45.91              & 54.56               & 63.79              \\
                             & \logo{bytedance}Seed2.0 Pro~\cite{Seed_2_0_2026}           & 56.32               & 44.77              & 45.85               & 61.70               & 66.89               & 61.87              & 55.73               & 52.05              & 46.53               & 71.49              \\
                             & \logo{gemini}Gemini 3.1 Pro~\cite{Gemini_3_1_2026}         & 59.53               & 46.83              & 54.89               & 62.62               & 63.37               & 63.96              & 54.53               & 64.29              & 55.30               & 70.02              \\
                             & \logo{qwen}Qwen3.5-A17B~\cite{Qwen3_5_2026}                & 59.76               & 44.92              & 52.56               & 58.16               & 71.54               & 67.42              & 55.98               & 62.58              & 56.11               & 68.56              \\
            \midrule
            \multirow{3}{*}[-0.1mm]
            {\makecell{OCR \nextline Models}}
                             & \logo{paddle}PaddleOCR-VL-1.6~\cite{PaddleOCR_VL_1_6_2026} & 61.95               & 41.36              & 72.20               & 58.56               & 70.61               & 65.24              & 61.85               & 63.63              & 54.60               & 69.52              \\
                             & \logo{hunyuan}HunyuanOCR~\cite{HunyuanOCR_2025}            & \ul{70.92}          & \tbf{56.76}        & 73.63               & \ul{73.54}          & \ul{77.10}          & \tbf{75.34}        & 63.51               & \tbf{76.58}        & \ul{64.56}          & \tbf{77.31}        \\
                             & \hybcell\logo{hunyuan}\tbf{HunyuanOCR-1.5}                 & \hybcell\tbf{71.40} & \hybcell\ul{53.21} & \hybcell\tbf{79.43} & \hybcell\tbf{75.84} & \hybcell\tbf{78.40} & \hybcell\ul{75.03} & \hybcell\tbf{65.22} & \hybcell\ul{74.51} & \hybcell\tbf{65.12} & \hybcell\ul{76.09} \\
            \bottomrule
        \end{tabular}
        \vspace{-10pt}
    }
\end{table}

\mypara{OmniDocBench.}
For end-to-end document parsing, HunyuanOCR-1.5 achieves an Overall score of 94.74 on OmniDocBench v1.6, reaching the SOTA performance among end-to-end OCR expert models, as shown in~\cref{tab:OmniDocBench_results}. This result shows that HunyuanOCR-1.5 further improves full-page document parsing while preserving the lightweight end-to-end architecture, with strong performance on structured parsing dimensions such as text, tables, and reading order. It is worth noting that HunyuanOCR-series models tend to parse multi-line formulas in a unified manner, using begin/end-style LaTeX syntax to represent the complete formula. However, the current OmniDocBench matching protocol splits independent multi-line formulas into single-line units before matching. This GT matching strategy is not fully aligned with complete multi-line formula outputs from end-to-end models, and may underestimate their actual formula parsing capability. This observation suggests that the evaluation protocol for long and multi-line formulas still has room for further refinement.

\mypara{Spotting Benchmark.}
For text spotting, HunyuanOCR-1.5 further improves over HunyuanOCR-1.0 on the in-house Spotting Benchmark, as shown in~\cref{tab:spotting_results}. In addition to regular text localization and recognition, HunyuanOCR-1.5 introduces negative-sample handling: when an input image contains no text, the model avoids producing hallucinated detection boxes and instead returns that no text is present. On an internal negative set of 1,000 text-free images, HunyuanOCR-1.5 achieves a no-text handling accuracy of 99.8\%, substantially outperforming HunyuanOCR-1.0 at 78.1\%. This ability is important for real-world OCR systems, where inputs do not always contain valid textual content.

\begin{table}[!t]
    \small
    \centering
    \caption{
        \textbf{Evaluation of text-image translation.}
        We report results on MMTIT (other-to-English and other-to-Chinese) and DoTA (English-to-Chinese) to evaluate the performance of text-image translation models.
    }
    \setlength{\abovecaptionskip}{0.2cm}
    \setlength{\tabcolsep}{3.5mm}{
        \begin{tabular}{llccc}
            \toprule
            \multirow{2}{*}{\raisebox{-0.7ex}{\textbf{Model}}}   &
            \multirow{2}{*}{\raisebox{-0.7ex}{\textbf{Size}}}    &
            \multicolumn{2}{c}{\textbf{MMTIT}}                   &
            \textbf{DoTA}                                                                                       \\
            \cmidrule(rl){3-4}
            \cmidrule(rl){5-5}
                                                                 &    & other2en    & other2zh    & en2zh       \\
            \midrule
            \logo{qwen}Qwen3-VL-8B-Instruct~\cite{Qwen3_VL_2025} & 8B & \ul{75.09}  & \ul{75.63}  & 79.86       \\
            \logo{qwen}Qwen3-VL-4B-Instruct~\cite{Qwen3_VL_2025} & 4B & 70.38       & 70.29       & 78.45       \\
            \logo{qwen}Qwen3-VL-2B-Instruct~\cite{Qwen3_VL_2025} & 2B & 66.30       & 66.77       & 73.49       \\
            \logo{paddle}PP-DocTranslation                       & -  & 52.63       & 52.43       & 82.09       \\
            \logo{hunyuan}HunyuanOCR~\cite{HunyuanOCR_2025}      & 1B & 73.38       & 73.62       & \ul{83.48}  \\
            \hybr\logo{hunyuan}\textbf{HunyuanOCR-1.5}           & 1B & \tbf{76.51} & \tbf{76.01} & \tbf{83.69} \\
            \bottomrule
        \end{tabular}
    }
    \label{tab:translation}
\end{table}

\mypara{Text Image Translation.}
We continue to monitor text image translation with DoTA and MMTIT, as shown in~\cref{tab:translation}. DoTA mainly evaluates English-to-Chinese translation for printed document images, where HunyuanOCR-1.5 preserves a capability level close to HunyuanOCR-1.0, indicating no clear degradation on the existing document translation setting. In contrast, MMTIT covers more languages and more diverse visual scenarios. Under this more challenging multilingual and multi-scenario setting, HunyuanOCR-1.5 is further optimized to improve its adaptability to multilingual text image translation.

\begin{table}[!t]
    \small
    \centering
    \caption{
        \textbf{Evaluation of information extraction (IE), video subtitles extraction, and visual question answering (VQA).}
        We report IE results on cards and receipts.
    }
    \setlength{\tabcolsep}{14pt}
    \resizebox{\linewidth}{!}{
        \begin{tabular}{lcccc}
            \toprule
            \multirow{2}{*}[-0.6mm]
            {\tbf{Model}}                                               &
            \multicolumn{2}{c}
            {\tbf{IE}}                                                  &
            \multirow{2}{*}[-0.6mm]
            {\tbf{\makecell{Video Subtitles \nextline Extraction}}}     &
            \multicolumn{1}{c}
            {\tbf{OCRBench}}
            \\
            \cmidrule(lr){2-3}
            \cmidrule(lr){5-5}
                                                                        & Cards       & Receipts    &             & Acc.      \\
            \midrule
            \logo{deepseek}DeepSeek-OCR~\cite{DeepSeek_OCR_2025}        & 10.04       & 40.54       & 5.41        & 430       \\
            \logo{paddle}PP-ChatOCR~\cite{PP_ChatOCR_2025}              & 57.02       & 50.26       & 3.10        & -         \\
            \logo{qwen}Qwen3-VL-2B-Instruct~\cite{Qwen3_VL_2025}        & 67.62       & 64.62       & 3.75        & 858       \\
            \logo{bytedance}Seed-1.6-Vision~\cite{Seed1_6_2025}         & 70.12       & 67.50       & 60.45       & \ul{881}  \\
            \logo{qwen}Qwen3-VL-235B-A22B-Instruct~\cite{Qwen3_VL_2025} & 75.59       & 78.40       & 50.74       & \tbf{920} \\
            \logo{gemini}Gemini 2.5 Pro~\cite{Gemini_2_5_2025}          & 80.59       & 80.66       & 53.65       & 872       \\
            \logo{hunyuan}HunyuanOCR~\cite{HunyuanOCR_2025}             & \ul{92.29}  & \ul{92.53}  & \ul{92.87}  & 860       \\
            \rowcolor{hunyuanblue}\logo{hunyuan}\tbf{HunyuanOCR-1.5}    & \tbf{92.40} & \tbf{92.55} & \tbf{93.07} & 861       \\
            \bottomrule
        \end{tabular}
    }
    \label{tab:IE_VQA_results}
\end{table}

\mypara{IE, Video Subtitle Extraction, and OCRBench.}
For information extraction, video subtitle extraction, and OCRBench, HunyuanOCR-1.5 largely maintains the capabilities established by HunyuanOCR-1.0, as shown in~\cref{tab:IE_VQA_results}. Information extraction and video subtitle extraction correspond to practical OCR applications such as structured field extraction and subtitle recognition from video frames, while OCRBench monitors general OCR-aware QA ability. These results indicate that HunyuanOCR-1.5 expands its boundary capabilities without sacrificing its existing core OCR abilities.

Overall, HunyuanOCR-1.5 shows further improvements on end-to-end document parsing and text spotting, preserves its printed-document translation capability while improving multilingual and multi-scenario translation adaptability, and maintains strong performance on information extraction, video subtitle extraction, and general OCR-aware QA. These results demonstrate that the capability boundary expansion of HunyuanOCR-1.5 is achieved without compromising the practical OCR abilities established in HunyuanOCR-1.0.

\section{Conclusion and Future Work}
\label{sec:conclusion}

We present HunyuanOCR-1.5, a lightweight end-to-end OCR-specialized VLM that advances HunyuanOCR-1.0 toward two goals: \textit{faster} inference and \textit{broader} OCR capabilities. Without redesigning the validated backbone, HunyuanOCR-1.5 integrates DFlash speculative decoding for long structured OCR generation, achieving substantial speedups under both Transformers and vLLM while also supporting PC-side deployment via llama.cpp. Meanwhile, Agentic Data Flow, together with upgraded pretraining and post-training recipes, extends the model toward 4K-resolution perception, 128K-context understanding, multi-image QA, low-resource multilingual OCR, ancient-script recognition, chart/table parsing, and more faithful document generation.
Through a capability-oriented evaluation tree, HunyuanOCR-1.5 demonstrates top-tier end-to-end document parsing performance, strong long-tail capability gains, and leading inference efficiency among compared OCR systems. We will release the model weights and training code to support reproducible research, user-side fine-tuning, and real-world deployment. Future work will further reduce high-resolution visual token redundancy, expand Agentic Data Flow toward continuous data-model co-evolution, and improve reliability for long and visually complex OCR generation.

\clearpage
{
    \small
    \setlength{\bibsep}{6pt}
    \bibliographystyle{unsrtnat}
    \bibliography{main}
}

\begin{CJK*}{UTF8}{gbsn}
    \appendix
\saferesetlinenumber[1]
\counterwithin{figure}{section}
\counterwithin{table}{section}
\maketitlesupplementary
\setcounter{page}{1}

\section*{Overview}
This material provides supplementary details to the main paper, organized as follows:
\vspace{-0.5em}
\begin{itemize}[
        label=\raisebox{0.5ex}{\tiny$\bullet$},
        leftmargin=1em,
        itemsep=0pt, 
        parsep=2pt, 
        partopsep=0pt 
    ]
    \item (\ref{supp:sec:model_architecture}) \textbf{Detailed Model Architecture}
    \item (\ref{supp:sec:recommend_instruction}) \textbf{Recommended Instruction}
    \item (\ref{supp:sec:chaos}) \textbf{CHAOS-Bench}
    \item (\ref{supp:sec:RL_details}) \textbf{Reinforcement Learning Details}
          \begin{itemize}[
                  label=\raisebox{0.2ex}{\tiny$\circ$},
                  leftmargin=1.4em,
                  itemsep=1pt,
                  parsep=0pt,
                  topsep=1pt,
                  partopsep=0pt
              ]
              \item (\ref{supp:subsec:RL_setup}) RL Setup
              \item (\ref{supp:subsec:RL_reward_design}) RL Reward Design
              \item (\ref{supp:subsec:RL_dynamics}) RL Dynamics
              \item (\ref{supp:subsec:RL_improvements}) Task-wise Performance Improvements
          \end{itemize}
    \item (\ref{supp:sec:qualitative_examples}) \textbf{Qualitative Examples}
\end{itemize}

\clearpage
\section{Detailed Model Architecture}
\label{supp:sec:model_architecture}

\mypara{Native-resolution visual encoder.}
The visual encoder is based on Hunyuan-ViT, a native-resolution Vision Transformer~\cite{ViT_2020, NaViT_2023}. It preserves the original aspect ratio of input images and encodes visual patches under their native spatial layout, which is important for OCR inputs with diverse shapes such as long receipts, dense documents, tables, charts, and scene-text images.

\mypara{Adaptive MLP connector.}
The adaptive MLP connector bridges the visual encoder and the language model. It performs learnable pooling and projection over high-resolution visual features, compressing dense visual patches into compact visual tokens while preserving key text-dense and layout-sensitive regions.

\mypara{Lightweight language model.}
The language component is a lightweight Hunyuan-0.5B~\cite{hunyuan2025llm} with a dense Transformer architecture~\cite{Attention_2017}. It takes visual tokens and instruction tokens as input and autoregressively generates OCR outputs. XD-RoPE~\cite{Qwen2_VL_2024, HunyuanOCR_2025, RoFormer_2021} is used to model text, height, width, and time dimensions, enabling unified reasoning over text sequences, document layouts, and multi-image or temporal inputs.

Overall, HunyuanOCR-1.5 keeps a fully end-to-end formulation. Given images and instructions, it directly generates task-specific outputs such as recognized text, spotting results, Markdown documents, HTML tables, LaTeX formulas, chart descriptions, and document-grounded answers.

\section{Recommended Instruction}
\label{supp:sec:recommend_instruction}

\Cref{supp:tab:recommend_instruction} summarizes the recommended instruction prompts for each task supported by HunyuanOCR-1.5, along with their English explanations for reference. We recommend using the Chinese instruction prompts directly to ensure the stability and reproducibility of benchmarking results.

Specifically, HunyuanOCR-1.5 newly supports Spotting outputs in the JSON format. For the ancient-text recognition benchmark Chronicles-OCR~\cite{Chronicles_OCR_2026}, we adopt the \textit{Parsing~v} prompt; for the chart deplotting benchmark ChartArena~\cite{ChartArena_2026}, we adopt the \textit{Parsing~iii} prompt. For OmniDocBench~\cite{OmniDocBench_2025}, we adopt the \textit{Parsing~iv} prompt, and for DoTA~\cite{DOTA_2024}, we adopt the \textit{Translation~i} prompt.

\begin{table}[htbp]
    \centering
    \caption{
        Recommended Chinese Instruction Prompts for Different Task Types, with English Explanations.
    }
    \label{supp:tab:recommend_instruction}
    \renewcommand{\arraystretch}{1.1}
    \small
    \begin{tabular}{C{1.6cm} C{0.3cm} L{5.75cm} L{5.75cm}}
        \toprule
        \textbf{Task}               &
        \textbf{\#}                 &
        \textbf{Instruction Prompt} &
        \textbf{Explanation}
        \\
        \midrule
        \multirow{8}{*}{\textbf{Spotting}}
                                    & \textit{i}
                                    & 检测并识别图片中的文字，将文本坐标格式化输出。
                                    & Detect and recognize text in the image, and output the text coordinates in a formatted manner.
        \\
        \cmidrule(lr){2-4}
                                    & \textit{ii}
                                    & 检测并识别图中所有的文字行，请按从上到下、从左到右的阅读顺序进行识别。输出格式为 JSON 数组，每个元素必须包含：``box'': [xmin, ymin, xmax, ymax]（坐标需归一化到 [0, 1000] 范围内）；``text'': ``识别出的文字内容''。注意：请直接输出 JSON 数组，不要包含任何多余的描述性文字。
                                    & Detect and recognize all text lines in the image in reading order (top-to-bottom, left-to-right). Output a JSON array where each element must contain: ``box'': [xmin, ymin, xmax, ymax] (coordinates normalized to the [0, 1000] range); ``text'': ``recognized text content''. Note: output the JSON array directly, without any additional descriptive text.
        \\
        \midrule

        \multirow{14}{*}{\textbf{Parsing}}
                                    & \textit{i}
                                    & 识别图片中的公式，用 \LaTeX 格式表示。
                                    & Identify the formula in the image and represent it using {\LaTeX} format.
        \\
        \cmidrule(lr){2-4}
                                    & \textit{ii}
                                    & 把图中的表格解析为 HTML。
                                    & Parse the table in the image into HTML.
        \\
        \cmidrule(lr){2-4}
                                    & \textit{iii}
                                    & 解析图中的图表，对于流程图使用 Mermaid 格式表示，其他图表使用 Markdown 格式表示。
                                    & Parse the chart in the image; use Mermaid format for flowcharts and Markdown for other charts.
        \\
        \cmidrule(lr){2-4}
                                    & \textit{iv}
                                    & 提取文档图片中正文的所有信息用 markdown 格式表示，其中页眉、页脚部分忽略，表格用 html 格式表达，文档中公式用 \LaTeX 格式表示，按照阅读顺序组织进行解析。
                                    & Extract all information from the main body of the document image and represent it in markdown format, ignoring headers and footers. Tables should be expressed in HTML format, formulas in the document should be represented using {\LaTeX} format, and the parsing should be organized according to the reading order.
        \\
        \cmidrule(lr){2-4}
                                    & \textit{v}
                                    & 提取图中的文字。
                                    & Extract the text in the image.
        \\
        \midrule

        \multirow{6}{*}{\textbf{\makecell{Information \nextline Extraction}}}
                                    & \textit{i}
                                    & 输出 $\mathrm{Key}$ 的值。
                                    & Output the value of $\mathrm{Key}$.
        \\
        \cmidrule(lr){2-4}
                                    & \textit{ii}
                                    & 提取图片中的: [`key1',`key2', \dots] 的字段内容，并按照 JSON 格式返回。
                                    & Extract the content of the fields: [`key1',`key2', \dots] from the image and return it in JSON format.
        \\
        \cmidrule(lr){2-4}
                                    & \textit{iii}
                                    & 提取图中的字幕。
                                    & Extract the subtitles from the image.
        \\
        \midrule
        \multirow{3}{*}{\textbf{Translation}}
                                    & \textit{i}
                                    & 先解析文档，再将文档内容翻译为中文，其中页眉、页脚忽略，公式用 \LaTeX 格式表示，表格用html格式表示。
                                    & First parse the document, then translate its content into Chinese. Ignore headers and footers; represent equations in \LaTeX; and render tables in HTML format.
        \\
        \cmidrule(lr){2-4}
                                    & \textit{ii}
                                    & 提取图中文字，并将其翻译成中文/英文。
                                    & Extract all text from the image and translate it into Chinese/English.
        \\
        \bottomrule
    \end{tabular}
\end{table}

\section{CHAOS-Bench Annotation Details}
\label{supp:sec:chaos}

CHAOS-Bench is designed to evaluate whether OCR-centric VLMs can faithfully preserve visually observed text of the provided images when visual evidence conflicts with language priors. The annotation process starts from document pages rendered by a PDF rendering tool. Annotators first parse the rendered page and select valid words from the page content. For each selected word, annotators modify part of its characters directly on the rendered image, transforming the original meaningful word into a meaningless visual word. For example, the word \textit{document} can be modified into \textit{dacument}. In this way, the perturbed word remains visually similar to the original word but no longer forms a semantically valid word, creating a controlled conflict between the visual evidence and the language prior.

\begin{figure*}[t]
    \centering
    \includegraphics[width=\textwidth]{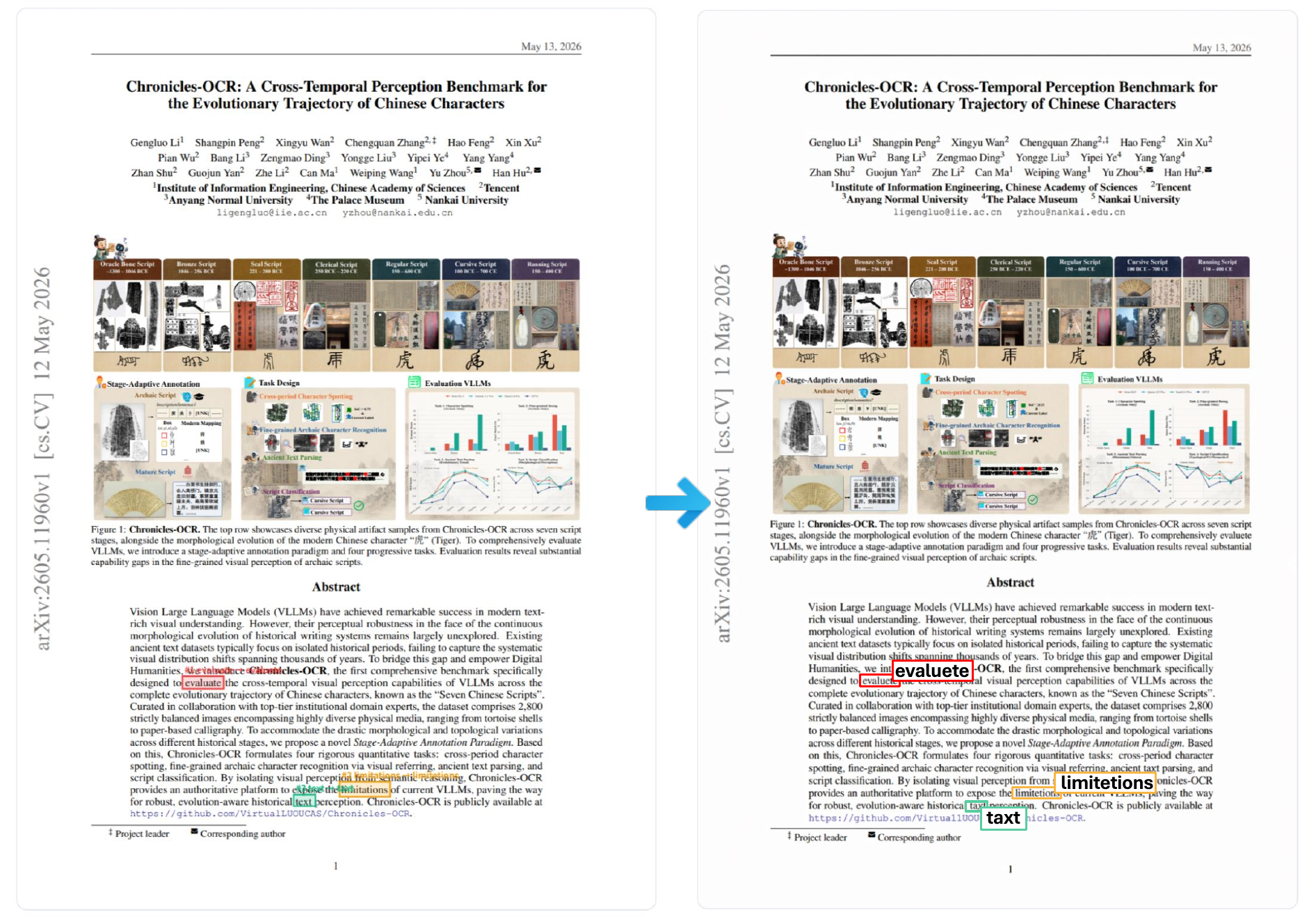}
    \vspace{-1em}
    \caption{
        \textbf{CHAOS-Bench annotation process.}
        Annotators modify characters in rendered document images to create meaningless perturbed words, record the word locations and before/after text, and use the perturbed pages to evaluate seen-text faithfulness under conflicts between visual evidence and language priors.
    }
    \label{fig:chaos_annotation}
\end{figure*}

As illustrated in~\cref{fig:chaos_annotation}, each edited page is saved as a perturbed image, together with annotation metadata including the location of each modified word, the original word, and the corresponding perturbed word. These annotations allow us to evaluate whether a model output faithfully contains the visually observed perturbed word. A model that strictly follows the seen-text principle should reproduce the perturbed word in its parsed output, while a model biased toward language priors may recover the original meaningful word or generate another plausible word that does not appear in the image.
CHAOS-Bench has been released as part of the HunyuanOCR benchmark suite at:
\url{https://github.com/Tencent-Hunyuan/HunyuanOCR/tree/main/benchmarks/CHAOS-Bench}.

\clearpage
\section{Reinforcement Learning Details}
\label{supp:sec:RL_details}

\subsection{RL Setup}
\label{supp:subsec:RL_setup}

The detailed training setup for RL is listed in~\cref{supp:tab:rl_setup}. We adopt a constant learning rate schedule with the Adam~\cite{Adam_2014} optimizer, a large global batch size, and long-context settings. For rollout generation, we use a temperature of $1.0$ with full-vocabulary sampling (i.e., top-$p = 1.0$ and top-$k = -1$), and sample $n = 16$ responses per prompt to obtain a diverse set of candidates for advantage calculation and policy updates. All RL experiments are performed on a cluster of GPUs with TFLOPS of BF16 compute equivalent to about 30 A100 GPUs, utilizing PyTorch 2.10.0~\cite{PyTorch_2019} and CUDA 12.9.

Compared with HunyuanOCR, HunyuanOCR-1.5 differs substantially in its RL configuration. First, we adopt a \textbf{vLLM}~\cite{vLLM_2023} inference backend together with an \textbf{FSDP}~\cite{FSDP_2023} training backend, in preparation for open-sourcing the training pipeline. Beyond infrastructure, we place particular emphasis on both the \textit{effectiveness} and the \textit{stability} of RL training. For \textit{effectiveness}, we use a larger rollout size ($n=16$) to sample a more diverse set of candidate responses, adopt a larger learning rate ($4\times10^{-6}$) for more aggressive optimization, and enlarge the input context length to support higher-resolution images and a broader range of tasks. For \textit{stability}, we observe that even for a 1B-parameter model, the train--inference mismatch phenomenon~\cite{RL_collapse_2025, Rollout_Training_Mismatch_2025} is non-negligible and has a noticeable impact on the final performance, and thus deserves careful treatment. To this end, we switch the RL algorithm to \textbf{IcePop}~\cite{IcePop_2025}, incorporate a KL loss term, and enlarge the training batch size to stabilize policy updates. In addition, we replace top-$k$ sampling in HunyuanOCR~\cite{HunyuanOCR_2025} with \textbf{full-vocabulary rollout sampling}, so as to remove the distributional discrepancy between the training and inference backends induced by truncated sampling~\cite{DeepSeek_V3_2_2025}. More detailed design considerations and empirical analyses of the RL configuration are provided in~\cref{supp:subsec:RL_dynamics}.

\begin{table}[bht!]
    \centering
    \small
    \caption{
        \textbf{Reinforcement Learning Training Configuration of HunyuanOCR-1.5.}
    }
    \label{supp:tab:rl_setup}
    {
        \renewcommand{\arraystretch}{0.98}
        \setlength{\tabcolsep}{22pt}
        \resizebox{\columnwidth}{!}{
            \begin{tabular}{c l >{\centering\arraybackslash}p{0.28\textwidth}}
                \toprule
                \textbf{Group} & \textbf{Setting}         & \textbf{Value}   \\
                \midrule

                \multirow{3}{*}{{\small\textit{Model}}}
                               & Tensor Parallelism (TP)  & 1                \\
                               & Micro batch size / GPU   & 1                \\
                               & PPO mini-batch size      & 512              \\
                \midrule
                \multirow{5}{*}{{\small\textit{Optimization}}}
                               & Training batch size      & 512              \\
                               & Optimizer                & Adam             \\
                               & Learning rate            & $4\times10^{-6}$ \\
                               & LR schedule              & Constant         \\
                               & Warm-up steps            & 0                \\
                \midrule
                \multirow{3}{*}{{\small\textit{Sequence}}}
                               & Sequence length          & 16384            \\
                               & Max prompt length        & 8192             \\
                               & Max response length      & 8192             \\
                \midrule
                \multirow{6}{*}{{\small\textit{RL Algorithm}}}
                               & Advantage estimator      & IcePop           \\
                               & Imp.\ ratio cap range    & $[0.2,\, 5.0]$   \\
                               & Use KL in reward         & False            \\
                               & Use KL loss              & True             \\
                               & KL loss coefficient      & 0.001            \\
                               & Entropy coefficient      & 0.0              \\
                \midrule
                \multirow{4}{*}{{\small\textit{Rollout}}}
                               & Samples per prompt ($n$) & 16               \\
                               & Temperature              & 1.0              \\
                               & Top-$p$                  & 1.0              \\
                               & Top-$k$                  & $-1$             \\
                \midrule
                \multirow{1}{*}{{\small\textit{Data}}}
                               & Training data size       & 50K              \\
                \midrule
                \multirow{6}{*}{{\small\textit{Backend}}}
                               & Inference backend        & vLLM             \\
                               & Training backend         & FSDP             \\
                               & Param offload            & True             \\
                               & Optimizer offload        & True             \\
                               & Model dtype              & bfloat16         \\
                               & Rollout dtype            & bfloat16         \\

                \bottomrule
            \end{tabular}
        }
    }
    \vspace{-2mm}
\end{table}

\subsection{RL Reward Design}
\label{supp:subsec:RL_reward_design}

In this subsection, we provide a detailed description of the reward design used in the reinforcement learning (RL) training of HunyuanOCR-1.5, serving as a supplement to~\cref{subsec:RL} in the main text. We first describe the overall reward computation flow, and then detail the reward for each task type.

Given a prompt $p$ and a rollout response $r$, the reward is computed as soon as $r$ is generated, rather than waiting until all rollouts have finished, which substantially improves training efficiency. We first apply the degeneration-suppression reward described in~\cref{subsubsec:degeneration_reward}: an overlong or repetitive response is penalized immediately and does not enter the subsequent stages, which removes redundant computation for degenerate rollouts. The remaining responses are then routed by task type, and the corresponding task-specific reward is computed for each of them. The following paragraphs describe the reward of each task in detail.

\mypara{Spotting.}
Text spotting requires joint bounding-box localization and text recognition. We first detect the format of the response, supporting both the new JSON output format introduced in HunyuanOCR-1.5 and the original HunyuanOCR format, and then extract the predicted boxes $\mathcal{B}_{p}$ and the ground-truth boxes $\mathcal{B}_{g}$ accordingly. Each predicted box $b_p\in\mathcal{B}_{p}$ is assigned to a ground-truth box $b_g\in\mathcal{B}_{g}$ by maximizing the Intersection over Union (IoU). For a matched pair with texts $t_{p}$ and $t_{g}$ (obtained by applying a shared text-normalization step to $b_p$ and $b_g$), the pair-level reward is computed as

\vspace{-1.0em}
\begin{equation}
    s
    =
    1 - \frac{\mathrm{EditDist}(t_{p}, t_{g})}{\max\!\left(|t_{p}|,\, |t_{g}|\right)}\,.
    \label{supp:eq:spotting_reward}
\end{equation}
\vspace{-0.8em}

Any unmatched prediction or ground-truth box is scored against an empty string, contributing $s=0$ (equivalently, a normalized distance of $1$) to the average. The final reward is the mean pair-level score across all evaluated pairs, providing a balanced measure of both localization and recognition accuracy. The full procedure of text spotting reward computation is summarized in~\cref{alg:spotting_reward}.

\begin{algorithm}[!t]
    \caption{Text Spotting Reward}
    \label{alg:spotting_reward}
    \begin{algorithmic}[1]
        \Statex \textbf{Inputs:} rollout response \textit{response}, reference answer \textit{ref\_answer}, IoU threshold $\tau$ (default $0.5$).
        \Statex \textbf{Output:} spotting reward $R\in[0,1]$.
        \Statex
        \Function{SpottingReward}{\textit{response}, \textit{ref\_answer}, $\tau$}
        \State $\mathcal{B}_{p}\leftarrow\textsc{ParseBoxes}(\textit{response})$,\quad $\mathcal{B}_{g}\leftarrow\textsc{ParseBoxes}(\textit{ref\_answer})$ \mycomment{format-aware box parsing}
        \State $\mathcal{M}\leftarrow\emptyset$;\quad $S\leftarrow 0$;\quad $c\leftarrow 0$ \mycomment{matched GT set, total pair reward, counter}
        \ForAll{$b_{p}\in\mathcal{B}_{p}$}
        \State $b_{g}^{\ast}\leftarrow\arg\max_{b_{g}\in\mathcal{B}_{g}\setminus\mathcal{M}}\textsc{IoU}(b_{p},b_{g})$
        \If{$b_{g}^{\ast}$ exists \textbf{and} $\textsc{IoU}(b_{p},b_{g}^{\ast})\ge\tau$}
        \State $t_{p}\leftarrow\textsc{Norm}(b_{p}.\text{text})$,\quad $t_{g}\leftarrow\textsc{Norm}(b_{g}^{\ast}.\text{text})$ \mycomment{text normalization}
        \State $s\leftarrow 1-\dfrac{\mathrm{EditDist}(t_{p},t_{g})}{\max(|t_{p}|,\,|t_{g}|)}$ \mycomment{\cref{supp:eq:spotting_reward}: matched pair}
        \State $\mathcal{M}\leftarrow\mathcal{M}\cup\{b_{g}^{\ast}\}$
        \Else
        \State $s\leftarrow 0$ \mycomment{unmatched prediction: compared against empty text}
        \EndIf
        \State $S\leftarrow S+s$;\quad $c\leftarrow c+1$
        \EndFor
        \ForAll{$b_{g}\in\mathcal{B}_{g}\setminus\mathcal{M}$}
        \State $S\leftarrow S+0$;\quad $c\leftarrow c+1$ \mycomment{unmatched GT box: $s=0$}
        \EndFor
        \State $R\leftarrow S/(c+\epsilon)$ \mycomment{mean pair-level reward}
        \State \Return $\textsc{Clamp}(R,0,1)$
        \EndFunction
    \end{algorithmic}
\end{algorithm}

\mypara{Parsing.}
For document parsing, the output may simultaneously contain plain text, tables, and charts, whose correctness is not well captured by a single text-level metric. Following~\cref{subsubsec:factual_reward}, we parse the response and the reference into a plain-text part and a set of special elements, score the plain text with a normalized edit-distance reward and each special element (table or chart) with its element-specific reward, and combine them into the structure-aware parsing reward of~\cref{eq:parsing_reward}. The overall procedure is summarized in~\cref{alg:parsing_reward}.

\begin{algorithm}[!t]
    \caption{Structure-Aware Parsing Reward}
    \label{alg:parsing_reward}
    \begin{algorithmic}[1]
        \Statex \textbf{Inputs:} rollout response \textit{response}, reference answer \textit{ref\_answer}, weights $\lambda_{1},\lambda_{2}$.
        \Statex \textbf{Output:} parsing reward $R_{\text{parse}}$.
        \Statex
        \Function{ParsingReward}{\textit{response}, \textit{ref\_answer}, $\lambda_{1},\lambda_{2}$}
        \State $(u_{p},\mathcal{E}_{p})\leftarrow\textsc{Split}(\textit{response})$,\quad $(u_{g},\mathcal{E}_{g})\leftarrow\textsc{Split}(\textit{ref\_answer})$ \mycomment{plain text $u$, special elements $\mathcal{E}$}
        \State $R_{\text{text}}\leftarrow 1-\textsc{NormEditDist}(u_{p},u_{g})$ \mycomment{plain-text reward}
        \State $\{(e_{j}^{p},e_{j}^{g})\}_{j=1}^{M}\leftarrow\textsc{PairElements}(\mathcal{E}_{p},\mathcal{E}_{g})$ \mycomment{align pred/GT elements in order; missing side $\to$ empty}
        \State $S\leftarrow 0$
        \For{$j=1,\ldots,M$}
        \If{$e_{j}^{g}$ is a table}
        \State $R_{\text{elem}}\leftarrow 0.5\,R_{\text{content}}(e_{j}^{p},e_{j}^{g}) + 0.5\,R_{\text{struct}}(e_{j}^{p},e_{j}^{g})$ \mycomment{destylized content + 1D-probe structure}
        \Else \mycomment{$e_{j}^{g}$ is a chart}
        \State $R_{\text{elem}}\leftarrow \textsc{SCRM\_mAP}(e_{j}^{p},e_{j}^{g})$ \mycomment{CSV conversion + order-invariant matching}
        \EndIf
        \State $S\leftarrow S + R_{\text{elem}}$
        \EndFor
        \State $R_{\text{parse}}\leftarrow \lambda_{1}\,R_{\text{text}} + \lambda_{2}\,\dfrac{S}{M}$ \mycomment{\cref{eq:parsing_reward}}
        \State \Return $R_{\text{parse}}$
        \EndFunction
    \end{algorithmic}
\end{algorithm}

\mypara{Visual question answering.}
For visual question answering, the reward is a binary consistency judgment produced by an LLM-as-a-judge, which decides whether the model answer is semantically consistent with the reference answer and assigns $1$ or $0$ accordingly. The judging prompt is given in the box below.

\begin{tcolorbox}[breakable, colback=gray!5, colframe=gray!55!black, title=Judging prompt for visual question answering, label=supp:box:vqa_prompt, fonttitle=\bfseries\small, fontupper=\small]
    \textbf{\# Task Description}\\
    You are given a question and two candidate answers: the \texttt{[Standard Answer]} and the \texttt{[Model Answer]}. Your goal is to determine whether the two are consistent in their core semantics.

    \medskip
    \textbf{\# Procedure}
    \begin{enumerate}[leftmargin=1.4em, itemsep=1pt, topsep=1pt]
        \item \textbf{Extract the core information.} From each answer, extract the key content that directly addresses the question. Ignore irrelevant pleasantries, prefixes, and summary statements.
        \item \textbf{Judge consistency.}
              \begin{itemize}[leftmargin=1.2em, itemsep=1pt, topsep=1pt]
                  \item \textit{Semantic consistency (default criterion):} the core information of the two answers should convey the same meaning. Differences in wording, order, and level of detail are allowed, provided they do not alter the meaning, omit key information, or cause misunderstanding.
                  \item \textit{Character-level consistency (strict criterion):} when the question requires exact transcription (e.g., ``What is the text in the image?''), the two answers must match exactly in characters and important punctuation.
              \end{itemize}
        \item \textbf{Output.}
              \begin{itemize}[leftmargin=1.2em, itemsep=1pt, topsep=1pt]
                  \item \textit{Reasoning:} briefly explain in one sentence why the answers are judged ``consistent'' or ``inconsistent''.
                  \item \textit{Judgement:} output \texttt{Judgement: 1} if consistent, otherwise \texttt{Judgement: 0}.
              \end{itemize}
    \end{enumerate}

    \medskip
    \texttt{[Question]}: \{question\}\\
    \texttt{[Standard Answer]}: \{gt\}\\
    \texttt{[Model Answer]}: \{answer\}\\
\end{tcolorbox}

\mypara{Translation.}
For text-image translation, the reward is a soft score in $[0,5]$ produced by an LLM-as-a-judge, which rates the model translation against the reference along semantic accuracy, fluency, cultural adequacy, and terminology consistency. The score is then normalized to $[0,1]$. The judging prompt is given below.

\begin{tcolorbox}[breakable, colback=gray!5, colframe=gray!55!black, title=Judging prompt for text-image translation, label=supp:box:translation_prompt, fonttitle=\bfseries\small, fontupper=\small]
    You are given a \textbf{target language}, a \textbf{source text} (which may contain multiple source languages), a \textbf{reference translation}, and the \textbf{scoring criteria and rubric} (0--5 points). Following the format of the \textbf{scoring example} and referring to the reference translation, score the model translation. Output only ``Reasoning'' and ``Final score'', and nothing else.

    \medskip
    \textbf{Scoring criteria}
    \begin{enumerate}[leftmargin=1.4em, itemsep=1pt, topsep=1pt]
        \item \textbf{Semantic accuracy:} faithfully conveys the meaning of the source.
        \item \textbf{Fluency:} reads naturally in the target language.
        \item \textbf{Cultural adequacy:} wording is appropriate for the target-language culture.
        \item \textbf{Terminology consistency:} technical terms and widely recognized names are preserved correctly.
    \end{enumerate}

    \textbf{Scoring rubric (0--5 points).}
    Score strictly, with higher scores for higher-quality translations. Semantic accuracy is fundamental: other dimensions cannot exceed the semantic-accuracy score. Use the reference translation as an anchor: on par with the reference may receive $4$; exceeding it may receive $4.0$--$5.0$. If the model answer contains image descriptions, format changes, or a restatement of the source, ignore them and evaluate only the extracted translation. Numbers and LaTeX formulas need not be translated.
    \begin{itemize}[leftmargin=1.2em, itemsep=1pt, topsep=1pt]
        \item \textbf{Very poor (0):} severe errors unrelated to the source, or harmful content (e.g., large-scale missing content or incomprehensible output).
        \item \textbf{Poor (1):} no severe errors and essentially harmless, but low quality on key dimensions (e.g., wrong key-term translation, many missing or mistranslated key contents, or hard to understand).
        \item \textbf{Fair (2):} basically meets the requirements with minor errors, such as a few wrong non-technical words, errors in less-known names, or slight loss of non-essential content.
        \item \textbf{Good (3):} good on all dimensions, fluent and accurate, faithfully conveying the source meaning with context-appropriate wording.
        \item \textbf{Very good (4.0--5.0):} high accuracy and fluency, fully conforming to target-language usage, with proper cultural adaptation and near-perfect performance on all dimensions.
    \end{itemize}
    \medskip
    \textbf{Question}: \{question\}\\
    \textbf{Target language}: \{reference\}\\
    \textbf{Source text}: \{text\}\\
    \textbf{Reference translation}: \{gt\}\\
    \textbf{Model translation}: \{answer\}
\end{tcolorbox}

\mypara{Repeated fragment detection.}
As part of the degeneration-suppression reward, we detect tail repetition in the response: a short unit of length at most \textit{max\_unit} that repeats at least \textit{min\_repeats} times consecutively at the end of the sequence. Only the tail is inspected to keep the cost low, which makes the check suitable for frequent calls during generation. A response flagged by this detector receives a reward of zero. The procedure is summarized in~\cref{alg:tail_repetition}.

\begin{algorithm}[!t]
    \caption{Tail Repetition Detection}
    \label{alg:tail_repetition}
    \begin{algorithmic}[1]
        \Statex \textbf{Inputs:} text \textit{text}, minimum repeats \textit{min\_repeats} (default $8$), maximum unit length \textit{max\_unit} (default $256$).
        \Statex \textbf{Output:} boolean flag indicating tail repetition.
        \Statex
        \Function{HasTailRepetition}{\textit{text}, \textit{min\_repeats}, \textit{max\_unit}}
        \State $n\leftarrow|\textit{text}|$
        \If{$n < 2\cdot\textit{min\_repeats}$}
        \State \Return \textsc{False}
        \EndIf
        \State $\textit{upper}\leftarrow\min(\textit{max\_unit},\ \lfloor n/\textit{min\_repeats}\rfloor)$
        \For{$\ell=1,\ldots,\textit{upper}$}
        \State $\textit{unit}\leftarrow\textit{text}[\,n-\ell:n\,]$ \mycomment{candidate repeating unit at the tail}
        \If{$\textit{unit}$ is empty or whitespace-only}
        \State \textbf{continue}
        \EndIf
        \State $\textit{ok}\leftarrow\textsc{True}$
        \For{$k=2,\ldots,\textit{min\_repeats}$}
        \If{$\textit{text}[\,n-\ell k : n-\ell(k-1)\,]\ne\textit{unit}$}
        \State $\textit{ok}\leftarrow\textsc{False}$;\quad \textbf{break}
        \EndIf
        \EndFor
        \If{$\textit{ok}$}
        \State \Return \textsc{True} \mycomment{a short unit repeats $\ge\textit{min\_repeats}$ times at the tail}
        \EndIf
        \EndFor
        \State \Return \textsc{False}
        \EndFunction
    \end{algorithmic}
\end{algorithm}

\subsection{RL Dynamics}
\label{supp:subsec:RL_dynamics}

\begin{figure}[t!]
    \centering
    \includegraphics[width=\linewidth]{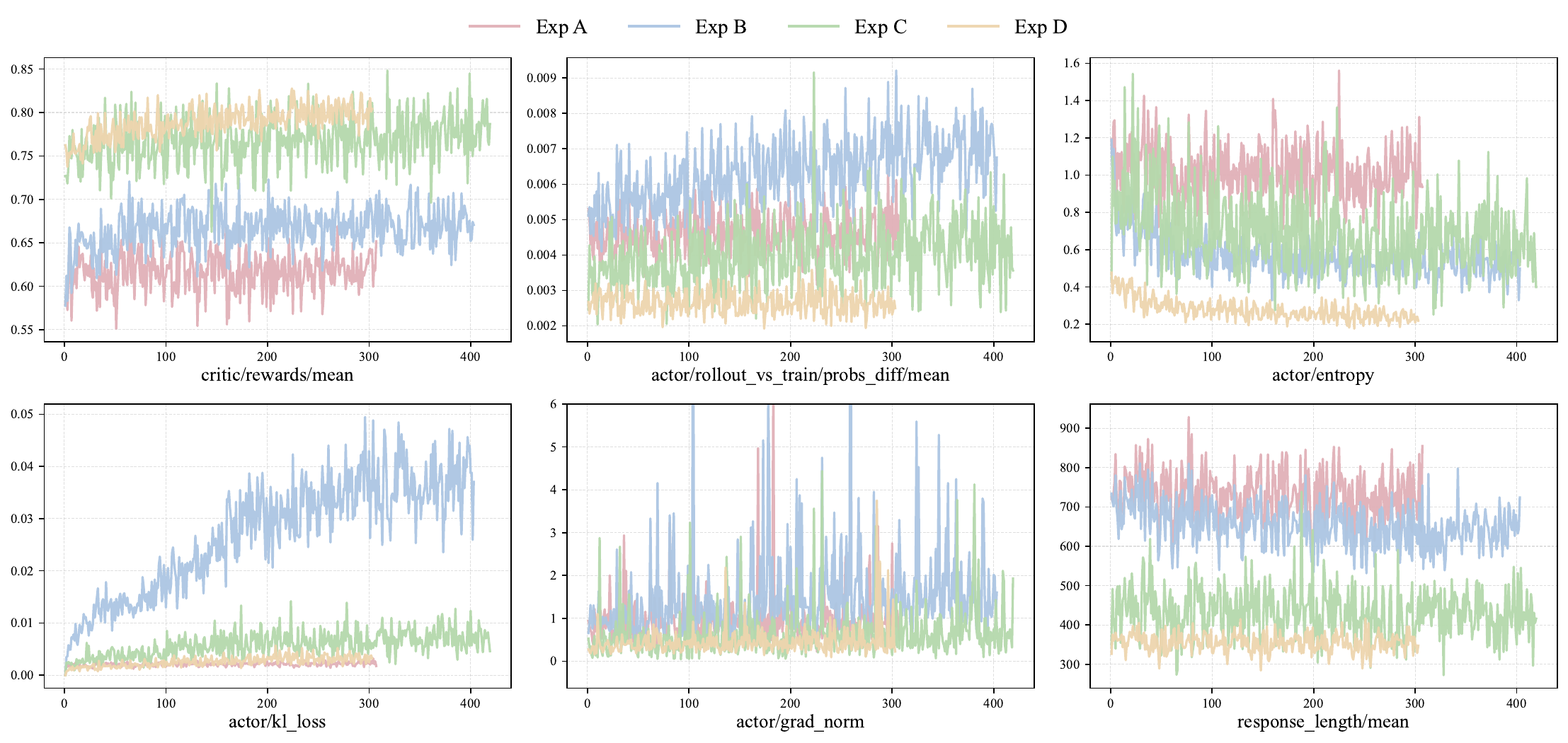}
    \vspace{-1.6em}
    \caption{
        \textbf{Training dynamics of the exploratory RL experiments for HunyuanOCR-1.5.}
        Each panel reports a different training metric:
        reward (\texttt{critic/rewards/mean}),
        train--inference mismatch (\texttt{actor/rollout\_vs\_train/probs\_diff/mean}),
        policy entropy (\texttt{actor/entropy}),
        KL loss (\texttt{actor/kl\_loss}),
        gradient norm (\texttt{actor/grad\_norm}),
        and response length (\texttt{response\_length/mean}).
        Exp~A--C are exploratory runs, while Exp~D is the finalized training configuration.
    }
    \label{fig:RL_dynamics}
    \vspace{-5pt}
\end{figure}

In this subsection, we discuss the training dynamics observed during the early-stage RL experiments, and how these observations motivated the adjustments that led to our final training configuration. The dynamics of each main training runs are provided in~\cref{fig:RL_dynamics}. The horizontal axis denotes the training iteration, and the vertical axis of each panel reports a different training metric. Among the four runs, Exp~A to Exp~C are exploratory experiments, while Exp~D is the finalized training configuration. Their configurations are summarized in~\cref{supp:tab:rl_dynamics_exp}, where each experiment is described by its difference from the previous one.

\begin{table}[h!]
    \centering
    \small
    \caption{
        \textbf{Configuration of the exploratory RL experiments.}
        Each row lists the change relative to the previous experiment; unlisted settings are inherited.
    }
    \label{supp:tab:rl_dynamics_exp}
    {
        \renewcommand{\arraystretch}{1.1}
        \setlength{\tabcolsep}{12pt}
        \resizebox{0.8\columnwidth}{!}{
            \begin{tabular}{c l}
                \toprule
                \textbf{Exp} & \textbf{Configuration / difference from the previous experiment}                                  \\
                \midrule
                Exp~A        & Baseline from HunyuanOCR: $\text{bs}=256$, $\text{lr}=1\times10^{-6}$, $\text{kl\_coef}=0$, $n=8$ \\
                Exp~B        & $\text{lr}=1\times10^{-5}$                                                                        \\
                Exp~C        & $\text{lr}=4\times10^{-6}$, $\text{kl\_coef}=0.001$, $n=16$, updated training data                \\
                Exp~D        & $\text{bs}=512$, rollout-based sampling and filtering of the training data                        \\
                \bottomrule
            \end{tabular}
        }
    }
    \vspace{-2mm}
\end{table}

We start from the RL training configuration of HunyuanOCR~\cite{HunyuanOCR_2025} and design Exp~A, shown as the red curves in~\cref{fig:RL_dynamics}. For the HunyuanOCR-1.5 SFT model, both \texttt{kl\_loss} and \texttt{rewards/mean} stay almost flat throughout training, indicating that the model barely improves under this configuration. To understand this behavior, we further compare the model weights before and after training in terms of their change ratio and similarity, and find that the weights change only marginally. This suggests that the model is effectively stuck, which motivates us to study the effect of a larger learning rate.

We then design Exp~B, raising the learning rate to $1\times10^{-5}$ while keeping other settings unchanged. As shown by the blue curves, \texttt{rewards/mean} is consistently higher than in Exp~A, indicating more effective optimization. However, \texttt{kl\_loss} grows substantially, and the train--inference mismatch metric \texttt{probs\_diff/mean} rises, indicating that training stability degrades. This instability is itself a problem that needs to be addressed.

To improve stability, we design Exp~C, lowering the learning rate to $4\times10^{-6}$ and adding a KL loss together with a larger sample count $n$ to regularize the updates. Note that this step also updates the training data, so \texttt{rewards/mean} is not directly comparable with the earlier experiments. As shown by the green curves, \texttt{probs\_diff/mean} remains stable, \texttt{entropy} decreases steadily, and \texttt{kl\_loss} rises smoothly and settles at a reasonable level, indicating improved training stability. Nevertheless, \texttt{rewards/mean} shows no clear gain, so the training effectiveness is still unsatisfactory and calls for further refinement.

Finally, we design Exp~D, replacing the training data with the data obtained by the sampling-and-filtering strategy described in the main text and enlarging the batch size to $512$. As shown by the orange curves, \texttt{rewards/mean} improves substantially, indicating clearly more effective training, while \texttt{kl\_loss} and \texttt{probs\_diff/mean} remain stable and vary more smoothly than in Exp~C, indicating better stability as well. Based on these observations, we adopt this configuration as our final RL setup, which realizes the balance between \textit{effectiveness} and \textit{stability} targeted in~\cref{supp:subsec:RL_setup}.

\subsection{Task-wise Performance Improvements}
\label{supp:subsec:RL_improvements}

Benefiting from the effective and stable RL training described above, HunyuanOCR-1.5 achieves consistent improvements over SFT checkpoint across a broad range of tasks. As shown in~\cref{supp:tab:rl_task_improvements}, the RL stage improves the model on end-to-end document parsing, text spotting, ancient-script recognition, structured chart parsing, and video subtitle extraction, covering both core OCR abilities and boundary capabilities. These gains show that the reward system described in~\cref{supp:subsec:RL_reward_design} provides useful and discriminative signals across task types, and that the training configuration converges toward a well-balanced improvement rather than favoring any single task.

\begin{table}[t!]
    \centering
    \small
    \caption{
        \textbf{Task-wise performance improvements from RL training.}
        We compare the SFT checkpoint with the final RL model of HunyuanOCR-1.5 across five representative benchmarks.
    }
    \label{supp:tab:rl_task_improvements}
    {
        \renewcommand{\arraystretch}{1.05}
        \setlength{\tabcolsep}{14pt}
        \resizebox{\columnwidth}{!}{
            \begin{tabular}{l ccccc}
                \toprule
                \textbf{Stage}             &
                \textbf{OmniDocBench v1.6} &
                \textbf{Spotting}          &
                \textbf{Chronicles-OCR}    &
                \textbf{ChartArena}        &
                \textbf{Video Subtitle}
                \\
                \midrule
                SFT                        & 92.92 & 70.1 & 0.50 / 0.73   & 47.1 / 62.2 & 92.4 \\
                RL                         & 94.74 & 71.4 & 0.54 / 0.79   & 48.9 / 64.1 & 93.1 \\
                \hybr
                $\Delta$                   & +1.82 & +1.3 & +0.04 / +0.06 & +1.8 / +1.9 & +0.7 \\
                \bottomrule
            \end{tabular}
        }
    }
    \vspace{-2mm}
\end{table}

\section{Qualitative Examples}
\label{supp:sec:qualitative_examples}

In this section, we present a set of qualitative examples to illustrate the newly supported capabilities and characteristic scenarios of HunyuanOCR-1.5.

\begin{figure*}[t]
    \centering
    \includegraphics[width=0.72\textwidth]{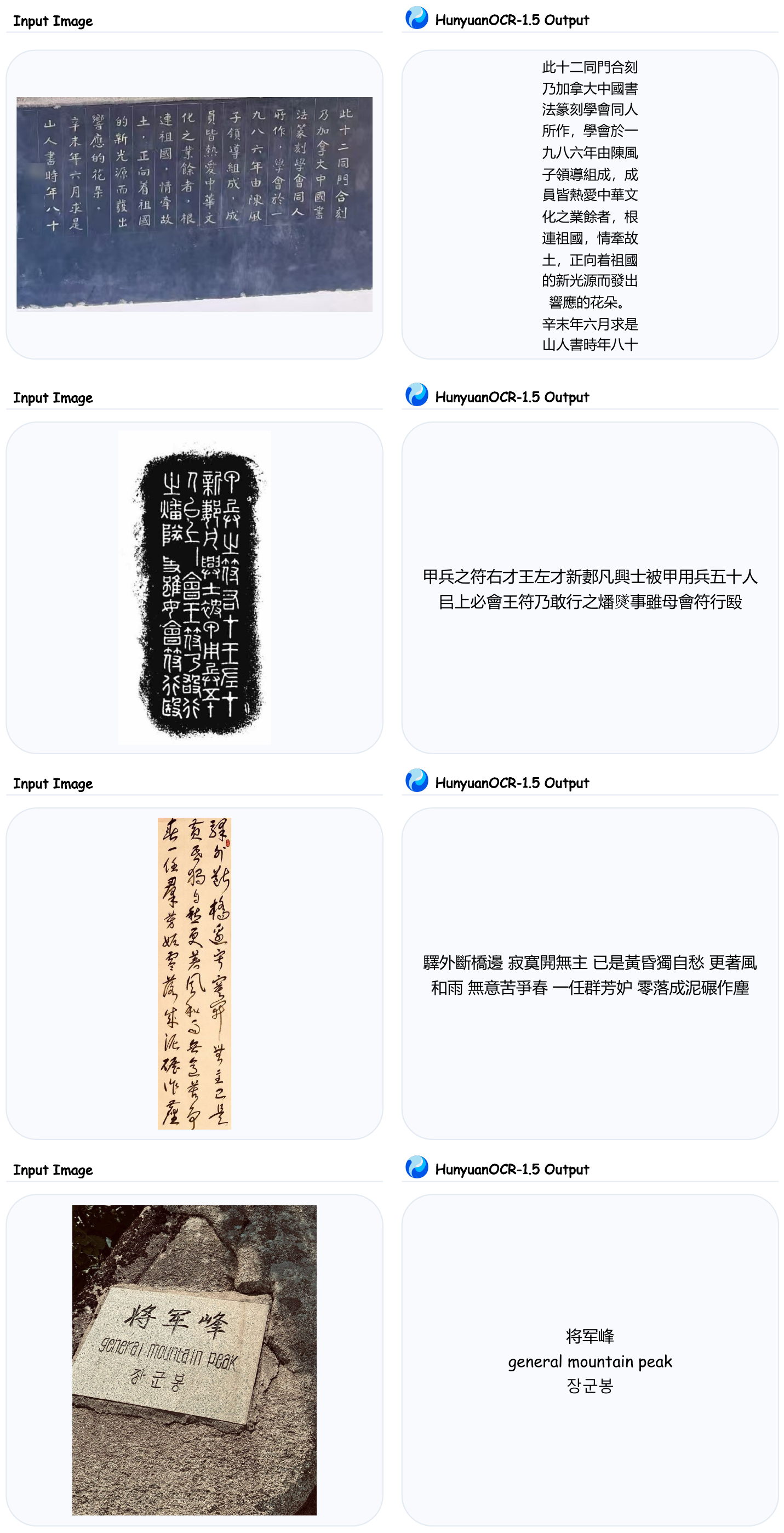}
    \caption{
        \textbf{Qualitative example on Chronicles-OCR~\cite{Chronicles_OCR_2026}.}
        HunyuanOCR-1.5 is able to handle the historical forms of Chinese characters, historical documents, and ancient-script images.
    }
    \label{fig:case_chronicles_ocr_1}
\end{figure*}

\begin{figure*}[t]
    \centering
    \includegraphics[width=0.72\textwidth]{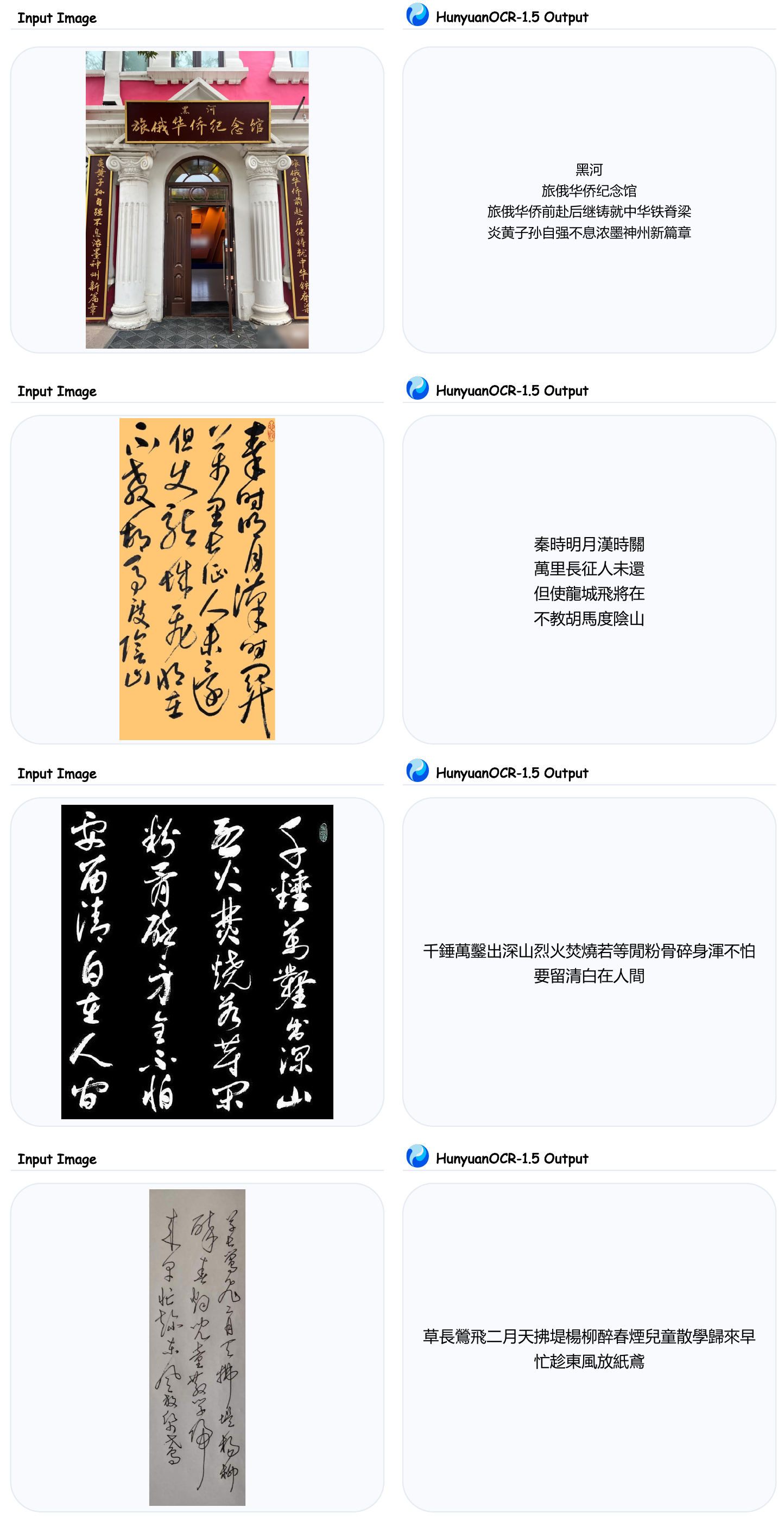}
    \caption{
        \textbf{Qualitative example on Chronicles-OCR~\cite{Chronicles_OCR_2026}.}
        HunyuanOCR-1.5 is able to handle the historical forms of Chinese characters, historical documents, and ancient-script images.
    }
    \label{fig:case_chronicles_ocr_2}
\end{figure*}

\begin{figure*}[t]
    \centering
    \includegraphics[width=0.72\textwidth]{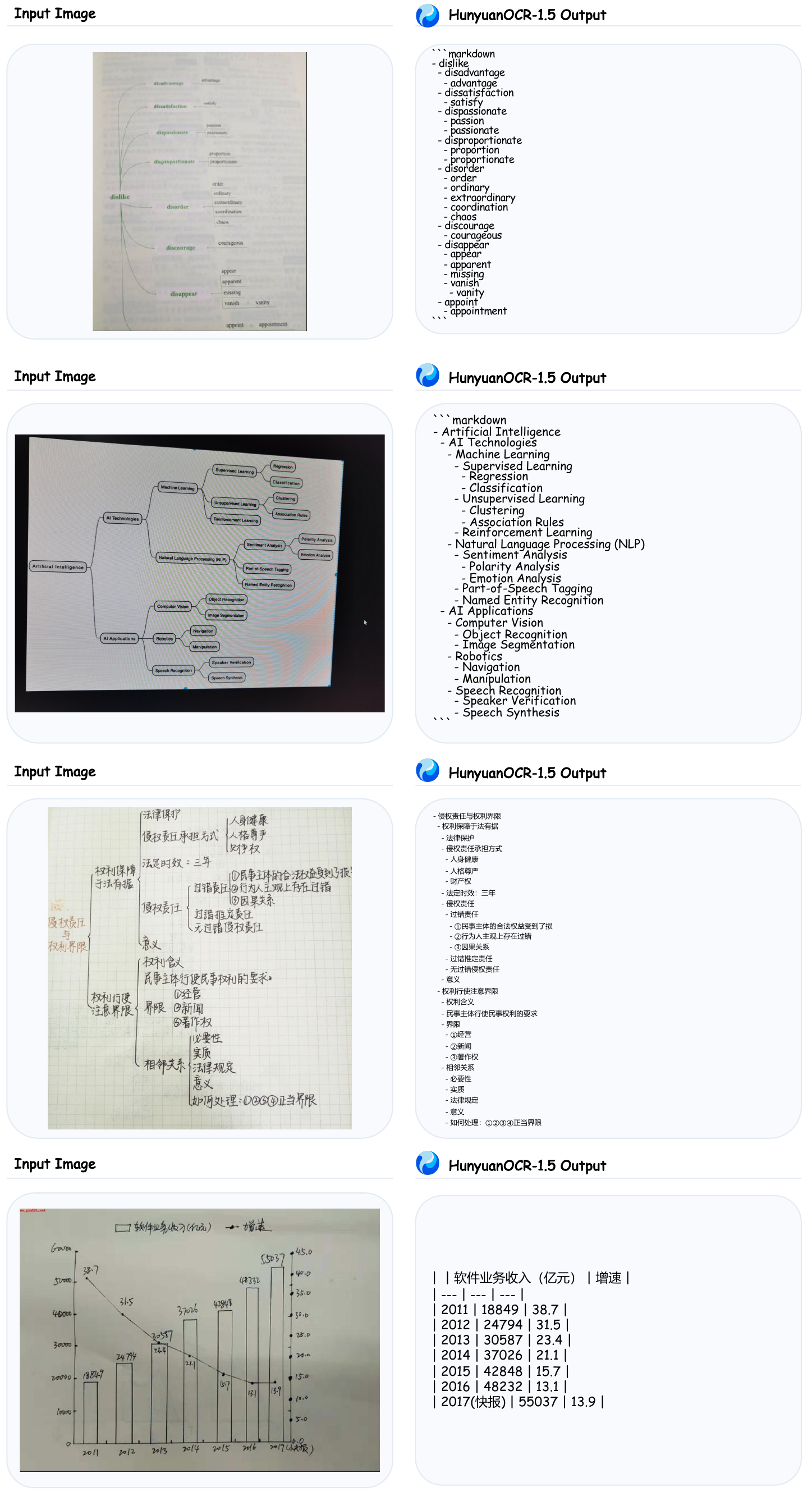}
    \caption{
        \textbf{Qualitative example on ChartArena~\cite{ChartArena_2026}.}
        HunyuanOCR-1.5 is able to parse charts across diverse chart families, visual scenarios, and languages.
    }
    \label{fig:case_chartarena_1}
\end{figure*}

\begin{figure*}[t]
    \centering
    \includegraphics[width=0.72\textwidth]{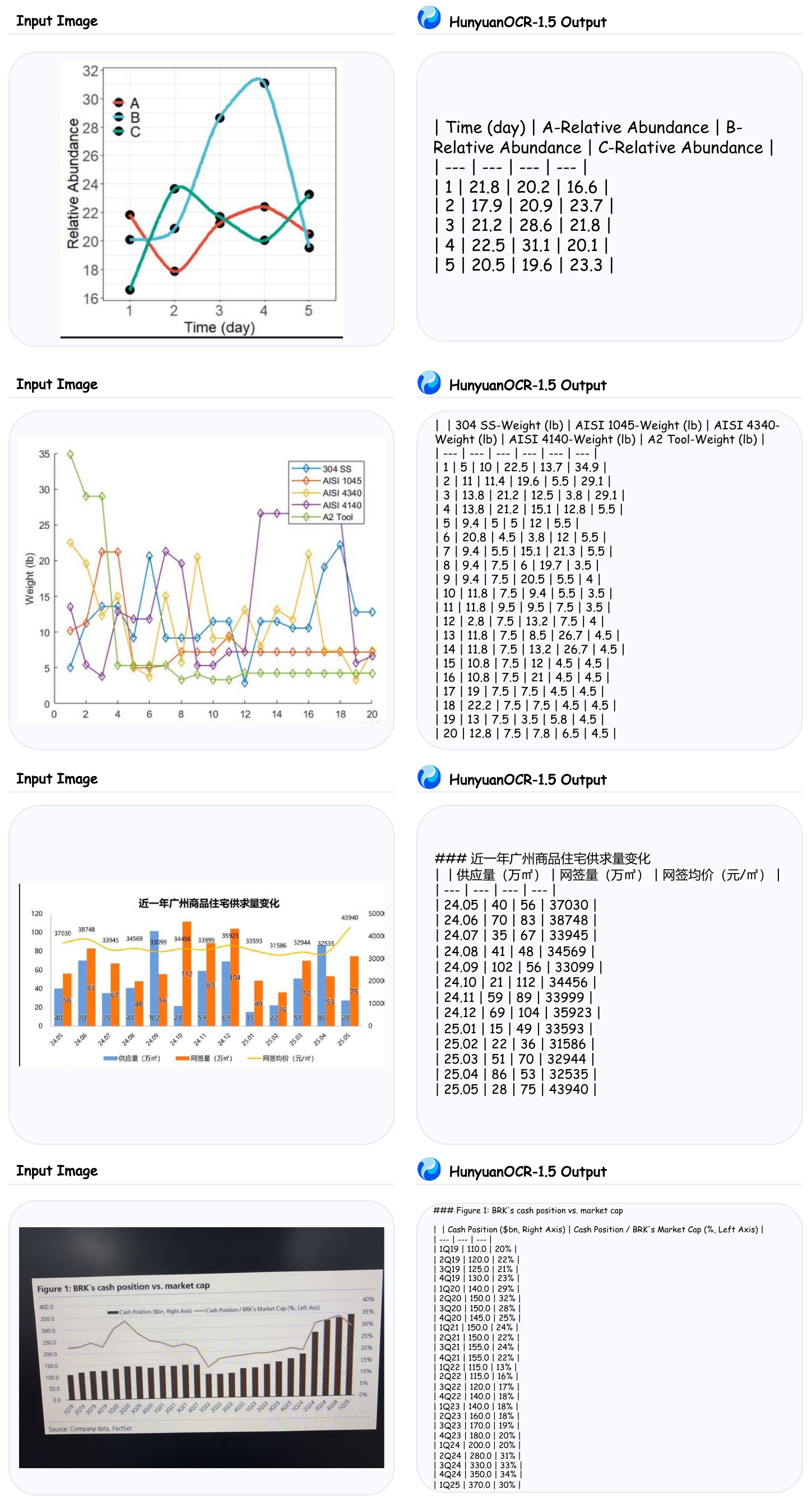}
    \caption{
        \textbf{Qualitative example on ChartArena~\cite{ChartArena_2026}.}
        HunyuanOCR-1.5 is able to parse charts across diverse chart families, visual scenarios, and languages.
    }
    \label{fig:case_chartarena_2}
\end{figure*}

\begin{figure*}[t]
    \centering
    \includegraphics[width=0.85\textwidth]{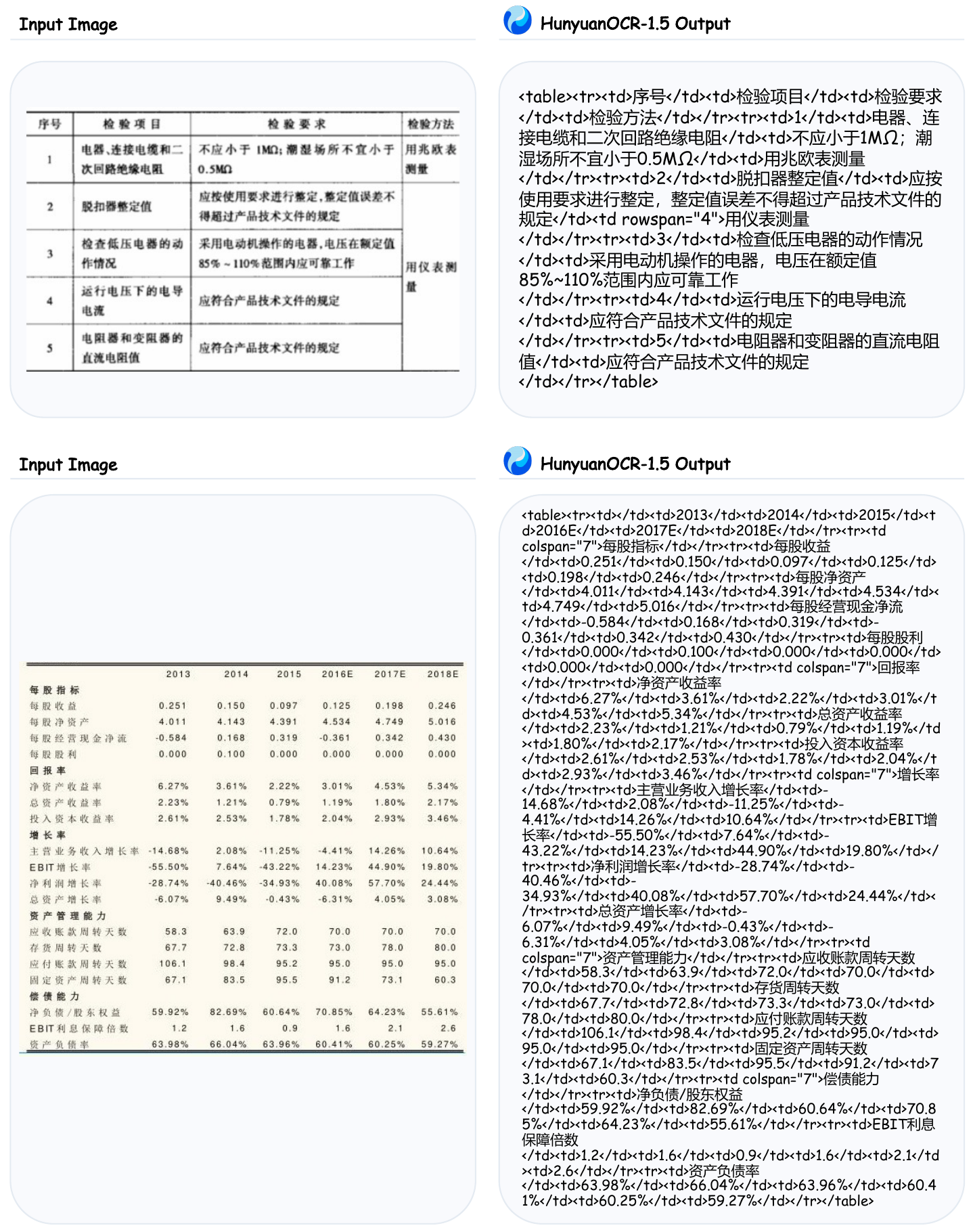}
    \caption{
        \textbf{Qualitative example on TableVerse-5K~\cite{StrucTab_2026}.}
        HunyuanOCR-1.5 is able to parse tables with diverse structures, styles, and content types.
    }
    \label{fig:case_tableverse_1}
\end{figure*}

\begin{figure*}[t]
    \centering
    \includegraphics[width=0.85\textwidth]{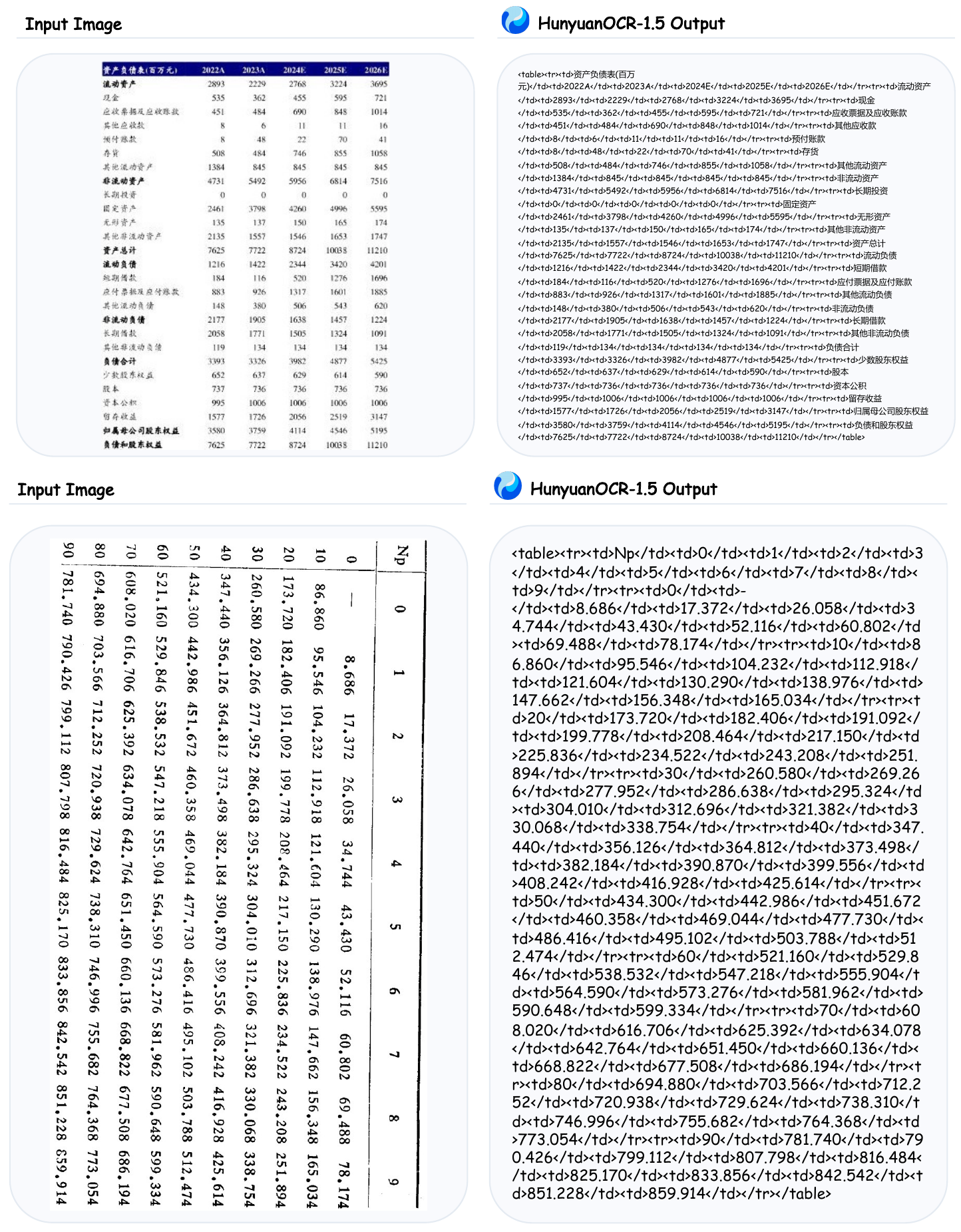}
    \caption{
        \textbf{Qualitative example on TableVerse-5K~\cite{StrucTab_2026}.}
        HunyuanOCR-1.5 is able to parse tables with diverse structures, styles, and content types.
    }
    \label{fig:case_tableverse_2}
\end{figure*}

\end{CJK*}

\end{document}